\documentclass[10pt,journal,compsoc]{IEEEtran}
\ifCLASSOPTIONcompsoc
  \usepackage[nocompress]{cite}
\else
  \usepackage{cite}
\fi
\ifCLASSINFOpdf
   \usepackage[pdftex]{graphicx}
   \DeclareGraphicsExtensions{.pdf,.jpeg,.png}
\else
\fi
\usepackage{amsmath}
\usepackage{array}

\ifCLASSOPTIONcompsoc
 \usepackage[caption=false,font=footnotesize,labelfont=sf,textfont=sf]{subfig}
\else
 \usepackage[caption=false,font=footnotesize]{subfig}
\fi
\usepackage{url}

\usepackage{algorithm}
\usepackage{algorithmic}
\usepackage{amsmath}
\usepackage{amssymb}
\usepackage{mathtools}
\usepackage{siunitx}
\usepackage{array}
\usepackage{multirow}
\usepackage{booktabs}
\usepackage{enumitem}

\DeclareMathOperator{\Beta}{Beta}
\DeclareMathOperator{\Bernoulli}{Ber}

\DeclareMathOperator{\Normal}{\mathcal{N}}
\DeclareMathOperator{\diag}{diag}
\DeclareMathOperator{\sigmoid}{sigmoid}
\DeclareMathOperator{\softmax}{softmax}
\DeclareMathOperator{\mean}{mean}
\DeclareMathOperator{\MI}{MI}
\DeclareMathOperator{\entropy}{H}

\DeclareMathOperator*{\argmax}{arg\,max}

\newcolumntype{C}[1]{>{\centering\let\newline\\\arraybackslash\hspace{0pt}}m{#1}}

\allowdisplaybreaks

\begin{document}
%
\title{Unsupervised Object-Centric Learning from Multiple Unspecified Viewpoints}
%
%
%
%

\author{Jinyang Yuan,
        Tonglin Chen$^*$,
        Zhimeng Shen$^*$,
        Bin Li,
        and Xiangyang Xue
\IEEEcompsocitemizethanks{\IEEEcompsocthanksitem The authors are with the Shanghai Key Laboratory of Intelligent Information Processing
and the School of Computer Science, Fudan University, Shanghai
200433, China.\protect\\
E-mail: \{yuanjinyang, tlchen18, libin, xyxue\}@fudan.edu.cn, zmshen22@m.fudan.edu.cn}
\thanks{Manuscript received Month Day, Year; revised Month Day, Year.\\
(Corresponding authors: Bin Li and Xiangyang Xue.)}}

%
%

\markboth{Journal of \LaTeX\ Class Files,~Vol.~14, No.~8, August~2015}%
{Shell \MakeLowercase{\textit{et al.}}: Bare Demo of IEEEtran.cls for Computer Society Journals}
%


\IEEEpubid{\begin{minipage}{\textwidth}\ \\[12pt]
	\copyright~2024 IEEE. Personal use of this material is permitted. Permission from IEEE must be obtained for all other uses, including reprinting/republishing this material for advertising or promotional purposes, collecting new collected works for resale or redistribution to servers or lists, or reuse of any copyrighted component of this work in other works. This work has been submitted to the IEEE for possible publication. Copyright may be transferred without notice, after which this version may no longer be accessible.
\end{minipage}}


\IEEEtitleabstractindextext{%
\begin{abstract}
Visual scenes are extremely diverse, not only because there are infinite possible combinations of objects and backgrounds but also because the observations of the same scene may vary greatly with the change of viewpoints. When observing a multi-object visual scene from multiple viewpoints, humans can perceive the scene compositionally from each viewpoint while achieving the so-called ``object constancy'' across different viewpoints, even though the exact viewpoints are untold. This ability is essential for humans to identify the same object while moving and to learn from vision efficiently. It is intriguing to design models that have a similar ability. In this paper, we consider a novel problem of learning compositional scene representations from multiple unspecified (i.e., unknown and unrelated) viewpoints without using any supervision and propose a deep generative model which separates latent representations into a viewpoint-independent part and a viewpoint-dependent part to solve this problem. During the inference, latent representations are randomly initialized and iteratively updated by integrating the information in different viewpoints with neural networks. Experiments on several specifically designed synthetic datasets have shown that the proposed method can effectively learn from multiple unspecified viewpoints.
\end{abstract}

}

\maketitle
\def\thefootnote{*}\footnotetext{Equal contribution}\def\thefootnote{\arabic{footnote}}

\IEEEdisplaynontitleabstractindextext

%
\IEEEpeerreviewmaketitle

\IEEEraisesectionheading{\section{Introduction}\label{sec:introduction}}

%
%
%
%


\IEEEPARstart{V}{ision} is an important way for humans to acquire knowledge about the world. Due to the diverse combinations of objects and backgrounds that constitute visual scenes, it is hard to model the whole scene directly. In the process of learning from the world, humans can develop the concept of objects \cite{Johnson2010How} and are thus capable of perceiving visual scenes compositionally. This type of learning is more efficient than perceiving the entire scene as a single entity \cite{Fodor1988Connectionism}. Compositionality is one of the fundamental ingredients for building artificial intelligence systems that learn efficiently and effectively like humans \cite{Lake2017Building}. To better capture the combinational property of visual scenes, instead of learning a single representation for the entire scene, it is desirable to build compositional scene representation models which learn \emph{object-centric representations} (i.e., learn separate representations for different objects and backgrounds).

\begin{figure}[t]
	\centering
	\includegraphics[width=0.95\columnwidth]{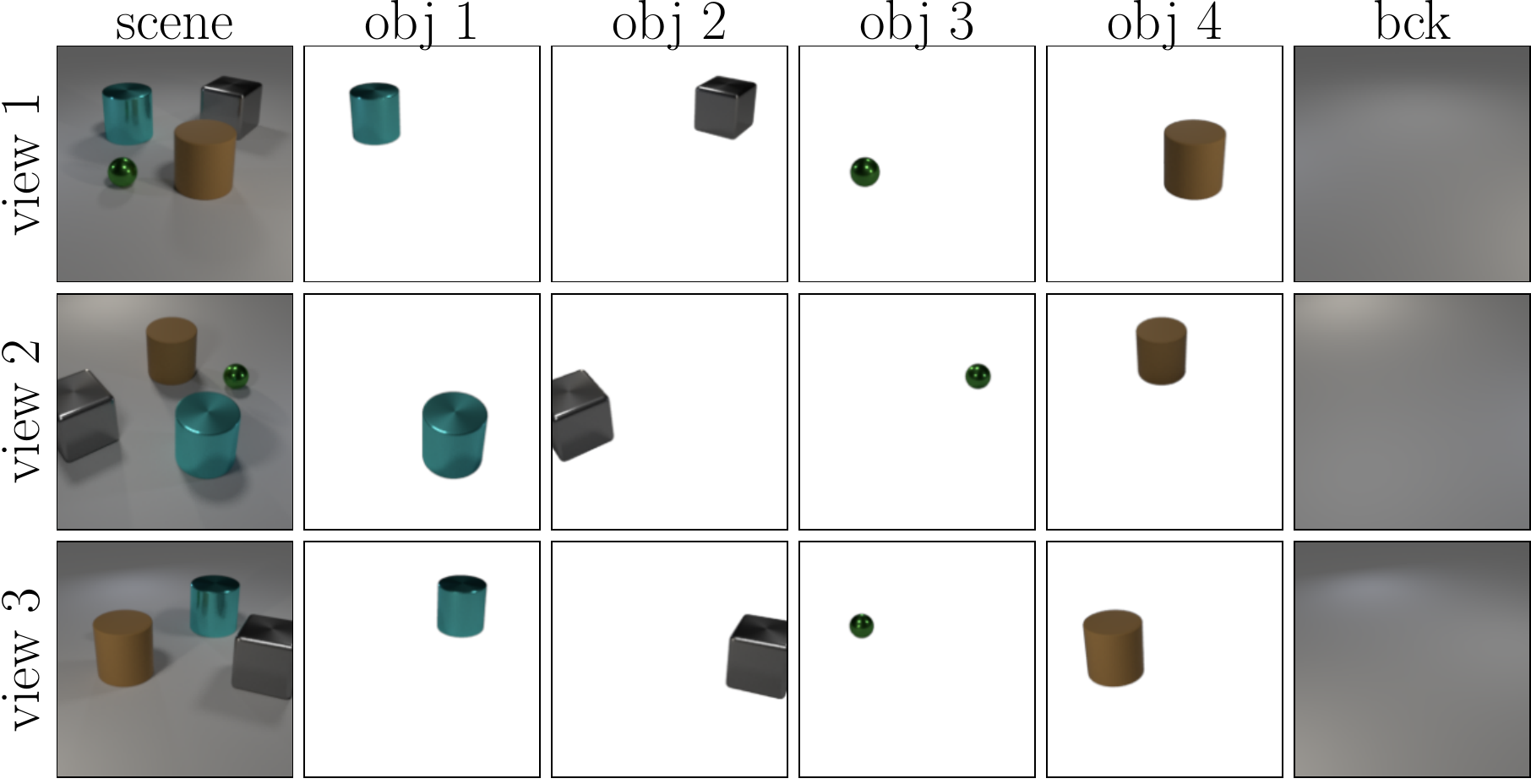}
	\caption{Humans can perceive visual scenes compositionally while maintaining object constancy across different viewpoints (the indexes of objects are arbitrarily chosen).}
	\label{fig:intro}
\end{figure}

In addition, humans can achieve the so-called ``object constancy'' in visual perception, i.e., recognizing the same object from different viewpoints \cite{Turnbull1997Neuropsychology}, possibly because of the mechanisms such as performing mental rotation \cite{Shepard1971Mental} or representing objects independently of viewpoint \cite{Marr1982Vision}. When observing a multi-object scene from multiple viewpoints, humans can separate different objects and identify the same one from different viewpoints. As shown in Figure \ref{fig:intro}, given three images of the same visual scene observed from different viewpoints (column 1), humans are capable of decomposing each image into \emph{complete} objects (columns 2-5) and background (column 6) that are \emph{consistent} across viewpoints, even though the viewpoints are \emph{unknown} and \emph{unrelated}, the poses of the same object may be significantly \emph{different} across viewpoints, and some objects may be partially (object 2 in viewpoint 1) or even completely (object 3 in viewpoint 3) \emph{occluded}. Observing visual scenes from multiple viewpoints gives humans a better understanding of the scenes, and it is intriguing to design compositional scene representation methods that can achieve object constancy and thus effectively learn from multiple viewpoints like humans.

In recent years, a variety of deep generative models have been proposed to learn compositional representations without object-level supervision. Most methods, such as AIR \cite{Eslami2016Attend}, N-EM \cite{Greff2017Neural}, MONet \cite{Burgess2019MONet}, IODINE \cite{Greff2019Multi}, and Slot Attention \cite{Locatello2020Object}, however, only learn from a \emph{single} viewpoint. Only a few methods, including MulMON \cite{Li2020Learning}, DyMON \cite{Nanbo2021Object}, ROOTS \cite{Chen2021ROOTS}, and SIMONe \cite{Kabra2021SIMONe}, have considered the problem of learning from multiple viewpoints. Among these methods, MulMON, DyMON, and ROOTS assume that the viewpoint annotations (under a certain global coordinate system) are given and aim to learn viewpoint-independent object-centric representations \emph{conditioned on} these annotations. Viewpoint annotations play fundamental roles in the initialization and updates of object-centric representations (for MulMON and DyMON) or the computations of perspective projections (for ROOTS). Although SIMONe does not require viewpoint annotations, it assumes that viewpoints of the same visual scene have temporal relationships and utilizes the frame indexes of viewpoints to assist the inference of compositional scene representations. As for the novel problem of learning compositional scene representations from multiple unspecified (i.e., unknown and unrelated) viewpoints \emph{without} any supervision, existing methods are \emph{not} directly applicable.

The problem setting considered in this paper is very challenging, as the object-centric representations that are shared across viewpoints and the viewpoint representations that are shared across objects and backgrounds both need to be learned. More specifically, there are two major reasons. \emph{Firstly}, object constancy needs to be achieved \emph{without the guidance} of viewpoints, which are the only variables among multiple images of the same visual scene and can be exploited to reduce the difficulty of learning the common factors, i.e., object-centric representations. \emph{Secondly}, the representations of images need to be disentangled into object-centric representations and viewpoint representations, even though there are \emph{infinitely many} possible solutions, e.g., due to the change of global coordinate system.

In this paper, we propose a deep generative model called \textbf{O}bject-\textbf{C}entric \textbf{L}earning with \textbf{O}bject \textbf{C}onstancy (OCLOC) to learn compositional representations of visual scenes observed from multiple viewpoints \emph{without any supervision} (including viewpoint annotations) under the assumptions that 1) objects are \emph{static}; and 2) different visual scenes may be observed from \emph{different} sets of viewpoints that are both \emph{unknown} and \emph{unrelated}. The proposed method models viewpoint-independent attributes of objects/backgrounds (e.g., 3D shapes and appearances in the global coordinate system) and viewpoints with separate latent variables. To infer latent variables, OCLOC adopts an amortized variational inference method that iteratively updates the parameters of approximated posteriors by integrating information from different viewpoints with inference neural networks.

To the best of the authors' knowledge, no existing object-centric learning method can learn from multiple unspecified (i.e., unknown and unrelated) viewpoints without viewpoint annotations. Therefore, the proposed OCLOC cannot be directly compared with existing ones in the considered problem setting. Experiments on several specifically designed synthetic datasets have shown that OCLOC can effectively learn from multiple unspecific viewpoints without supervision, i.e., it \emph{competes with} or \emph{slightly outperforms} state-of-the-art methods that either use viewpoint annotations in the learning or assume relationships among viewpoints. The preliminary version of this paper has been published as \cite{Yuan2022Unsupervised}. Compared with the preliminary version, the method proposed in this paper explicitly considers the shadows of objects in the modeling of visual scenes, and the experimental results in this paper are more extensive.

\section{Related Work}

Object-centric representations are compositional scene representations that treat objects or backgrounds as the basic entities of the visual scene and represent different basic entities separately. In recent years, various methods have been proposed to learn object-centric representations without any supervision or only using scene-level annotations. Based on whether learning from multiple viewpoints and whether considering the movements of objects, these methods can be roughly divided into four categories.

\textbf{Single-Viewpoint Static Scenes:}
CST-VAE \cite{Huang2016Efficient}, AIR \cite{Eslami2016Attend}, and MONet \cite{Burgess2019MONet} extract the representation of each object sequentially based on the attention mechanism. GMIOO \cite{Yuan2019Generative} sequentially initializes the representation of each object and iteratively updates the representations, both with attention to objects. SPAIR \cite{Crawford2019Spatially} and SPACE \cite{Lin2020SPACE} generate object proposals with convolutional neural networks and are applicable to large visual scenes containing a relatively large number of objects. N-EM \cite{Greff2017Neural}, LDP \cite{Yuan2019Spatial}, IODINE \cite{Greff2019Multi}, Slot Attention \cite{Locatello2020Object}, and EfficientMORL \cite{Emami2021Efficient} first initialize representations of all the objects and then iteratively update the representations in parallel based on competition among objects. ObSuRF \cite{Stelzner2021Decomposing} represents objects with Neural Radiance Fields (NeRFs). When viewpoints are known, it can extract compositional scene representations from a single viewpoint and render the visual scene from multiple novel viewpoints. GENESIS \cite{Engelcke2020GENESIS} and GNM \cite{Jiang2020Generative} consider the structures of visual scenes in the generative models to generate more coherent samples. ADI \cite{Yuan2021Knowledge} considers the acquisition and utilization of knowledge. These methods provide mechanisms to separate objects and form the foundations of learning object-centric representations with the existence of object motions or from multiple viewpoints.

\textbf{Single-Viewpoint Dynamic Scenes:}
Inspired by the methods proposed for learning from single-viewpoint static scenes, several methods, such as Relational N-EM \cite{Steenkiste2018Relational}, SQAIR \cite{Kosiorek2018Sequential}, R-SQAIR \cite{Stanic2019R}, TBA \cite{He2019Tracking}, SILOT \cite{Crawford2020Exploiting}, SCALOR \cite{Jiang2020SCALOR}, OP3 \cite{Veerapaneni2020Entity}, PROVIDE \cite{Zablotskaia2021PROVIDE}, SAVi \cite{Kipf2022Conditional}, and Gao \& Li \cite{Gao2023Time}, have been proposed for learning from video sequences. The difficulties of this problem setting include modeling object motions and relationships, as well as maintaining the identities of objects even if objects disappear and reappear after full occlusion \cite{Weis2021Benchmarking}. Although these methods can identify the same object across adjacent frames, they cannot be directly applied to the problem setting considered in this paper for two major reasons: 1) multiple viewpoints of the same visual scene are assumed to be unrelated, and the positions of the same object may differ significantly in images observed from different viewpoints; and 2) viewpoints are shared among all the objects in multiple images of the same visual scene, while object motions do not have such a property because different objects may move differently.

\begin{figure}[t]
	\centering
	\includegraphics[width=0.99\columnwidth]{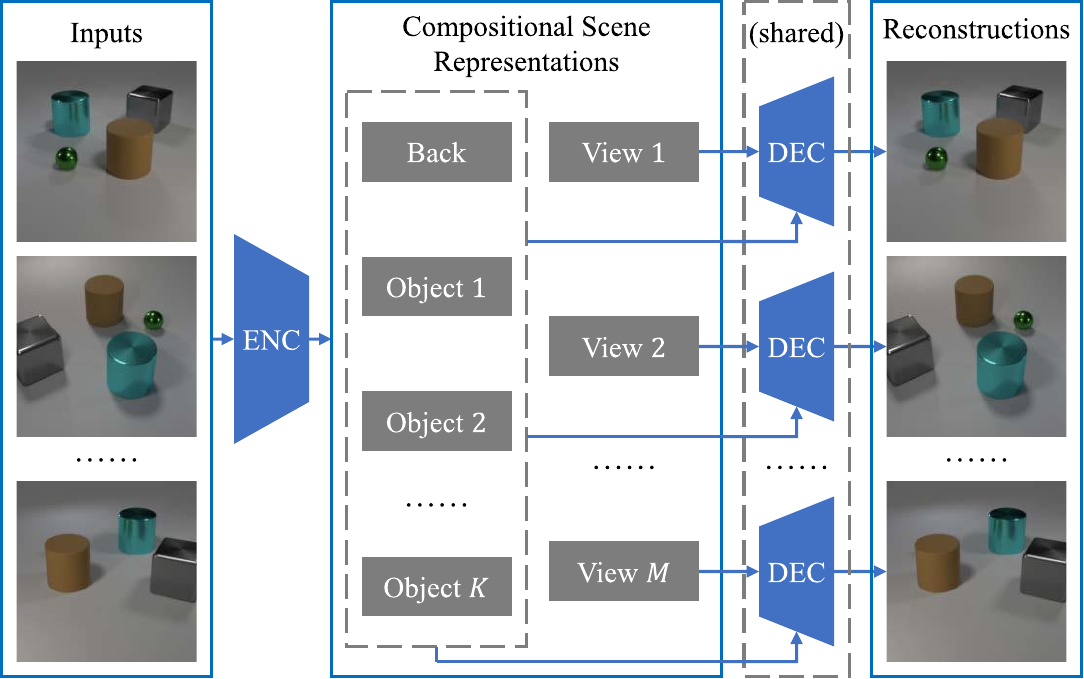}
	\caption{The overall framework of the proposed OCLOC. The main objective of the learning is to reconstruct images of the same visual scene observed from different viewpoints.}
	\label{fig:overview}
\end{figure}

\textbf{Multi-Viewpoint Static Scenes:}
MulMON \cite{Li2020Learning}, ROOTS \cite{Chen2021ROOTS}, and SIMONe \cite{Kabra2021SIMONe} are representative methods proposed for learning compositional representations of static scenes from multiple viewpoints. MulMON extends the iterative amortized inference \cite{Marino2018Iterative} used in IODINE \cite{Greff2019Multi} to sequences of images observed from different viewpoints. Object-centric representations are first initialized based on the first image and its \emph{viewpoint annotation} and then iteratively refined by processing the rest pairs of images and annotations one by one. At each iteration, the previously estimated posteriors of latent variables are used as the current object-wise priors to guide the inference. ROOTS adopts the idea of using grid cells like SPAIR \cite{Crawford2019Spatially} and SPACE \cite{Lin2020SPACE}, and it generates object proposals in a bounded 3D region. The 3D center position of each object proposal is estimated and projected into different images with transformations that are computed based on the \emph{annotated viewpoints}. After extracting crops of images corresponding to each object proposal, a type of GQN \cite{Eslami2018Neural} is applied to infer object-centric representations. SIMONe assumes that both the object-centric representations and viewpoint representations are fully independent in the generative model. Although relationships between viewpoints are not modeled in the generative model, images observed from different viewpoints are assumed to have temporal relationships during the inference. Inspired by Transformer \cite{Vaswani2017Attention}, the encoder of SIMONe first extracts feature maps of images observed from different viewpoints and applies positional embeddings both spatially and temporally, then uses the self-attention mechanism to transform feature maps, and finally obtains viewpoint representations and object-centric representations by spatial and temporal averages, respectively.
Because MulMON and ROOTS heavily rely on viewpoint annotations, and SIMONe exploits the temporal relationships among viewpoints during the inference, they are not well suited for the fully unsupervised scenario where viewpoints are both unknown and unrelated.

\textbf{Multi-Viewpoint Dynamic Scenes:}
Learning compositional representations of dynamic scenes from multiple viewpoints is a challenging problem that has only been considered recently. A representative method proposed for this problem is DyMON \cite{Nanbo2021Object}, which extends MulMON \cite{Li2020Learning} to videos observed from multiple viewpoints. To decouple the influence of viewpoint change and object motion, DyMON makes two assumptions. The first is that the frame rate of the video is very high, and the second is that either viewpoint change or object motion is the main reason for the change of adjacent frames. In the inference of latent variables, DyMON first determines the main reason for the change of adjacent frames and then chooses the frequencies accordingly to update viewpoints and compositional scene representations iteratively. Same as MulMON, DyMON does not learn in the fully unsupervised setting because it assumes that viewpoint annotations are given.

\section{Proposed Method}

The proposed OCLOC assumes that objects in the visual scenes are static, and different visual scenes may be observed from different sets of unknown and unrelated viewpoints. Compositional scene representations are learned mainly by reconstructing images of the same visual scene observed from different viewpoints. As shown in Figure \ref{fig:overview}, compositional representations of visual scenes are divided into a viewpoint-independent part (i.e., object-centric representations) and a viewpoint-dependent part (i.e., viewpoint representations). The viewpoint-independent part characterizes intrinsic attributes of objects and backgrounds, e.g., 3D shapes and appearances in the global coordinate system. The viewpoint-dependent part models the rest attributes that may vary as the viewpoint changes. To extract compositional scene representations from images, OCLOC adopts an amortized variational inference method that iteratively updates parameters of approximated posteriors by integrating information from different viewpoints with inference neural networks (i.e., encoder networks). To reconstruct each image, decoder networks that consider the compositionality of visual scenes are applied to transform object-centric representations and the corresponding viewpoint representation. Parameters of decoder networks are shared across all the viewpoints and all the objects. Therefore, the proposed OCLOC is applicable to visual scenes with different numbers of objects and viewpoints.

\begin{figure}[t]
	\centering
	\includegraphics[width=0.55\columnwidth]{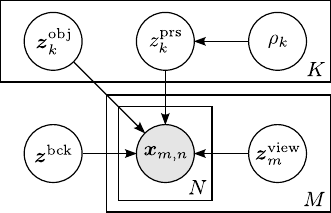}
	\caption{The probabilistic graphical model of visual scene modeling. $K$ is the maximum number of objects that may appear in the visual scene. $N$ is the number of pixels in each visual scene image. $M$ is the number of viewpoints to observe the visual scene. $\boldsymbol{z}_{1:K}^{\text{obj}}$ and $\boldsymbol{z}^{\text{bck}}$ are continuous latent variables that characterize the viewpoint-independent attributes of objects and the background, respectively. $z_k^{\text{prs}}$ with $1 \!\leq\! k \!\leq\! K$ is a binary latent variable that indicates whether the $k$th object is included in the visual scene. This type of latent variables makes it possible to model the varying number of objects in different visual scenes. $\rho_k$ is a continuous latent variable that defines the distribution to generate $z_k^{\text{prs}}$. $\boldsymbol{z}_{m}^{\text{view}}$ with $1 \!\leq\! m \!\leq\! M$ is a continuous latent variable that determine the $m$th viewpoint to observe the visual scene. $\boldsymbol{x}_{1:M,1:N}$ represents the observed visual scene image. Neural networks are applied to compute parameters of the likelihood function $p(\boldsymbol{x}_{m,n}|\boldsymbol{z}_{m}^{\text{view}}, \boldsymbol{z}_{1:K}^{\text{obj}}, \boldsymbol{z}^{\text{bck}}, \boldsymbol{z}_{1:K}^{\text{prs}})$.}
	\label{fig:bn}
\end{figure}

\subsection{Modeling of Visual Scenes}

Visual scenes are assumed to be independent and identically distributed, and they are modeled in a generative way. For simplicity, the index of the visual scene is omitted in the generative model, and the procedure to generate images of a single visual scene is described. Let $M$ denote the number of images observed from different viewpoints (\emph{may vary} in different visual scenes), $N$ and $C$ denote the respective numbers of pixels and channels in each image, and $K$ denote the \emph{maximum} number of objects that may appear in the visual scene. The image of the $m$th viewpoint $\boldsymbol{x}_{m} \in \mathbb{R}^{N \times C}$ is assumed to be generated via a pixel-wise mixture of $K + 1$ layers, with $K$ layers ($1 \!\leq\! k \!\leq\! K$) describing the objects and $1$ layer ($k \!=\! 0$) describing the background. The pixel-wise weights $\boldsymbol{\pi}_{m,0:K} \!\in\! [0, 1]^{(K + 1) \times N}$ and the images of layers $\boldsymbol{a}_{m,0:K} \!\in\! \mathbb{R}^{(K + 1) \times N \times C}$ are computed based on latent variables $\boldsymbol{z}_{1:M}^{\text{view}}$ (contains the information of $M$ viewpoints), $\boldsymbol{z}_{1:K}^{\text{obj}}$ (characterizes the viewpoint-independent attributes of objects), $\boldsymbol{z}^{\text{bck}}$ (characterizes the viewpoint-independent attributes of the background), and $\boldsymbol{z}_{1:K}^{\text{prs}}$ (determines the number of objects in the visual scene). The probabilistic graphical model of visual scene modeling is shown in Figure \ref{fig:bn}. In the following, we first express the generative model in mathematical form and then describe the latent variables and the likelihood function in detail.

\subsubsection{Generative Model}

\label{sec:generative_model}

The mathematical expressions of the generative model are
\begin{align*}
	\boldsymbol{z}_{m}^{\text{view}} & \sim \Normal\big(\boldsymbol{0}, \boldsymbol{I}\big); \mkern27mu \boldsymbol{z}_{k}^{\text{obj}} \sim \Normal\big(\boldsymbol{0}, \boldsymbol{I}\big); \mkern27mu \boldsymbol{z}^{\text{bck}} \sim \Normal\big(\boldsymbol{0}, \boldsymbol{I}\big) \\
	\rho_{k} & \sim \Beta\big(\alpha / K, 1\big); \qquad\qquad z_{k}^{\text{prs}} \sim \Bernoulli\big(\rho_{k}\big) \\
	s_{m,k,n}^{\text{sdw}} & = z_{k}^{\text{prs}} \sigmoid(f_{\text{slt}}^{\text{sdw}}(\boldsymbol{z}_{m}^{\text{view}}, \boldsymbol{z}_{k}^{\text{obj}})_n) \\
	s_{m,k,n}^{\text{obj}} & = z_{k}^{\text{prs}} (1 - s_{m,k,n}^{\text{sdw}}) \sigmoid(f_{\text{slt}}^{\text{obj}}(\boldsymbol{z}_{m}^{\text{view}}, \boldsymbol{z}_{k}^{\text{obj}})_n) \\
	o_{m,k} & = f_{\text{ord}}(\boldsymbol{z}_{m}^{\text{view}}, \boldsymbol{z}_{k}^{\text{obj}}) \\
	\zeta_{m,k,n} & =
	\begin{dcases}
		\prod\nolimits_{k'=1}^{K}(1 - s_{m,k',n}^{\text{sdw}}), \mkern114mu k = 0 \\
		s_{m,k,n}^{\text{sdw}} \prod\nolimits_{k'\!: o_{m,k'} \!> o_{m,k}}\!\!(1 - s_{m,k',n}^{\text{sdw}}), \mkern12mu 1 \leq k \leq K
	\end{dcases} \\
	\pi_{m,k,n} & =
	\begin{dcases}
		\prod\nolimits_{k'=1}^{K}(1 - s_{m,k',n}^{\text{obj}}), \mkern114mu k = 0 \\
		s_{m,k,n}^{\text{obj}} \prod\nolimits_{k'\!: o_{m,k'} \!> o_{m,k}}\!\!(1 - s_{m,k',n}^{\text{obj}}), \mkern12mu 1 \leq k \leq K
	\end{dcases} \\
	\boldsymbol{b}_{m,k,n} & =
	\begin{dcases}
		f_{\text{bck}}\big(\boldsymbol{z}_{m}^{\text{view}}, \boldsymbol{z}^{\text{bck}}\big)_{n}, \mkern137mu k = 0 \\
		\boldsymbol{b}_{m,0,n} \sigmoid(f_{\text{apc}}^{\text{sdw}}(\boldsymbol{z}_{m}^{\text{view}}, \boldsymbol{z}_{k}^{\text{obj}})_n), \mkern11mu 1 \leq k \leq K
	\end{dcases} \\
	\boldsymbol{a}_{m,k,n} & =
	\begin{dcases}
		\sum\nolimits_{k'=0}^{K}{\zeta_{m,k',n} \, \boldsymbol{b}_{m,k',n}}, \mkern98mu k = 0 \\
		f_{\text{apc}}^{\text{obj}}\big(\boldsymbol{z}_{m}^{\text{view}}, \boldsymbol{z}_{k}^{\text{obj}}\big)_{n}, \mkern139mu 1 \leq k \leq K
	\end{dcases} \\
	\boldsymbol{x}_{m,n} & \sim \sum\nolimits_{k=0}^{K}{\pi_{m,k,n} \, \mathcal{N}\Big(\boldsymbol{a}_{m,k,n}, \, \sigma_{\text{x}}^2 \boldsymbol{I}\Big)}
\end{align*}
In the above expressions, some of the ranges of indexes, i.e., $1 \!\leq\! m \!\leq\! M$, $1 \!\leq\! n \!\leq\! N$, and $1 \!\leq\! k \!\leq\! K$, are omitted for simplicity. $\alpha$ and $\sigma_{\text{x}}$ are tunable hyperparameters. Let $\boldsymbol{\Omega} = \{$$\boldsymbol{z}^{\text{view}}$, $\boldsymbol{z}^{\text{obj}}$, $\boldsymbol{z}^{\text{bck}}$, $\boldsymbol{\rho}$, $\boldsymbol{z}^{\text{prs}}$$\}$ be the collection of all latent variables. The joint probability of $\boldsymbol{x}$ and $\boldsymbol{\Omega}$ is
\begin{align}
	\label{equ:generative}
	& p(\boldsymbol{x}, \boldsymbol{\Omega}) = p(\boldsymbol{z}^{\text{bck}}) \prod_{k=1}^{K}{\!p(\boldsymbol{z}_{k}^{\text{obj}}) p(\rho_{k}) p(z_{k}^{\text{prs}}|\rho_{k})} \prod_{m=1}^{M}{\!p(\boldsymbol{z}_{m}^{\text{view}})} \nonumber \\
	& \mkern72mu\prod_{m=1}^{M}{\prod_{n=1}^{N}{\!p(\boldsymbol{x}_{m,n}|\boldsymbol{z}_{m}^{\text{view}}, \boldsymbol{z}_{1:K}^{\text{obj}}, \boldsymbol{z}^{\text{bck}}, \boldsymbol{z}_{1:K}^{\text{prs}})}}
\end{align}

\subsubsection{Latent Variables}

According to whether depending on viewpoints, latent variables can be categorized into two parts. Viewpoint-dependent latent variables may vary as the viewpoint changes. These latent variables include $\boldsymbol{z}_{m}^{\text{view}}$ with $1 \!\leq\! m \!\leq\! M$. Viewpoint-independent latent variables are shared across different viewpoints and are introduced in the generative model to achieve object constancy. These variables include $\boldsymbol{z}^{\text{obj}}$, $\boldsymbol{z}^{\text{bck}}$, $\boldsymbol{\rho}$, and $\boldsymbol{z}^{\text{prs}}$.
\begin{itemize}[leftmargin=*]
	\item $\boldsymbol{z}_{m}^{\text{view}}$ determines the viewpoint (in an automatically chosen global coordinate system) of the $m$th image. It is drawn from a standard normal prior distribution.
	\item $\boldsymbol{z}_{1:K}^{\text{obj}}$ and $\boldsymbol{z}^{\text{bck}}$ characterize the viewpoint-independent attributes of objects and the background, respectively. These attributes include the 3D shapes and appearances of objects and the background in an automatically chosen global coordinate system. The priors of both $\boldsymbol{z}_{1:K}^{\text{obj}}$ and $\boldsymbol{z}^{\text{bck}}$ are standard normal distributions.
	\item $\boldsymbol{\rho}_{1:K}$ and $\boldsymbol{z}^{\text{prs}}_{1:K}$ are used to model the number of objects in the visual scene, considering that different visual scenes may contain different numbers of objects. The binary latent variable $\boldsymbol{z}^{\text{prs}}_{k} \!\in\! \{0, 1\}$ indicates whether the $k$th object is included in the visual scene (i.e., the number of objects is $\sum_{k=1}^{K}{\boldsymbol{z}^{\text{prs}}_{k}}$) and is sampled from a Bernoulli distribution with the latent variable $\boldsymbol{\rho}_{k}$ as its parameter. The priors of all the $\boldsymbol{\rho}_{k}$ with $1 \!\leq\! k \!\leq\! K$ are beta distributions parameterized by hyperparameters $\alpha$ and $K$.
\end{itemize}

\subsubsection{Likelihood Function}

\begin{figure}[t]
	\centering
	\includegraphics[width=0.99\columnwidth]{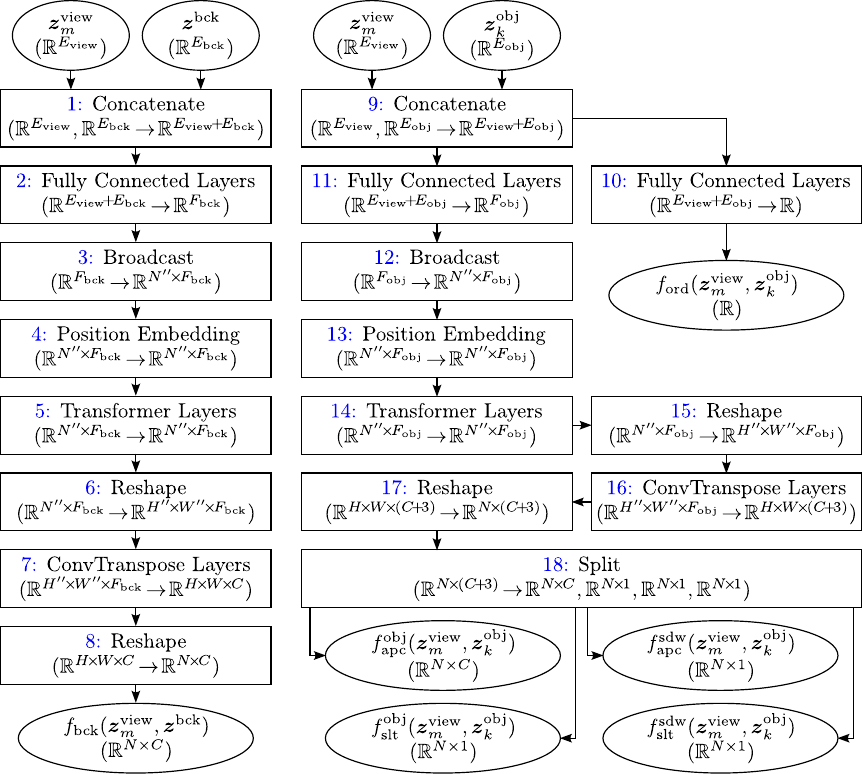}
	\caption{The architecture of decoder networks. The batch dimension is omitted for simplicity.}
	\label{fig:dedoce}
\end{figure}

All the pixels of images $\boldsymbol{x}_{1:M,1:N}$ are assumed to be conditional independent of each other given all the latent variables $\boldsymbol{\Omega}$, and the likelihood function $p(\boldsymbol{x}|\boldsymbol{\Omega})$ is assumed to be factorized as the product of several mixture models. To compute the parameters (i.e., $\boldsymbol{\pi}$ and $\boldsymbol{a}$) of these mixture models, neural networks are applied to transform latent variables. The structure of decoder networks is shown in Figure \ref{fig:dedoce}. The meanings of the variables $\boldsymbol{s}^{\text{sdw}}$, $\boldsymbol{s}^{\text{obj}}$, $\boldsymbol{o}$, $\boldsymbol{\zeta}$, $\boldsymbol{\pi}$, $\boldsymbol{b}$, and $\boldsymbol{a}$ in the transformation are presented below.
\begin{itemize}[leftmargin=*]
	\item $\boldsymbol{s}_{m,1:K,1:N}^{\text{sdw}} \in [0, 1]^{K \times N}$ and $\boldsymbol{s}_{m,1:K,1:N}^{\text{obj}} \in [0, 1]^{K \times N}$ indicate the shadows and complete silhouettes of objects in the image coordinate system determined by the $m$th viewpoint, respectively. They are computed by first applying the sigmoid function to the outputs of neural networks $f_{\text{slt}}^{\text{sdw}}$ and $f_{\text{slt}}^{\text{obj}}$ to restrict the ranges and then multiplying the results with $\boldsymbol{z}_{1:K}^{\text{prs}}$ to ensure that the shadows and silhouettes of objects not in the visual scene are empty.
	\item $\boldsymbol{o}_{m,1:K}$ characterizes the depth ordering of objects in the image observed from the $m$th viewpoint. If multiple objects overlap, the object with the largest value of $o_{m,k}$ occludes the others. It is computed by transforming latent variables $\boldsymbol{z}_{m}^{\text{view}}$ and $\boldsymbol{z}_{1:K}^{\text{obj}}$ with the neural network $f_{\text{ord}}$.
	\item $\boldsymbol{\zeta}_{m,0:K,1:N}$ and $\boldsymbol{\pi}_{m,0:K,1:N}$ indicate the perceived silhouettes of shadows and objects in the $m$th image. These variables satisfy the constraints that $(\forall m, k, n) \, 0 \!\leq\! \zeta_{m,k,n} \!\leq\! 1$, $(\forall m, k, n) \, 0 \!\leq\! \pi_{m,k,n} \!\leq\! 1$, $(\forall m, n) \sum_{k=0}^{K}{\zeta_{m,k,n}} \!=\! 1$, and $(\forall m, n) \sum_{k=0}^{K}{\pi_{m,k,n}} \!=\! 1$. They are computed based on $\boldsymbol{s}_{m,1:K,1:N}^{\text{sdw}}$, $\boldsymbol{s}_{m,1:K,1:N}^{\text{obj}}$, and $\boldsymbol{o}_{m,1:K}$.
	\item $\boldsymbol{b}_{m,0:K,1:N}$ describes the background in the image observed from the $m$th viewpoint without ($k \!=\! 0$) and with ($1 \!\leq\! k \!\leq\! K$) shadows on it. $\boldsymbol{b}_{m,0,1:N}$ is computed by the neural network $f_{\text{bck}}$, whose inputs are the latent variables $\boldsymbol{z}_{m}^{\text{view}}$ and $\boldsymbol{z}^{\text{bck}}$. As for $\boldsymbol{b}_{m,k,1:N}$ ($1 \!\leq\! k \!\leq\! K$), which places the shadow of the $k$th object on the background, it is computed by transforming $\boldsymbol{z}_{m}^{\text{view}}$ and $\boldsymbol{z}_{k}^{\text{obj}}$ with the neural network $f_{\text{apc}}^{\text{sdw}}$, applying the sigmoid function, and multiplying the results with $\boldsymbol{b}_{m,0,1:N}$.
	\item $\boldsymbol{a}_{m,0:K,1:N}$ contains information about the complete appearances of the background ($k \!=\! 0$, with shadows of objects on it) and objects ($1 \!\leq\! k \!\leq\! K$, without shadows) in the $m$th image. $\boldsymbol{a}_{m,0,1:N}$ is computed as the summation of variable $\boldsymbol{b}_{m,0:K,1:N}$ weighted by $\boldsymbol{\zeta}_{m,0:K,1:N}$. $\boldsymbol{a}_{m,k,1:N}$ with $1 \!\leq\! k \!\leq\! K$ is computed by transforming latent variables $\boldsymbol{z}_{m}^{\text{view}}$ and $\boldsymbol{z}_{k}^{\text{obj}}$ with the neural network $f_{\text{apc}}^{\text{obj}}$.
\end{itemize}

\subsection{Inference of Latent Variables}

The exact posterior distribution $p(\boldsymbol{\Omega}|\boldsymbol{x})$ of latent variables is intractable to compute. Therefore, we adopt amortized variational inference, which approximates the complex posterior distribution with a tractable variational distribution $q(\boldsymbol{\Omega}|\boldsymbol{x})$, and apply neural networks to transform $\boldsymbol{x}$ into parameters of the variational distribution.

\subsubsection{Variational Distribution}

The variational distribution $q(\boldsymbol{\Omega}|\boldsymbol{x})$ is factorized as
\begin{align}
	\label{equ:inference}
	& q(\boldsymbol{\Omega}|\boldsymbol{x}) = q(\boldsymbol{z}^{\text{bck}}|\boldsymbol{x}) \prod\nolimits_{k=1}^{K}{q(\boldsymbol{z}_{k}^{\text{obj}}|\boldsymbol{x})} \\
	& \mkern73mu \prod\nolimits_{k=1}^{K}{q(\rho_{k}|\boldsymbol{x}) q(z_{k}^{\text{prs}}|\boldsymbol{x})} \prod\nolimits_{m=1}^{M}{q(\boldsymbol{z}_{m}^{\text{view}}|\boldsymbol{x})}  \nonumber
\end{align}
The choices of terms on the right-hand side of Eq. \eqref{equ:inference} are
\begin{align}
	q(\boldsymbol{z}^{\text{bck}}|\boldsymbol{x}) & = \Normal\big(\boldsymbol{z}^{\text{bck}}; \boldsymbol{\mu}^{\text{bck}}, \diag(\boldsymbol{\sigma}^{\text{bck}})^2\big) \\
	q(\boldsymbol{z}_{k}^{\text{obj}}|\boldsymbol{x}) & = \Normal\big(\boldsymbol{z}_{k}^{\text{obj}}; \boldsymbol{\mu}_{k}^{\text{obj}}, \diag(\boldsymbol{\sigma}_{k}^{\text{obj}})^2\big) \\
	q(\rho_{k}|\boldsymbol{x}) & = \Beta\big(\rho_{k}; \tau_{k,1}, \tau_{k,2}\big) \\
	q(z_{k}^{\text{prs}}|\boldsymbol{x}) & = \Bernoulli\big(z_{k}^{\text{prs}}; \kappa_{k}\big) \\
	q(\boldsymbol{z}_{m}^{\text{view}}|\boldsymbol{x}) & = \Normal\big(\boldsymbol{z}_{m}^{\text{view}}; \boldsymbol{\mu}_{m}^{\text{view}}, \diag(\boldsymbol{\sigma}_{k}^{\text{attr}})^2\big)
\end{align}
In the above expressions, $q(\boldsymbol{z}^{\text{bck}}|\boldsymbol{x})$, $q(\boldsymbol{z}_{k}^{\text{obj}}|\boldsymbol{x})$ and $q(\boldsymbol{z}_{m}^{\text{view}}|\boldsymbol{x})$ are normal distributions with diagonal covariance matrices. $z_{k}^{\text{prs}}$ is assumed to be independent of $\rho_{k}$ given $\boldsymbol{x}$, and $q(\rho_{k}|\boldsymbol{x})$ and $q(z_{k}^{\text{prs}}|\boldsymbol{x})$ are chosen to be a beta distribution and a Bernoulli distribution, respectively. The advantage of this formulation is that the Kullback-Leibler (KL) divergence between $q(\rho_{k}|\boldsymbol{x}) q(z_{k}^{\text{prs}}|\boldsymbol{x})$ and $p(\rho_{k}) p(z_{k}^{\text{prs}}|\rho_{k})$ has a closed-form solution. The parameters $\boldsymbol{\mu}^{\text{bck}}$, $\boldsymbol{\sigma}^{\text{bck}}$, $\boldsymbol{\mu}^{\text{obj}}$, $\boldsymbol{\sigma}^{\text{obj}}$, $\boldsymbol{\tau}$, $\boldsymbol{\kappa}$, $\boldsymbol{\mu}^{\text{view}}$, and $\boldsymbol{\sigma}^{\text{view}}$ of these distributions are estimated by transforming $\boldsymbol{x}$ with inference networks.

\begin{figure}[t]
	\centering
	\includegraphics[width=0.9\columnwidth]{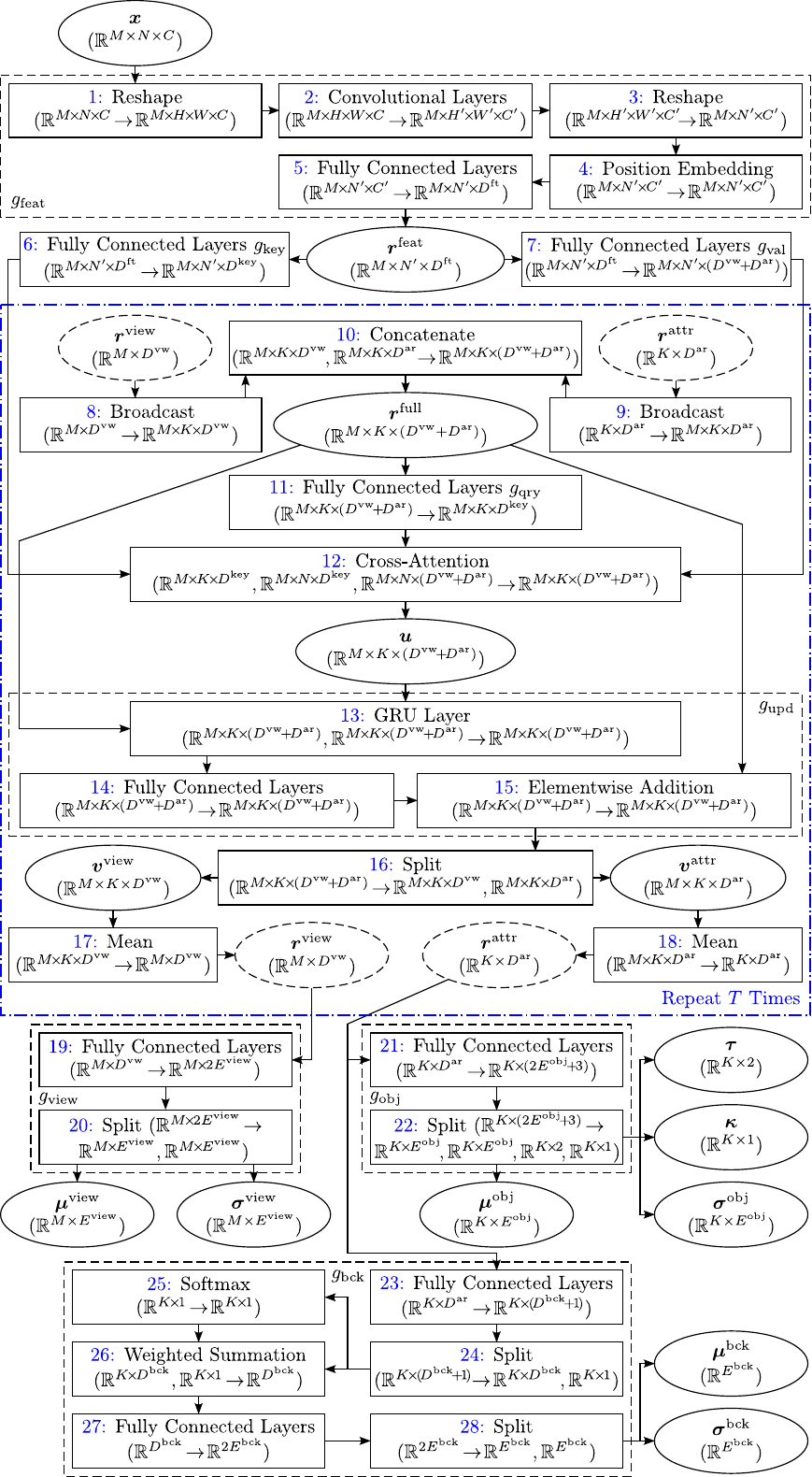}
	\caption{The architecture of encoder networks. The batch dimension is omitted for simplicity. In the dashed box that is repeated $T$ times, variables $\boldsymbol{r}^{\text{view}}$ and $\boldsymbol{r}^{\text{attr}}$ are initialized randomly and updated iteratively.}
	\label{fig:encode}
\end{figure}

\subsubsection{Inference Networks}

As shown in Figure \ref{fig:encode}, the parameters $\boldsymbol{\mu}^{\text{bck}}$, $\boldsymbol{\sigma}^{\text{bck}}$, $\boldsymbol{\mu}^{\text{obj}}$, $\boldsymbol{\sigma}^{\text{obj}}$, $\boldsymbol{\tau}$, $\boldsymbol{\kappa}$, $\boldsymbol{\mu}^{\text{view}}$, and $\boldsymbol{\sigma}^{\text{view}}$ of the variational distribution $q(\boldsymbol{\Omega}|\boldsymbol{x})$ are estimated with neural networks $g_{\text{feat}}$, $g_{\text{key}}$, $g_{\text{val}}$, $g_{\text{qry}}$, $g_{\text{upd}}$, $g_{\text{bck}}$, $g_{\text{obj}}$, $g_{\text{view}}$. Inspired by Slot Attention \cite{Locatello2020Object}, the inference is performed by first randomly initializing parameters of the variational distribution and then iteratively updating these parameters based on cross-attention. The procedure for applying inference networks is presented in Algorithm \ref{alg:infer}, and explanations are given below.
\begin{itemize}[leftmargin=*]
	\item $g_{\text{feat}}$ is the combination of convolution layers, position embedding, and fully connected layers. It transforms the image $\boldsymbol{x}_{m} \!\in\! \mathbb{R}^{N \times C}$ observed from each viewpoint into feature maps $\boldsymbol{r}_{m}^{\text{feat}} \!\in\! \mathbb{R}^{N' \times D_{\text{ft}}}$ that summarize the information of local regions in the image.
	\item $g_{\text{key}}$, $g_{\text{qry}}$, $g_{\text{val}}$, and $g_{\text{upd}}$ are neural networks used to transform feature maps $\boldsymbol{r}^{\text{feat}} \!\in\! \mathbb{R}^{M \times N' \times D_{\text{ft}}}$ into intermediate variables $\boldsymbol{r}^{\text{view}} \!\in\! \mathbb{R}^{M \!\times\! D_{\text{vw}}}$ and $\boldsymbol{r}^{\text{attr}} \!\in\! \mathbb{R}^{K \!\times\! D_{\text{at}}}$ that characterize the parameters of the viewpoint-dependent part ($\boldsymbol{\mu}^{\text{view}}$ and $\boldsymbol{\sigma}^{\text{view}}$) and the viewpoint-independent part ($\boldsymbol{\mu}^{\text{bck}}$, $\boldsymbol{\sigma}^{\text{bck}}$, $\boldsymbol{\mu}^{\text{obj}}$, $\boldsymbol{\sigma}^{\text{obj}}$, $\boldsymbol{\tau}$, and $\boldsymbol{\kappa}$) of the variational distribution $q(\boldsymbol{\Omega}|\boldsymbol{x})$, respectively. Among these networks, $g_{\text{key}}$, $g_{\text{qry}}$, and $g_{\text{val}}$ are fully connected layers, and $g_{\text{upd}}$ is a gated recurrent unit
	(GRU) followed by a residual multilayer perceptron (MLP).
	\item $g_{\text{view}}$, $g_{\text{obj}}$, and $g_{\text{bck}}$ are neural networks that transform intermediate variables into parameters of the variational distribution, i.e., the final outputs of the inference. $g_{\text{obj}}$ and $g_{\text{view}}$ are fully connected layers. $g_{\text{bck}}$ is the combination of fully connected layers. It aggregates the information of background from all the viewpoint-independent slots via weighted summation.
\end{itemize}

\subsection{Learning of Neural Networks}

The neural networks in the generative model, as well as the inference networks (including learnable parameters $\hat{\boldsymbol{\mu}}^{\text{view}}$, $\hat{\boldsymbol{\sigma}}^{\text{view}}$, $\hat{\boldsymbol{\mu}}^{\text{attr}}$, and $\hat{\boldsymbol{\sigma}}^{\text{attr}}$), are jointly learned with the goal of minimizing the negative value of evidence lower bound (ELBO). Detailed expressions of the loss function and the optimization of network parameters are described below.

\subsubsection{Loss Function}

The loss function $\mathcal{L}$ can be decomposed as
\begin{equation}
	\label{equ:loss}
	\mathcal{L} = \!\sum_{m=1}^{M}{\!\sum_{n=1}^{N}{\!\mathcal{L}_{m,n}^{\text{nll}}}} \!+\! \sum_{m=1}^{M}{\!\mathcal{L}_{m}^{\text{view}}} \!+\! \mathcal{L}^{\text{bck}} \!+\! \sum_{k=1}^{K}\!\big(\mathcal{L}_{k}^{\text{obj}} \!+\! \mathcal{L}_{k}^{\rho} \!+\! \mathcal{L}_{k}^{\text{prs}}\big)
\end{equation}
In Eq. \eqref{equ:loss}, the first term is negative log-likelihood, and the rest five terms are Kullback-Leibler (KL) divergences that are computed by $D_{\text{KL}}(q||p) \!=\! \mathbb{E}_{q}[\log{q} - \log{p}]$. Let $\Gamma$ and $\phi$ denote gamma and digamma functions, respectively. Detailed expressions of these terms are
\begin{align}
	\mathcal{L}_{m,n}^{\text{nll}} = \, & - \mathbb{E}_{q(\boldsymbol{\Omega}|\boldsymbol{x})}\big[\log{p(\boldsymbol{x}_{m,n}|\boldsymbol{z}_{m}^{\text{view}}\!, \boldsymbol{z}^{\text{bck}}\!, \boldsymbol{z}_{1:K}^{\text{obj}}\!, \boldsymbol{z}_{1:K}^{\text{prs}})}\big] \\
	= \, & \text{const} - \mathbb{E}_{q(\boldsymbol{\Omega}|\boldsymbol{x})}{\!\Bigg[\!\log\!\bigg(\!\sum_{k=0}^{K}{\!\pi_{m,k,n} \, e^{-\frac{(\boldsymbol{x}_{m,n} - \boldsymbol{a}_{m,k,n})^2}{2 \sigma_{\text{x}}^2}}}\bigg)\!\Bigg]} \nonumber \\
	\mathcal{L}_{m}^{\text{view}} = \, & D_{\text{KL}}\big(q(\boldsymbol{z}_{m}^{\text{view}}|\boldsymbol{x})||p(\boldsymbol{z}_{m}^{\text{view}})\big) \\
	= \, & \frac{1}{2} \sum_{i}{\Big(\mbox{$\mu_{m,i}^{\text{view}}$}^2 + \mbox{$\sigma_{m,i}^{\text{view}}$}^2 - \log{\mbox{$\sigma_{m,i}^{\text{view}}$}^2} - 1\Big)} \nonumber \\
	\mathcal{L}^{\text{bck}} = \, & D_{\text{KL}}\big(q(\boldsymbol{z}^{\text{bck}}|\boldsymbol{x})||p(\boldsymbol{z}^{\text{bck}})\big) \\
	= \, & \frac{1}{2} \sum_{i}{\Big(\mbox{$\mu_{i}^{\text{bck}}$}^2 + \mbox{$\sigma_{i}^{\text{bck}}$}^2 - \log{\mbox{$\sigma_{i}^{\text{bck}}$}^2} - 1\Big)} \nonumber \\
	\mathcal{L}_{k}^{\text{obj}} = \, & D_{\text{KL}}\big(q(\boldsymbol{z}_{k}^{\text{obj}}|\boldsymbol{x})||p(\boldsymbol{z}_{k}^{\text{obj}})\big) \\
	= \, & \frac{1}{2} \sum_{i}{\Big(\mbox{$\mu_{k,i}^{\text{obj}}$}^2 + \mbox{$\sigma_{k,i}^{\text{obj}}$}^2 - \log{\mbox{$\sigma_{k,i}^{\text{obj}}$}^2} - 1\Big)} \nonumber \\
	\mathcal{L}_{k}^{\rho} = \, & D_{\text{KL}}\big(q(\rho_{k}|\boldsymbol{x})||p(\rho_{k})\big) \\
	= \, & \log{\frac{\Gamma(\tau_{k,1} + \tau_{k,2})}{\Gamma(\tau_{k,1}) \Gamma(\tau_{k,2})}} - \log{\frac{\alpha}{K}} \nonumber \\
	& + \Big(\tau_{k,1} - \frac{\alpha}{K}\Big) \psi(\tau_{k,1}) + (\tau_{k,2} - 1) \psi(\tau_{k,2}) \nonumber \\
	& - \Big(\tau_{k,1} + \tau_{k,2} - \frac{\alpha}{K} - 1\Big) \psi(\tau_{k,1} + \tau_{k,2}) \nonumber \\
	\mathcal{L}_{k}^{\text{prs}} = \, & \mathbb{E}_{q(\rho_{k}|\boldsymbol{x})}\big[D_{\text{KL}}\big(q(z_{k}^{\text{prs}}|\boldsymbol{x})||p(z_{k}^{\text{prs}}|\rho_{k})\big)\big] \\
	= \, & \psi(\tau_{k,1} + \tau_{k,2}) + \kappa_{k} \big(\!\log(\kappa_{k}) - \psi(\tau_{k,1})\big) \nonumber \\
	& + (1 - \kappa_{k}) \big(\!\log(1 - \kappa_{k}) - \psi(\tau_{k,2})\big) \nonumber
\end{align}

\begin{algorithm}[tb]
\caption{Estimation of $q(\boldsymbol{\Omega}|\boldsymbol{x})$ with inference networks}
\label{alg:infer}
\textbf{Input}: Images of $M$ viewpoints $\boldsymbol{x}_{1:M}$ \\
\textbf{Output}: Parameters of $q(\boldsymbol{\Omega}|\boldsymbol{x})$
\begin{algorithmic}[1] 
	\STATE // Extract features and initialize intermediate variables
	\STATE $\boldsymbol{r}_{m}^{\text{feat}} \gets g_{\text{feat}}(\boldsymbol{x}_m), \quad \forall \, 1 \!\leq\! m \!\leq\! M$
	\STATE $\boldsymbol{r}_{m}^{\text{view}} \sim \mathcal{N}(\hat{\boldsymbol{\mu}}^{\text{view}}, \diag(\hat{\boldsymbol{\sigma}}^{\text{view}})), \quad \forall \, 1 \!\leq\! m \!\leq\! M$
	\STATE $\boldsymbol{r}_{k}^{\text{attr}} \sim \mathcal{N}(\hat{\boldsymbol{\mu}}^{\text{attr}}, \diag(\hat{\boldsymbol{\sigma}}^{\text{attr}})), \quad \forall \, 1 \!\leq\! k \!\leq\! K$
	\STATE // Update intermediate variables $\boldsymbol{r}_{1:M}^{\text{view}}$ and $\boldsymbol{r}_{1:K}^{\text{attr}}$
	\FOR[$\forall \, 1 \!\leq\! m \!\leq\! M, 1 \!\leq\! k \!\leq\! K$ in the loop]{$t \gets 1$ to $T'$}
	\STATE $\boldsymbol{r}_{m,k}^{\text{full}} \!\gets [\boldsymbol{r}_{m}^{\text{view}}, \boldsymbol{r}_{k}^{\text{attr}}]$
	\STATE $\boldsymbol{w}_{m,k} \!\gets \softmax_K\!\big(g_{\text{key}}(\boldsymbol{r}_{m}^{\text{feat}}) g_{\text{qry}}(\boldsymbol{r}_{m,1:K}^{\text{full}}) / \!\sqrt{\!D_{\text{key}}}\big)$
	\STATE $\boldsymbol{u}_{m,k} \!\gets \sum_{N'}{\softmax_{N'}(\log{\boldsymbol{w}_{m,k}}) \, g_{\text{val}}(\boldsymbol{r}_{m}^{\text{feat}})}$
	\STATE $[\boldsymbol{v}_{1:M,1:K}^{\text{view}}, \boldsymbol{v}_{1:M,1:K}^{\text{attr}}] \gets g_{\text{upd}}(\boldsymbol{r}_{1:M,1:K}^{\text{full}}, \boldsymbol{u}_{1:M,1:K})$
	\STATE $\boldsymbol{r}_{m}^{\text{view}} \gets \mean_{K}(\boldsymbol{v}_{m,1:K}^{\text{view}})$
	\STATE $\boldsymbol{r}_{k}^{\text{attr}} \gets \mean_{M}(\boldsymbol{v}_{1:M,k}^{\text{attr}})$
	\ENDFOR
	\STATE // Convert $\boldsymbol{r}_{1:M}^{\text{view}}$ and $\boldsymbol{r}_{1:K}^{\text{attr}}$ to parameters of $q(\boldsymbol{\Omega}|\boldsymbol{x})$
	\STATE $\boldsymbol{\mu}^{\text{bck}}, \boldsymbol{\sigma}^{\text{bck}} \gets g_{\text{bck}}(\boldsymbol{r}_{1:K}^{\text{attr}})$
	\STATE $\boldsymbol{\mu}_{k}^{\text{obj}}, \boldsymbol{\sigma}_{k}^{\text{obj}}, \boldsymbol{\tau}_{k}, \kappa_{k} \gets g_{\text{obj}}(\boldsymbol{r}_{k}^{\text{attr}}), \quad \forall \, 1 \!\leq\! k \!\leq\! K$
	\STATE $\boldsymbol{\mu}_{m}^{\text{view}}, \boldsymbol{\sigma}_{m}^{\text{view}} \gets g_{\text{view}}(\boldsymbol{r}_{m}^{\text{view}}), \quad \forall \, 1 \!\leq\! m \!\leq\! M$
	\STATE \textbf{return} $\boldsymbol{\mu}^{\text{bck}}, \boldsymbol{\sigma}^{\text{bck}}, \boldsymbol{\mu}_{1:K}^{\text{obj}}, \boldsymbol{\sigma}_{1:K}^{\text{obj}}, \boldsymbol{\tau}_{1:K}, \boldsymbol{\kappa}_{1:K}, \boldsymbol{\mu}_{1:M}^{\text{view}}, \boldsymbol{\sigma}_{1:M}^{\text{view}}$
\end{algorithmic}
\end{algorithm}

\subsubsection{Optimization of Network Parameters}

The loss function described in Eq. \eqref{equ:loss} is optimized using the gradient-based method. All the KL divergences have closed-form solutions, and the gradients of these terms can be easily computed. The negative log-likelihood cannot be computed analytically, and the gradients of this term are approximated by sampling latent variables $\boldsymbol{z}^{\text{view}}$, $\boldsymbol{z}^{\text{bck}}$, $\boldsymbol{z}^{\text{obj}}$, and $\boldsymbol{z}^{\text{prs}}$ from the variational distribution $q(\boldsymbol{\Omega}|\boldsymbol{x})$. To reduce the variances of gradients, the continuous variables $\boldsymbol{z}^{\text{view}}$ and $\boldsymbol{z}^{\text{attr}}$ are sampled using the reparameterization trick \cite{Salimans2013Fixed,Kingma2014Auto}, and the discrete variables $\boldsymbol{z}^{\text{prs}}$ and $\boldsymbol{z}^{\text{shp}}$ are approximated using a continuous relaxation \cite{Maddison2017Concrete,Jang2017Categorical}. Because the relative ordering instead of the value of the variable $o_{m,k}$ is used in the computation of the loss function, gradients cannot be backpropagated through this type of variable. To solve this problem, the straight-through estimator is applied. In the forward pass, variables $\zeta_{m,k,n}$ and $\pi_{m,k,n}$ are computed as described in Section \ref{sec:generative_model}. In the backward pass, the gradients are backpropagated as if these variables are computed using the following expressions.
\begin{align*}
	\zeta_{m,k,n} & =
	\begin{dcases}
		\prod\nolimits_{k'=1}^{K}(1 - s_{m,k',n}^{\text{sdw}}), \mkern117mu k = 0 \\
		\frac{(1 - \zeta_{m,0,n}) \, s_{m,k,n}^{\text{sdw}} \, \exp(o_{m,k})}{\sum_{k'=1}^{K}{s_{m,k',n}^{\text{sdw}} \, \exp(o_{m,k'})}}, \mkern37mu \text{otherwise}
	\end{dcases} \\
	\pi_{m,k,n} & =
	\begin{dcases}
		\prod\nolimits_{k'=1}^{K}(1 - s_{m,k',n}^{\text{obj}}), \mkern117mu k = 0 \\
		\frac{(1 - \pi_{m,0,n}) \, s_{m,k,n}^{\text{obj}} \, \exp(o_{m,k})}{\sum_{k'=1}^{K}{s_{m,k',n}^{\text{obj}} \, \exp(o_{m,k'})}}, \mkern37mu \text{otherwise}
	\end{dcases}
\end{align*}

\section{Experiments}

In this section, we aim to verify that the proposed method\footnote{Code is available at \url{https://git.io/JDnne}.}:
\begin{itemize}[leftmargin=*]
	\item can learn compositional scene representations of static scenes from multiple unspecified (unknown and unrelated) viewpoints \emph{without any supervision}, which have \emph{not} been considered by existing methods;
	\item competes with a state-of-the-art that uses \emph{viewpoint annotations} in the learning of compositional scene representations from multiple viewpoints, even though viewpoint annotations are \emph{not} utilized by the proposed method.
	\item outperforms a state-of-the-art proposed for learning from multiple ordered viewpoints of static scenes (i.e., it is assumed that viewpoints have temporal relationships and adjacent viewpoints do not differ too much) under the circumstance that the ordering of viewpoints is unknown and viewpoints may differ significantly;
	\item outperforms a state-of-the-art proposed for learning from videos (i.e., it is assumed that object motions may exist and adjacent video frames do not differ too much) in the considered problem setting (i.e., observations of static visual scenes from unordered viewpoints are treated as video sequences).
\end{itemize}
In the following, we will first describe the datasets, compared methods, and evaluation metrics that are used in the experiments, then present experimental results.

\subsection{Datasets}

To evaluate the performance of multi-viewpoint compositional scene representation learning methods, four multi-viewpoint datasets (referred to as CLEVR, SHOP, GSO, and ShapeNet, respectively) are constructed based on the CLEVR dataset \cite{Johnson2017CLEVR}, the SHOP-VRB dataset \cite{Nazarczuk2020SHOP}, the combination of GSO \cite{Downs2022Google} and HDRI-Haven datasets, and the combination of ShapeNet \cite{Chang2015ShapeNet} and HDRI-Haven datasets. The configurations of these datasets are shown in Table \ref{tab:data_multi}. All the datasets are generated based on the official code provided by \cite{Johnson2017CLEVR}, \cite{Nazarczuk2020SHOP}, and \cite{Greff2022Kubric}. Images in the CLEVR and SHOP datasets are generated with size $214 \!\times\! 160$ and cropped to size $128 \!\times\! 128$ at locations $19$ (up), $147$ (down), $43$ (left), and $171$ (right). Images in the GSO and ShapeNet datasets are generated with the default size $128 \!\times\! 128$.

\begin{table}[ht]
	\renewcommand*{\arraystretch}{1.5}
	\centering
	\caption{Configurations of datasets. Row 1: names of datasets. Row 2: splits of datasets. Row 3: the number of visual scenes in each split. Row 4: the ranges to sample the number of objects per scene. Row 5: the number of viewpoints to observe each visual scene. Row 6: the height and width of each image. Rows 7-9: the ranges to sample viewpoints.}
	\label{tab:data_multi}
	\begin{small}
		\addtolength{\tabcolsep}{-5pt}
		\begin{tabular}{|c|C{0.32in}|C{0.32in}|C{0.32in}|C{0.32in}|C{0.32in}|C{0.32in}|C{0.32in}|C{0.32in}|}
			\hline
			Dataset &               \multicolumn{4}{c|}{CLEVR / SHOP}                &       \multicolumn{4}{c|}{GSO / ShapeNet}       \\ \hline
			         Split           &     Train      &     Valid      &     Test 1     &     Test 2      &     Train      &     Valid      &     Test 1     &     Test 2      \\ \hline
			         Scenes          &      5000      &      100       &      100       &       100       &      5000      &      100       &      100       &       100       \\ \hline
			        Objects          & 3 $\!\sim\!$ 6 & 3 $\!\sim\!$ 6 & 3 $\!\sim\!$ 6 & 7 $\!\sim\!$ 10 & 3 $\!\sim\!$ 6 & 3 $\!\sim\!$ 6 & 3 $\!\sim\!$ 6 & 7 $\!\sim\!$ 10 \\ \hline
			       Viewpoints        &                                                         \multicolumn{4}{c|}{60} & \multicolumn{4}{c|}{12}                                                         \\ \hline
			       Image Size        &                                                  \multicolumn{8}{c|}{128 $\times$ 128}                                                  \\ \hline
			        Azimuth          &                                                    \multicolumn{8}{c|}{$[0, 2\pi]$}                                                     \\[-2pt]
			       Elevation         &                                                \multicolumn{8}{c|}{$[0.15\pi, 0.3\pi]$}                                                 \\[-2pt]
			        Distance         &                                                    \multicolumn{8}{c|}{$[10.5, 12]$}                                                    \\ \hline
		\end{tabular}
	\end{small}
\end{table}

\subsection{Compared Methods}

It is worth noting that the proposed OCLOC \emph{cannot} be directly compared with existing methods in the novel problem setting considered in this paper. To verify the effectiveness of OCLOC, three methods that are originally proposed for problem settings different from the considered one are compared with:
\begin{itemize}[leftmargin=*]
	\item MulMON \cite{Li2020Learning}, a method proposed for learning compositional scene representations from multiple \emph{known} viewpoints of static scenes. It solves a \emph{simpler} problem by using viewpoint annotations in both training and testing.
	\item SIMONe \cite{Kabra2021SIMONe}, a method proposed for learning from multiple unknown viewpoints under the assumption that viewpoints have temporal relationships. When trained and tested in the considered problem setting, the ordering of viewpoints is random. Therefore, the temporal relationships provided to this method are wrong in most cases.
	\item SAVi \cite{Kipf2022Conditional}, a method proposed for learning object-centric representations from videos. This method can be applied to the considered problem setting by treating each viewpoint as a video frame. Since the assumption that adjacent frames do not differ too much does not hold for unordered viewpoints, SAVi may not be well suited for the considered problem setting. To verify this, SAVi is also trained and tested in a different setting where viewpoints are ordered and adjacent viewpoints are not very different.
\end{itemize}
To verify the effectiveness of shadow modeling in the proposed method, an ablation method that does not explicitly consider shadows in the modeling of visual scenes is also compared with. This ablation method differs from OCLOC only in the generative model. The variables $\boldsymbol{b}_{1:M,1:K,1:N}$ is not computed, and the computation of $\boldsymbol{s}_{1:M,0:K,1:N}^{\text{sdw}}$ is replaced with $(\forall m, k, n) \, \boldsymbol{s}_{m,k,n}^{\text{sdw}} = 0$.

\subsection{Evaluation Metrics}

\begin{table*}[ht]
	\centering
	\caption{Comparison of multi-viewpoint learning on the Test 1 splits. All methods are trained with $M \!\in\! [1, 8]$ and $K \!=\! 7$ and tested with $M \!=\! 8$ and $K \!=\! 7$. The reported scores are averaged on datasets with similar properties. The top 2 are underlined, with the best in bold and the second best in italics.}
	\label{tab:multi_8_avg_test_1}
	\begin{small}
		\addtolength{\tabcolsep}{-1pt}
		\begin{tabular}{c|c|C{0.5in}C{0.5in}C{0.5in}C{0.5in}C{0.5in}C{0.5in}C{0.5in}C{0.5in}}
			\toprule
			Dataset          &  Method  &          ARI-A          &          AMI-A          &          ARI-O          &          AMI-O          &           IoU           &           F1            &           OCA           &           OOA           \\ \midrule
			\multirow{6}{*}{CLEVR \& SHOP} & SAVi (video) & 0.060 & 0.252 & 0.715 & 0.796 & N/A & N/A & 0.000 & N/A \\
			& SAVi & 0.003 & 0.021 & 0.046 & 0.068 & N/A & N/A & 0.000 & N/A \\
			& SIMONe & 0.296 & 0.434 & 0.667 & 0.708 & N/A & N/A & 0.000 & N/A \\
			& MulMON & \underline{\textit{0.473}} & \underline{\textit{0.494}} & 0.782 & 0.794 & N/A & N/A & \underline{\textbf{0.227}} & N/A \\
			& Ablation & 0.395 & 0.454 & \underline{\textbf{0.903}} & \underline{\textbf{0.891}} & \underline{\textit{0.444}} & \underline{\textit{0.571}} & 0.195 & \underline{\textbf{0.916}} \\
			& OCLOC & \underline{\textbf{0.734}} & \underline{\textbf{0.650}} & \underline{\textit{0.859}} & \underline{\textit{0.881}} & \underline{\textbf{0.606}} & \underline{\textbf{0.722}} & \underline{\textit{0.205}} & \underline{\textit{0.890}} \\
			\midrule
			\multirow{6}{*}{GSO \& ShapeNet} & SAVi (video) & 0.014 & 0.075 & 0.155 & 0.229 & N/A & N/A & 0.000 & N/A \\
			& SAVi & 0.005 & 0.023 & 0.041 & 0.069 & N/A & N/A & 0.000 & N/A \\
			& SIMONe & 0.404 & 0.390 & 0.327 & 0.432 & N/A & N/A & 0.000 & N/A \\
			& MulMON & 0.220 & 0.200 & 0.225 & 0.274 & N/A & N/A & \underline{\textit{0.025}} & N/A \\
			& Ablation & \underline{\textit{0.429}} & \underline{\textit{0.469}} & \underline{\textit{0.884}} & \underline{\textit{0.843}} & \underline{\textit{0.514}} & \underline{\textit{0.668}} & 0.000 & \underline{\textit{0.953}} \\
			& OCLOC & \underline{\textbf{0.831}} & \underline{\textbf{0.738}} & \underline{\textbf{0.934}} & \underline{\textbf{0.911}} & \underline{\textbf{0.707}} & \underline{\textbf{0.817}} & \underline{\textbf{0.715}} & \underline{\textbf{0.965}} \\
			\bottomrule
		\end{tabular}
	\end{small}
\end{table*}
\begin{table*}[ht]
	\centering
	\caption{Comparison of multi-viewpoint learning on the Test 2 splits. All methods are trained with $M \!\in\! [1, 8]$ and $K \!=\! 7$ and tested with $M \!=\! 8$ and $K \!=\! 11$. The reported scores are averaged on datasets with similar properties. The top 2 are underlined, with the best in bold and the second best in italics.}
	\label{tab:multi_8_avg_test_2}
	\begin{small}
		\addtolength{\tabcolsep}{-1pt}
		\begin{tabular}{c|c|C{0.5in}C{0.5in}C{0.5in}C{0.5in}C{0.5in}C{0.5in}C{0.5in}C{0.5in}}
			\toprule
			Dataset          &  Method  &          ARI-A          &          AMI-A          &          ARI-O          &          AMI-O          &           IoU           &           F1            &           OCA           &           OOA           \\ \midrule
			\multirow{6}{*}{CLEVR \& SHOP} & SAVi (video) & 0.060 & 0.329 & 0.690 & 0.796 & N/A & N/A & 0.000 & N/A \\
			& SAVi & 0.004 & 0.042 & 0.048 & 0.102 & N/A & N/A & 0.000 & N/A \\
			& SIMONe & 0.255 & 0.396 & 0.573 & 0.623 & N/A & N/A & 0.000 & N/A \\
			& MulMON & \underline{\textit{0.457}} & \underline{\textit{0.523}} & 0.768 & 0.800 & N/A & N/A & \underline{\textit{0.131}} & N/A \\
			& Ablation & 0.265 & 0.430 & \underline{\textbf{0.837}} & \underline{\textit{0.838}} & \underline{\textit{0.331}} & \underline{\textit{0.449}} & 0.100 & \underline{\textbf{0.909}} \\
			& OCLOC & \underline{\textbf{0.545}} & \underline{\textbf{0.548}} & \underline{\textit{0.817}} & \underline{\textbf{0.844}} & \underline{\textbf{0.449}} & \underline{\textbf{0.572}} & \underline{\textbf{0.175}} & \underline{\textit{0.794}} \\
			\midrule
			\multirow{6}{*}{GSO \& ShapeNet} & SAVi (video) & 0.019 & 0.119 & 0.146 & 0.257 & N/A & N/A & 0.000 & N/A \\
			& SAVi & 0.006 & 0.043 & 0.041 & 0.095 & N/A & N/A & 0.000 & N/A \\
			& SIMONe & 0.326 & 0.316 & 0.221 & 0.361 & N/A & N/A & 0.000 & N/A \\
			& MulMON & 0.291 & 0.378 & 0.449 & 0.531 & N/A & N/A & 0.008 & N/A \\
			& Ablation & \underline{\textit{0.372}} & \underline{\textit{0.481}} & \underline{\textit{0.772}} & \underline{\textit{0.766}} & \underline{\textit{0.427}} & \underline{\textit{0.578}} & \underline{\textit{0.020}} & \underline{\textit{0.900}} \\
			& OCLOC & \underline{\textbf{0.708}} & \underline{\textbf{0.641}} & \underline{\textbf{0.816}} & \underline{\textbf{0.810}} & \underline{\textbf{0.562}} & \underline{\textbf{0.686}} & \underline{\textbf{0.260}} & \underline{\textbf{0.918}} \\
			\bottomrule
		\end{tabular}
	\end{small}
\end{table*}

The evaluation metrics are modified based on the ones described in \cite{Yuan2023Compositional} by considering the object constancy among viewpoints. These metrics evaluate the performance of different methods from four aspects. 1) \emph{Adjusted Rand Index} (ARI) \cite{Hubert1985Comparing} and \emph{Adjusted Mutual Information} (AMI) \cite{Nguyen2010Information} assess the quality of segmentation, i.e., how accurately images are partitioned into different objects and background. Previous work usually evaluates ARI and AMI only at pixels belong to objects, and how accurately background is separated from objects is unclear. We evaluate ARI and AMI under two conditions. ARI-A and AMI-A are computed considering both objects and background, while ARI-O and AMI-O are computed considering only objects. 2) \emph{Intersection over Union} (IoU) and \emph{$F_1$ score} (F1) assess the quality of amodal segmentation, i.e., how accurately complete shapes of objects are estimated. 3) \emph{Object Counting Accuracy} (OCA) assesses the accuracy of the estimated number of objects. 4) \emph{Object Ordering Accuracy} (OOA) as used in \cite{Yuan2019Generative} assesses the accuracy of the estimated pairwise ordering of objects. Formal definitions of these metrics are described in the Supplementary Material.

\subsection{Scene Decomposition}

\begin{figure*}[p]
	\captionsetup[subfigure]{labelformat=empty}
	\centering
	\subfloat[(a) SAVi (video)]{\includegraphics[width=0.67\columnwidth]{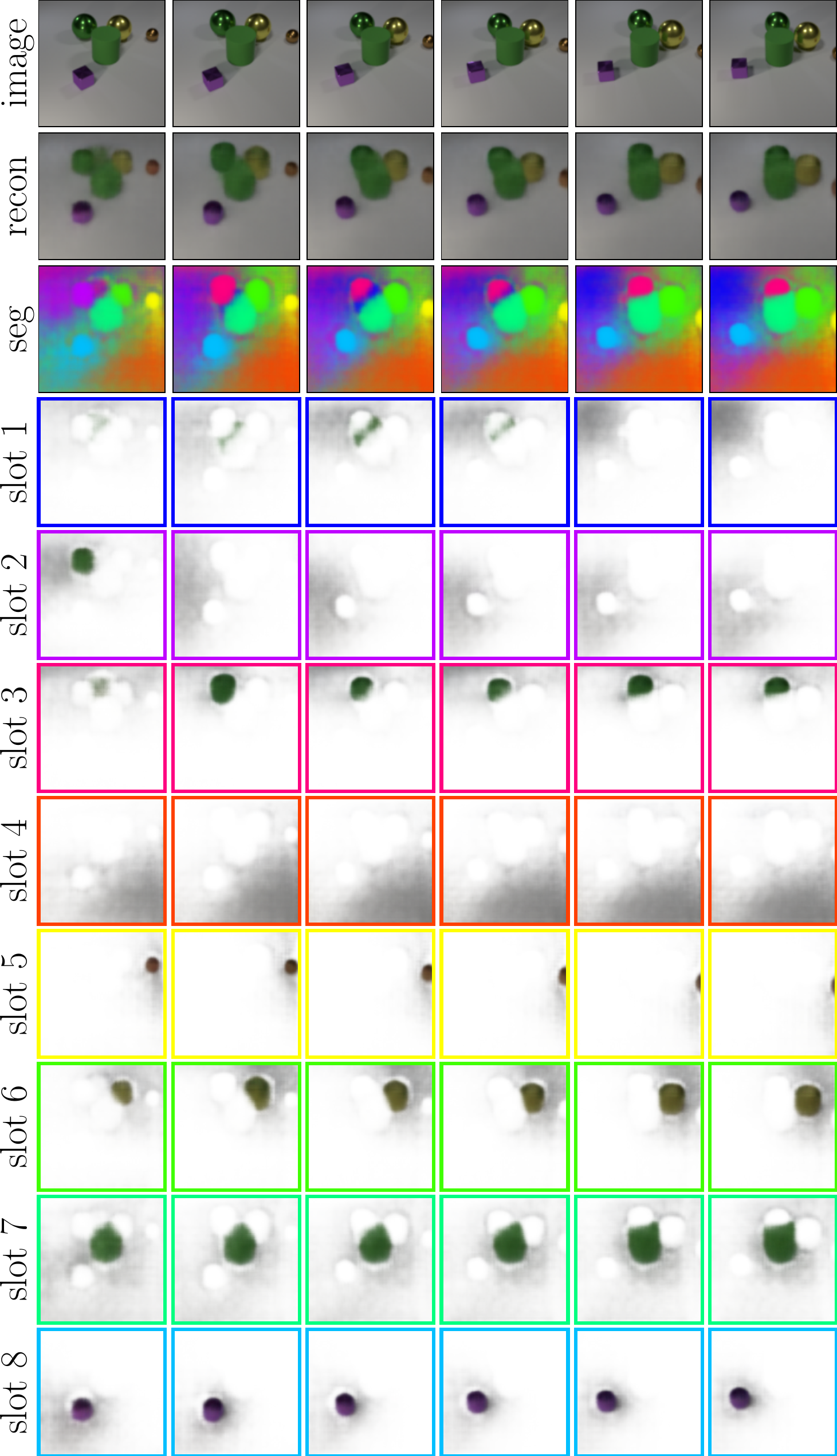}}%
	\hfill
	\subfloat[(b) SAVi]{\includegraphics[width=0.67\columnwidth]{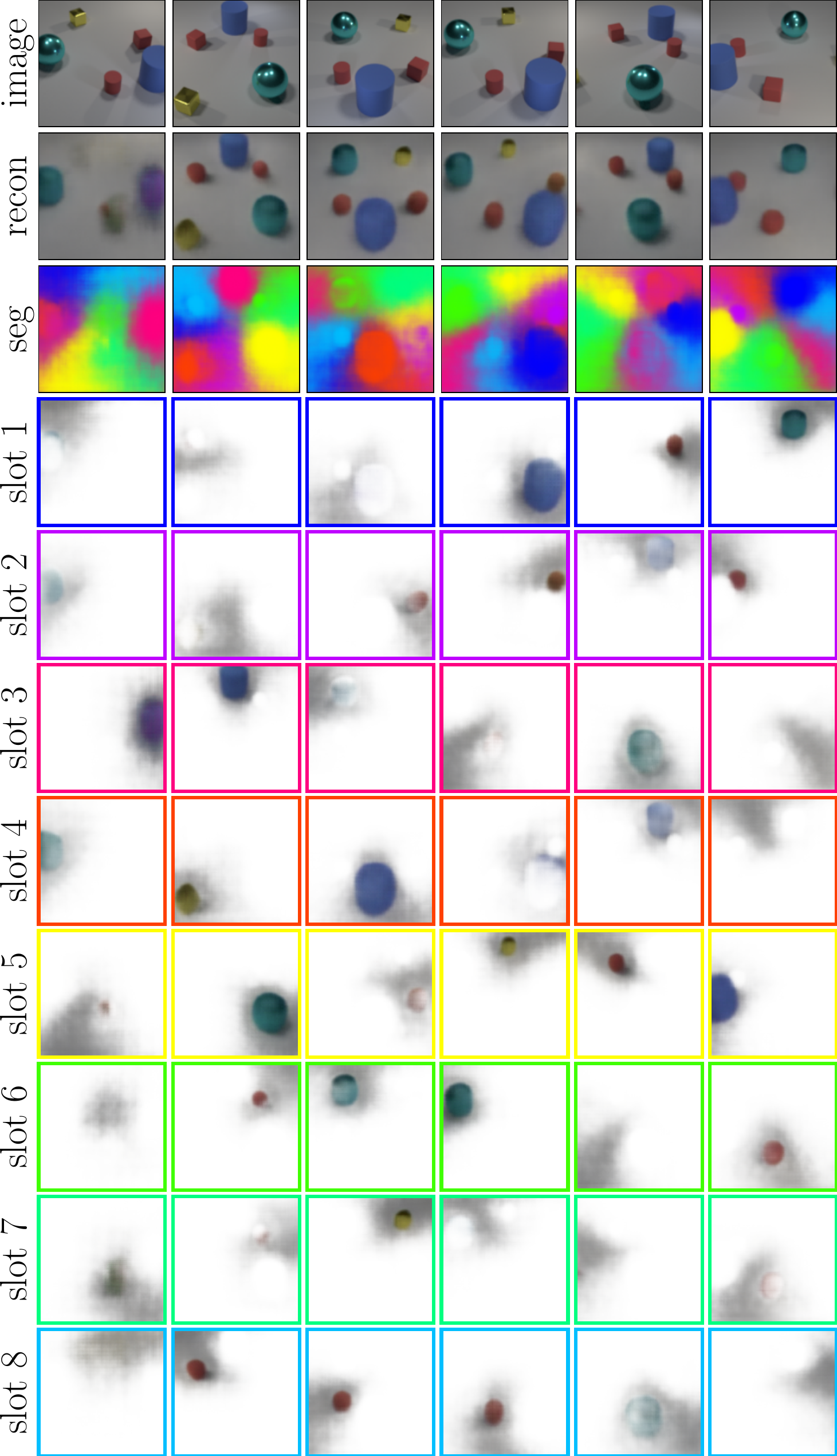}}%
	\hfill
	\subfloat[(c) SIMONe]{\includegraphics[width=0.67\columnwidth]{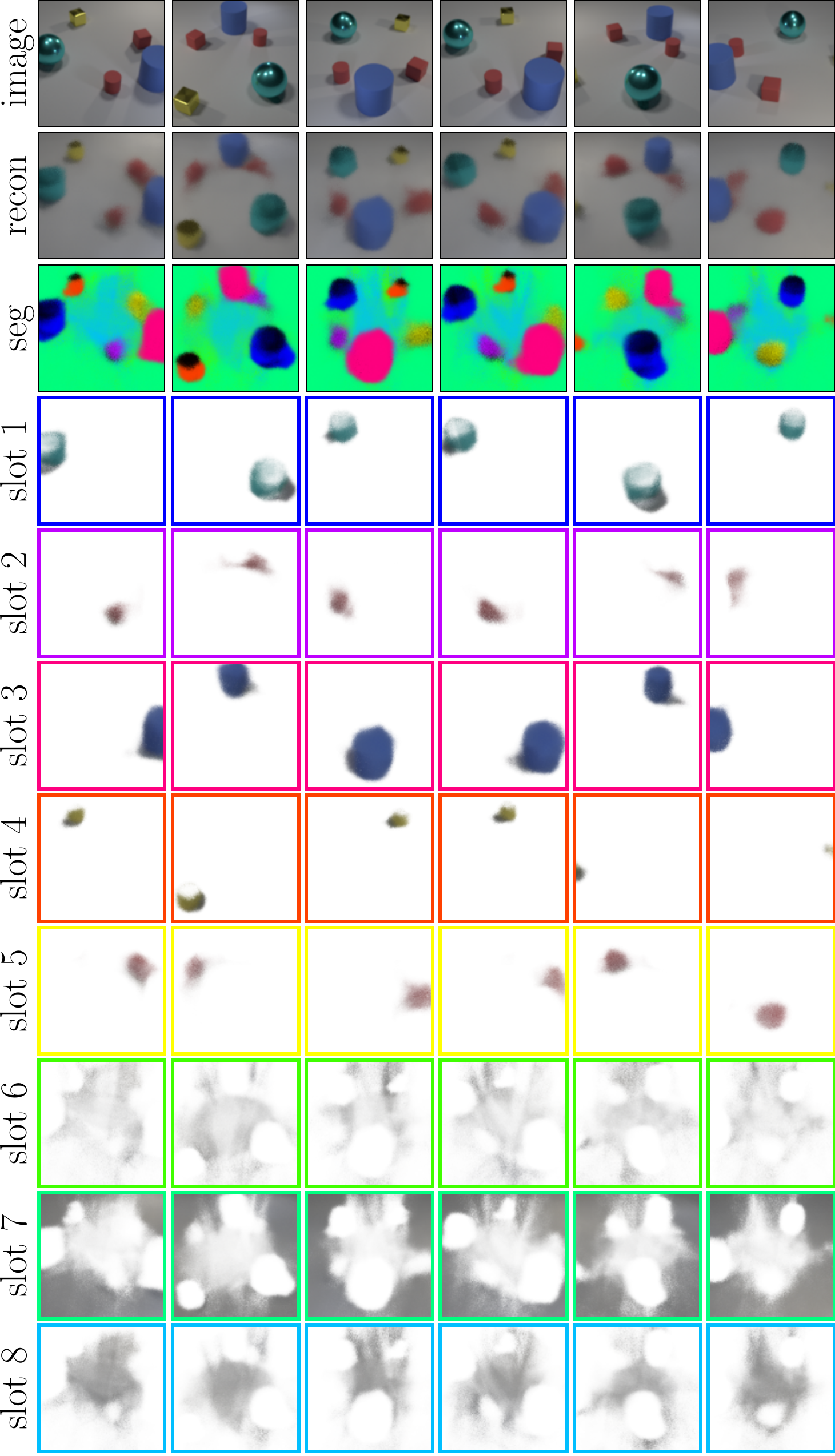}}%
	\\
	\subfloat[(d) MulMON]{\includegraphics[width=0.67\columnwidth]{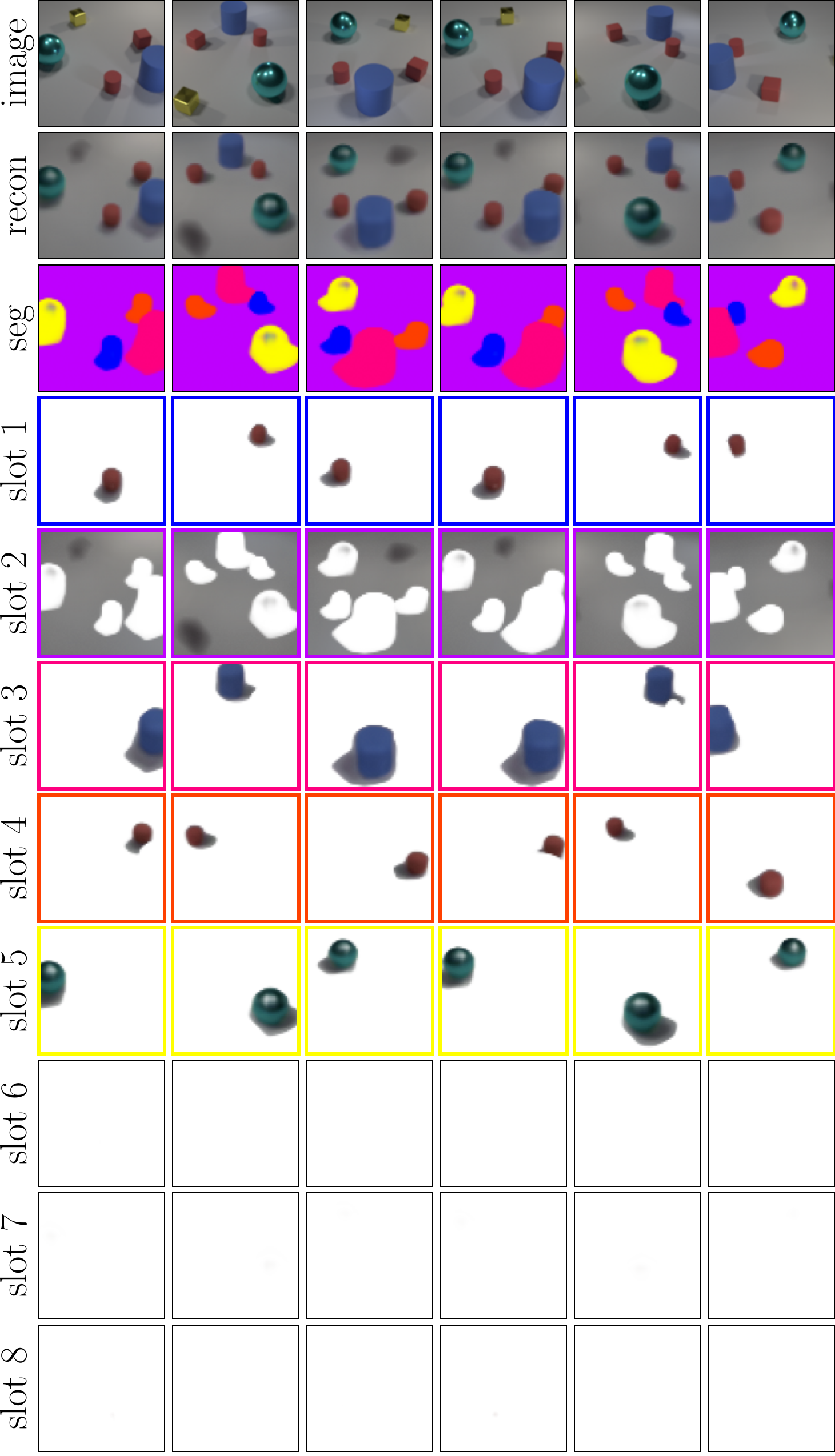}}%
	\hfill
	\subfloat[(e) Ablation]{\includegraphics[width=0.67\columnwidth]{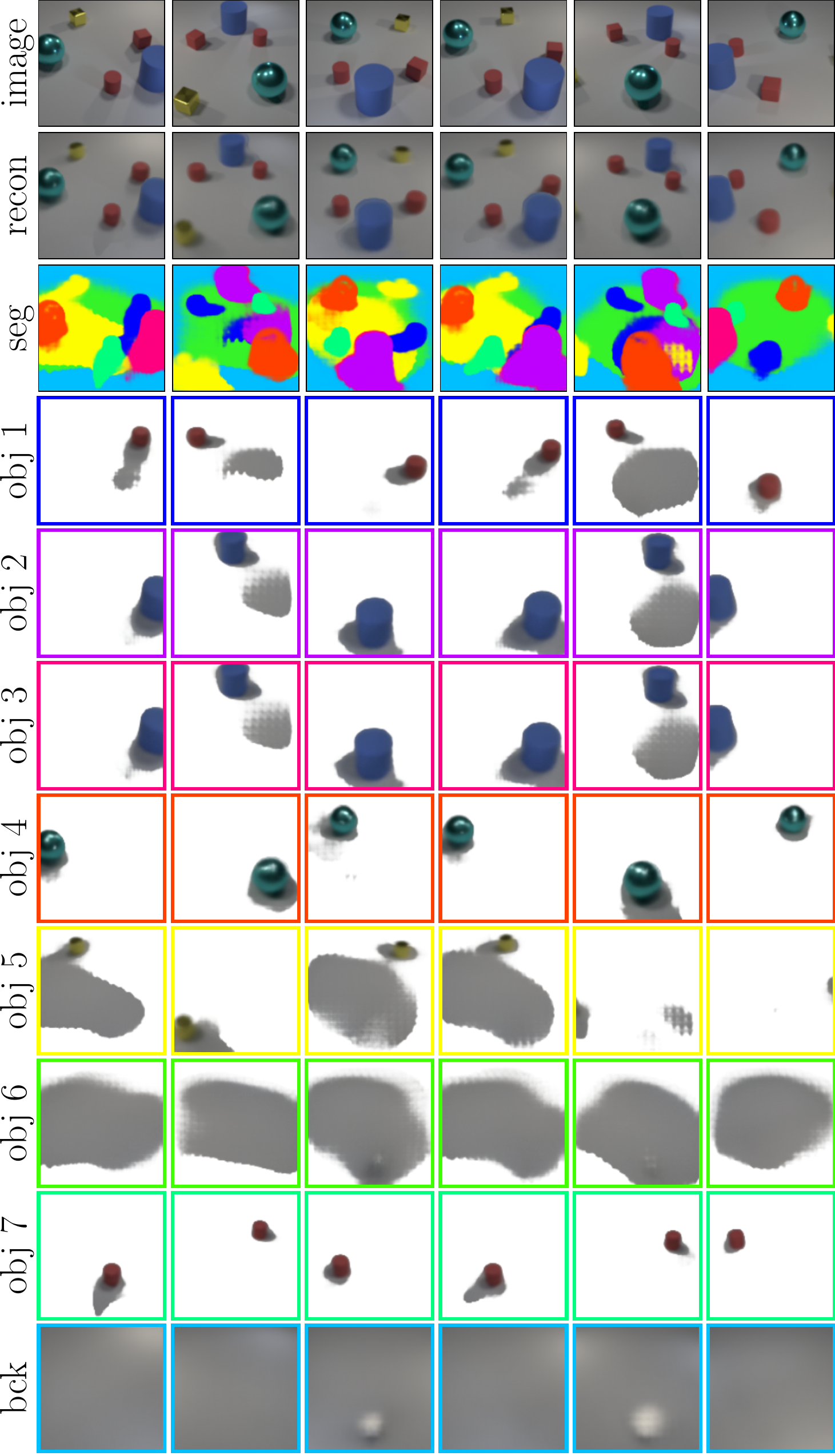}}%
	\hfill
	\subfloat[(f) OCLOC]{\includegraphics[width=0.67\columnwidth]{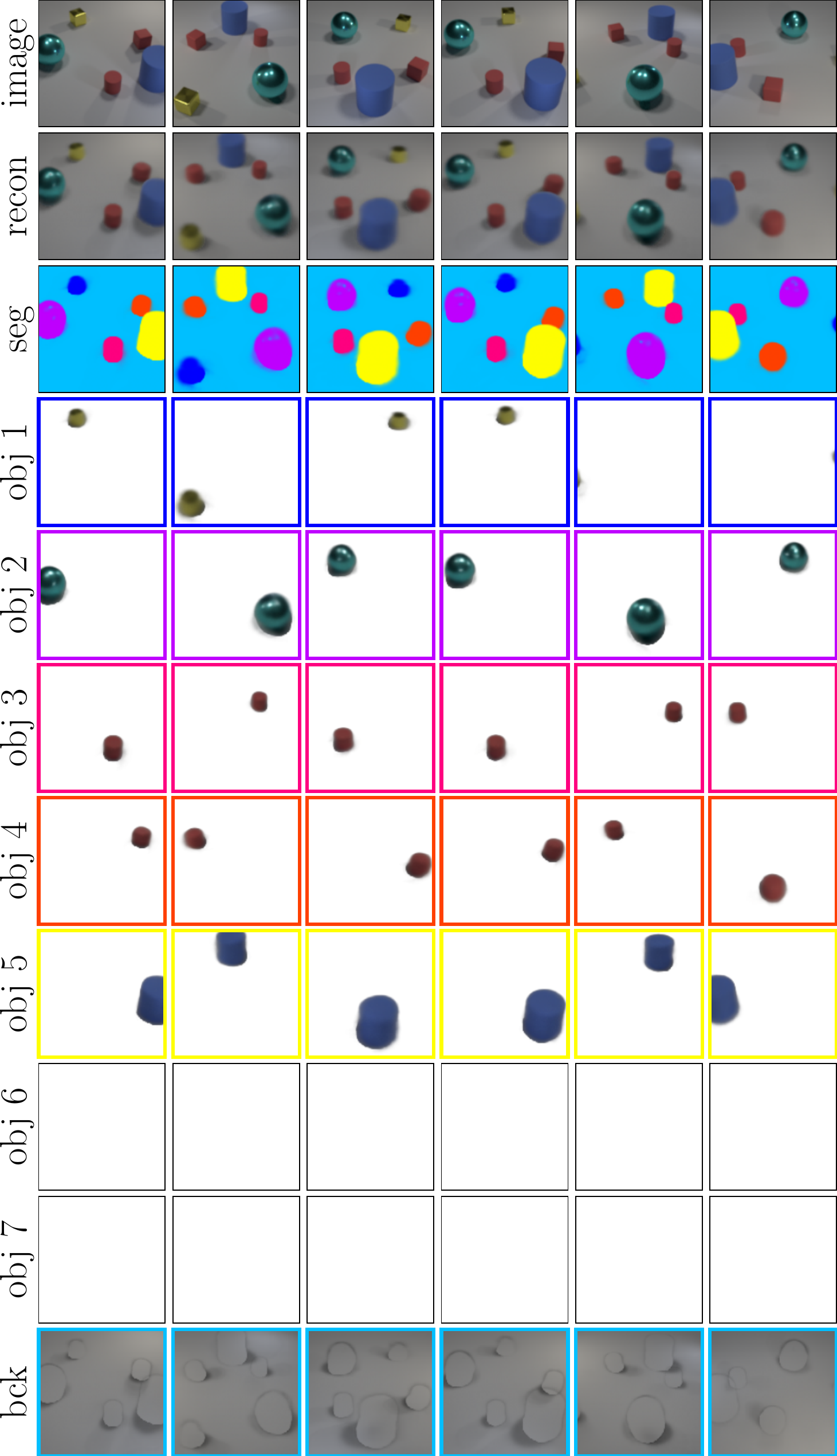}}%
	\caption{Scene decomposition results of different methods on the CLEVR dataset.}
	\label{fig:decompose_multi_clevr}
\end{figure*}

\begin{figure*}[p]
	\captionsetup[subfigure]{labelformat=empty}
	\centering
	\subfloat[(a) SAVi (video)]{\includegraphics[width=0.67\columnwidth]{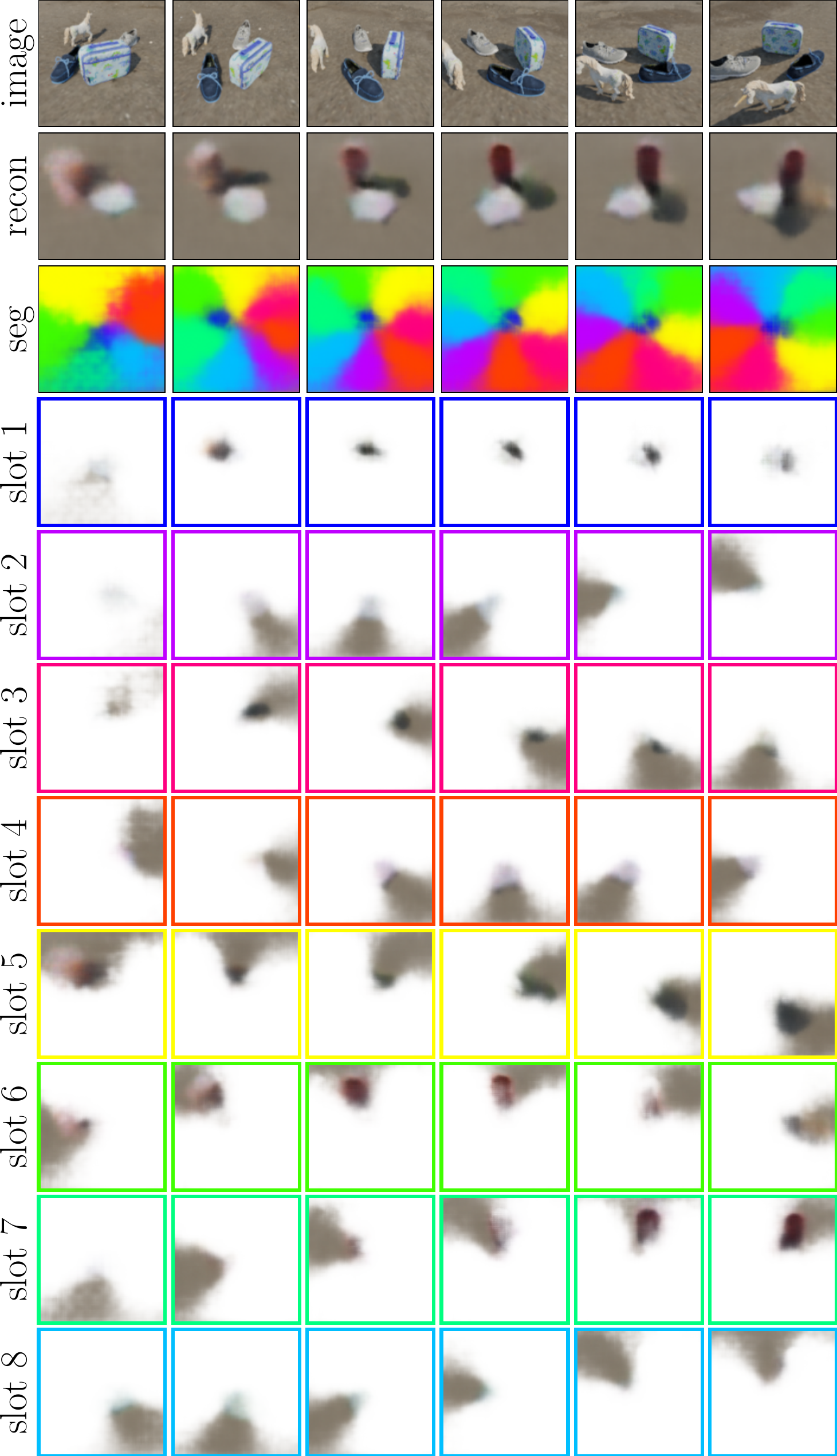}}%
	\hfill
	\subfloat[(b) SAVi]{\includegraphics[width=0.67\columnwidth]{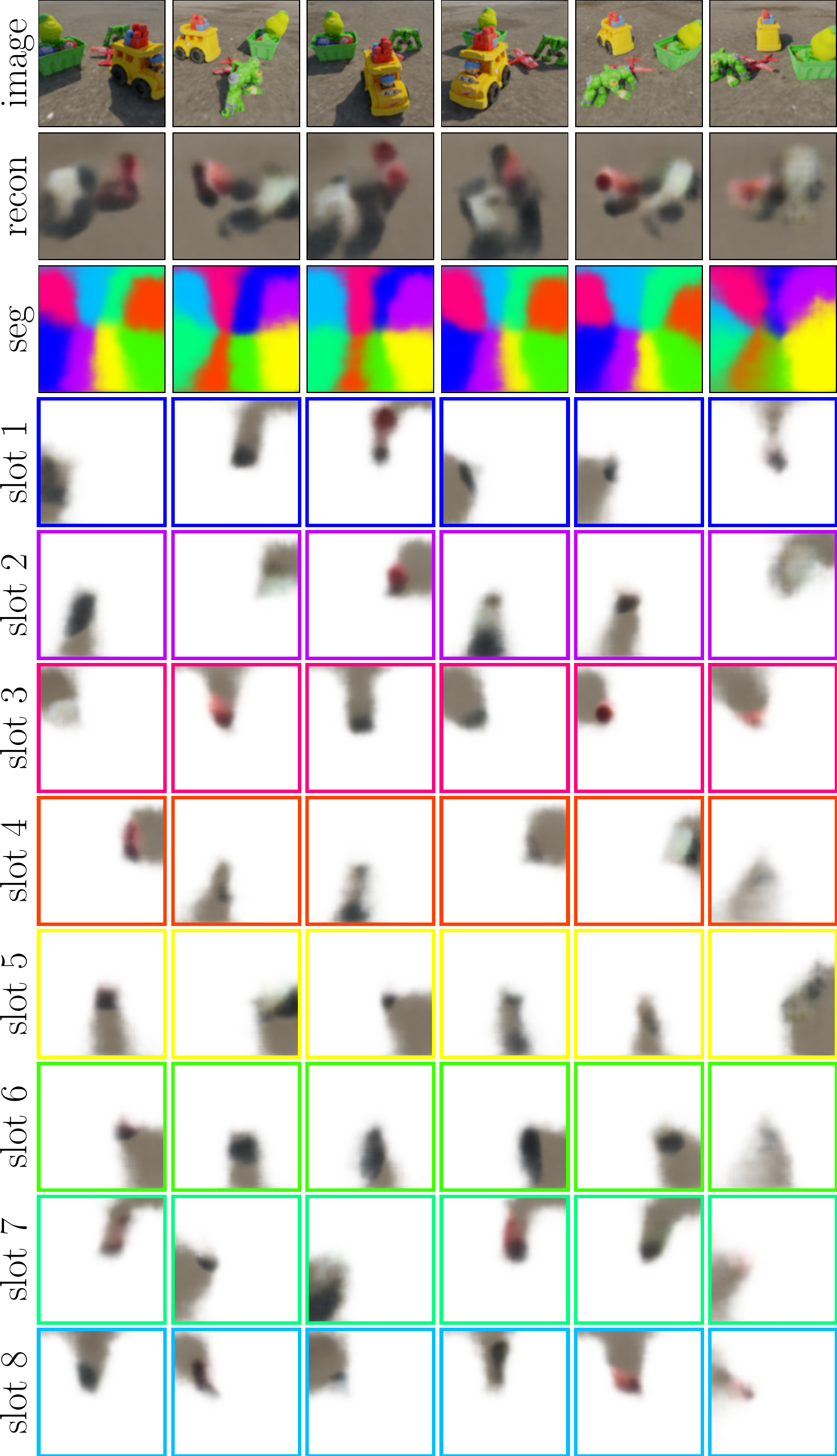}}%
	\hfill
	\subfloat[(c) SIMONe]{\includegraphics[width=0.67\columnwidth]{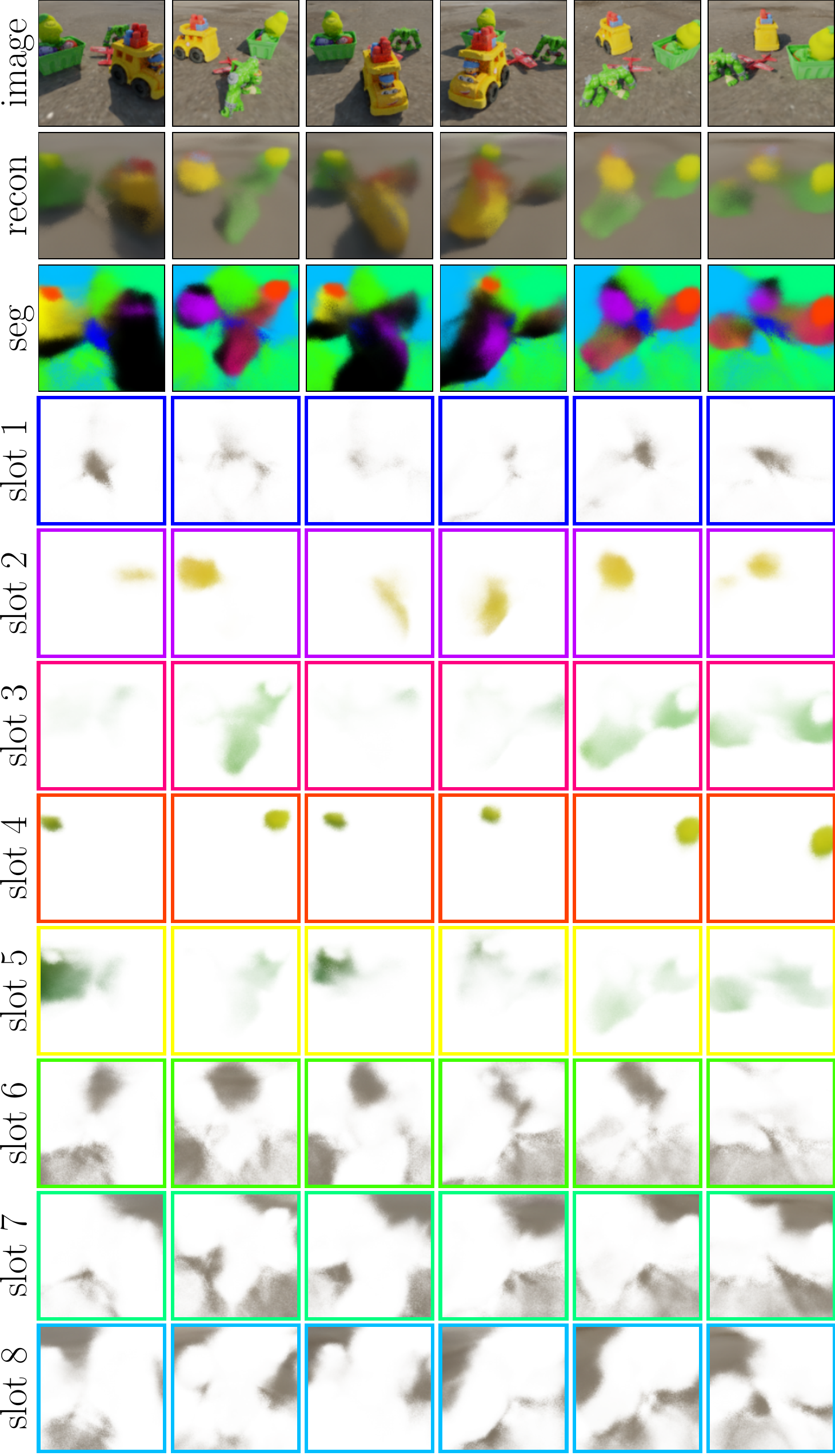}}%
	\\
	\subfloat[(d) MulMON]{\includegraphics[width=0.67\columnwidth]{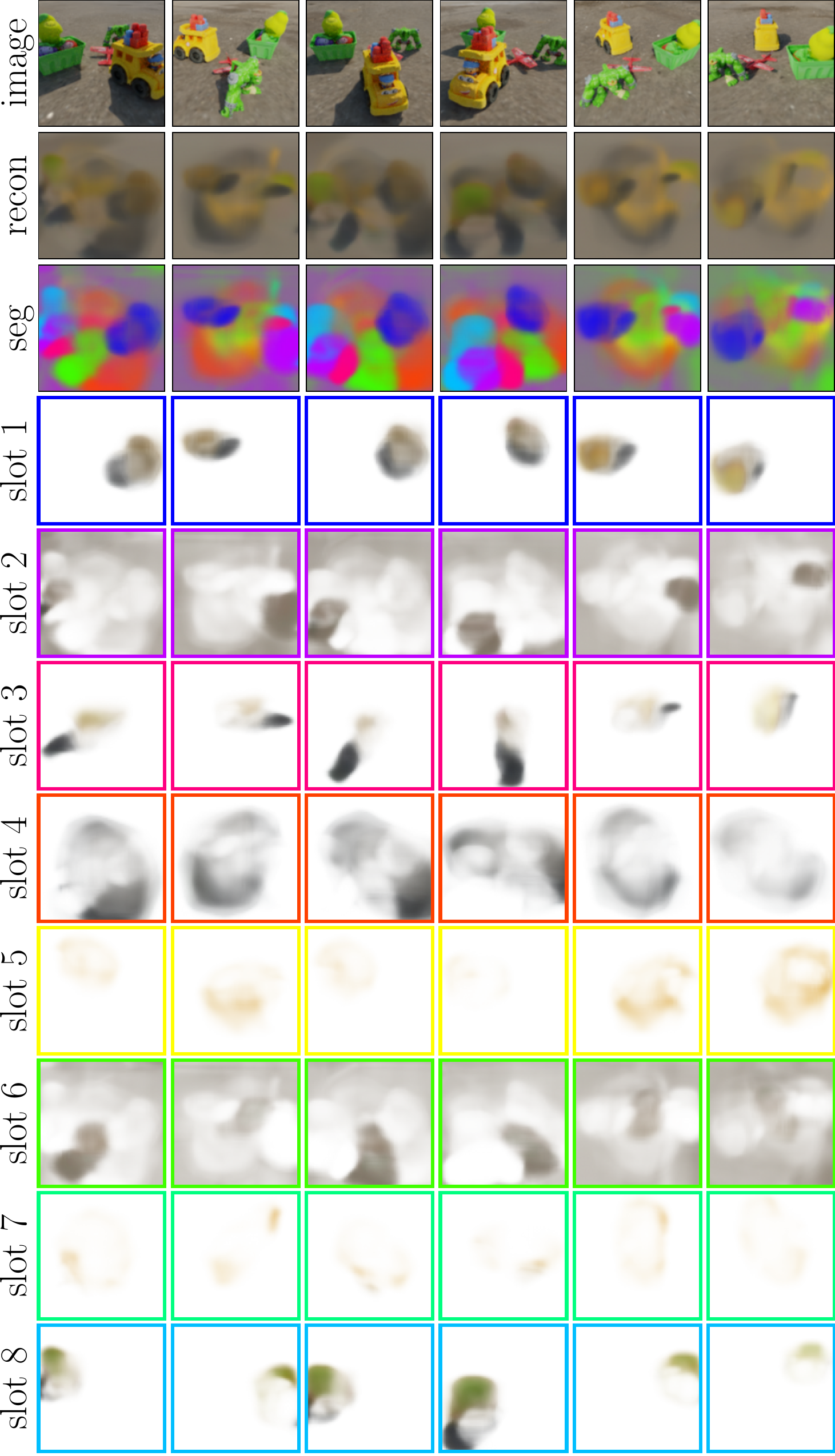}}%
	\hfill
	\subfloat[(e) Ablation]{\includegraphics[width=0.67\columnwidth]{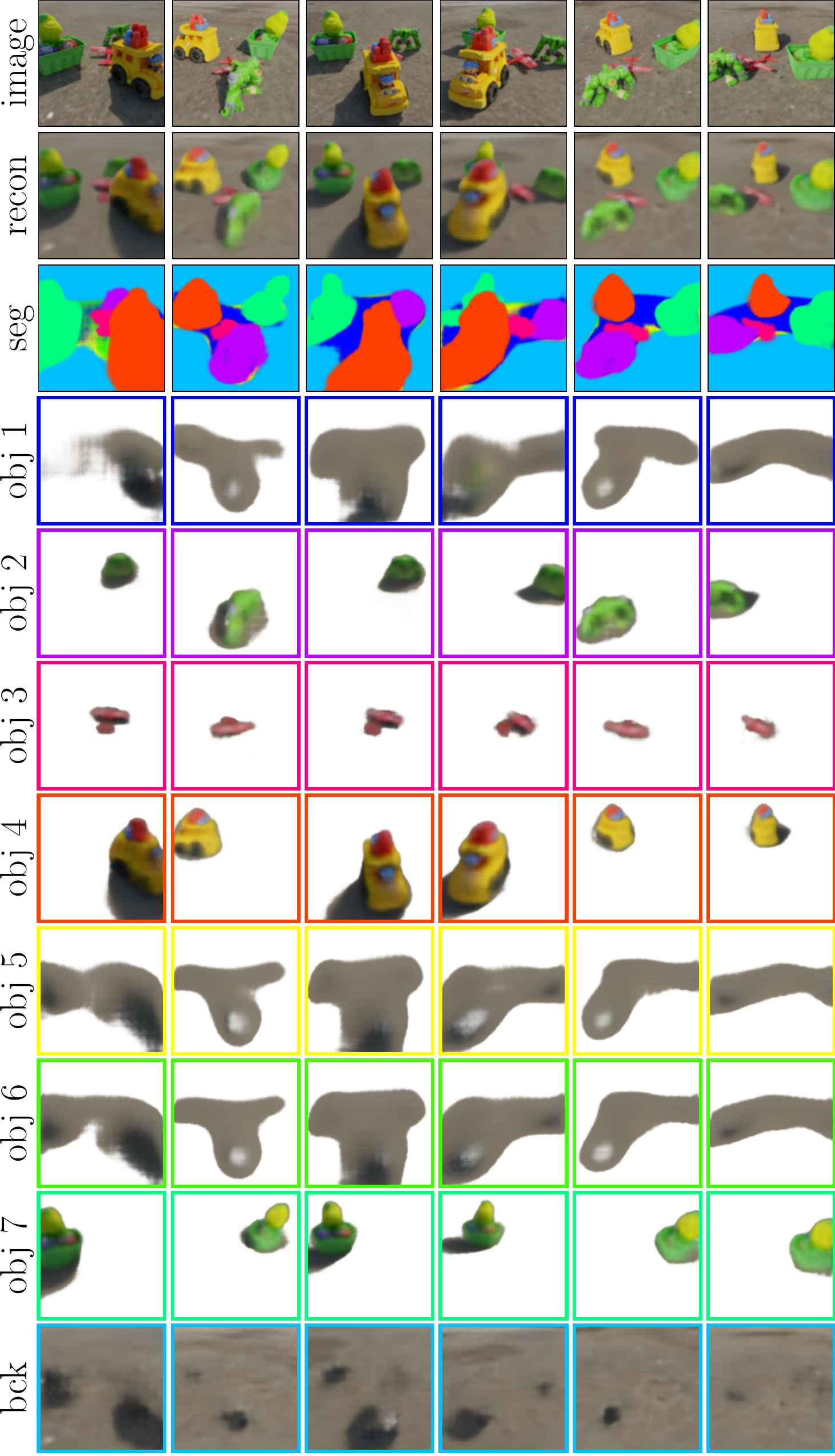}}%
	\hfill
	\subfloat[(f) OCLOC]{\includegraphics[width=0.67\columnwidth]{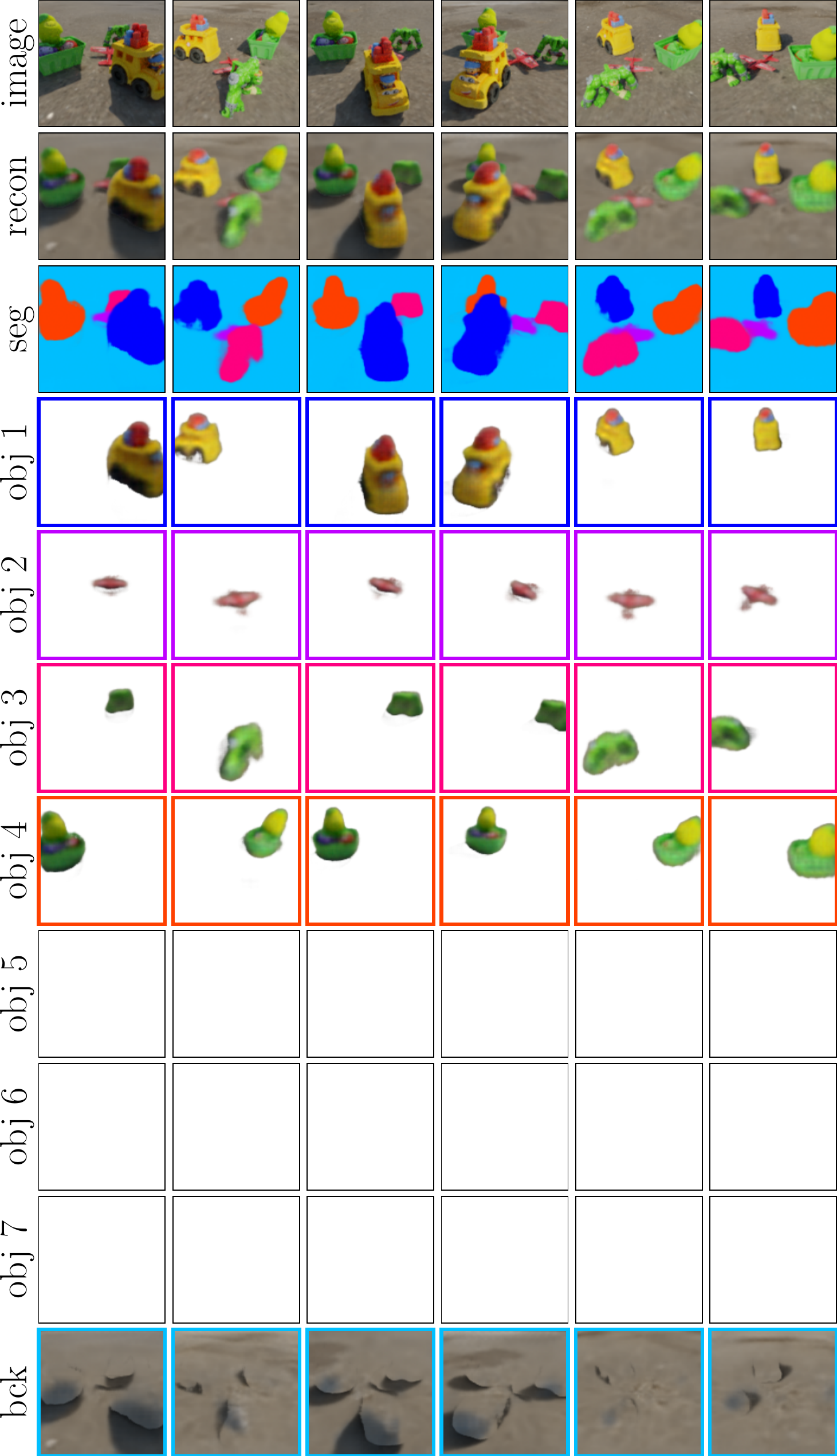}}%
	\caption{Scene decomposition results of different methods on the GSO dataset.}
	\label{fig:decompose_multi_gso}
\end{figure*}

\begin{figure}[!t]
	\centering
	\subfloat[Interpolating viewpoint latent variables]{\includegraphics[width=0.99\columnwidth]{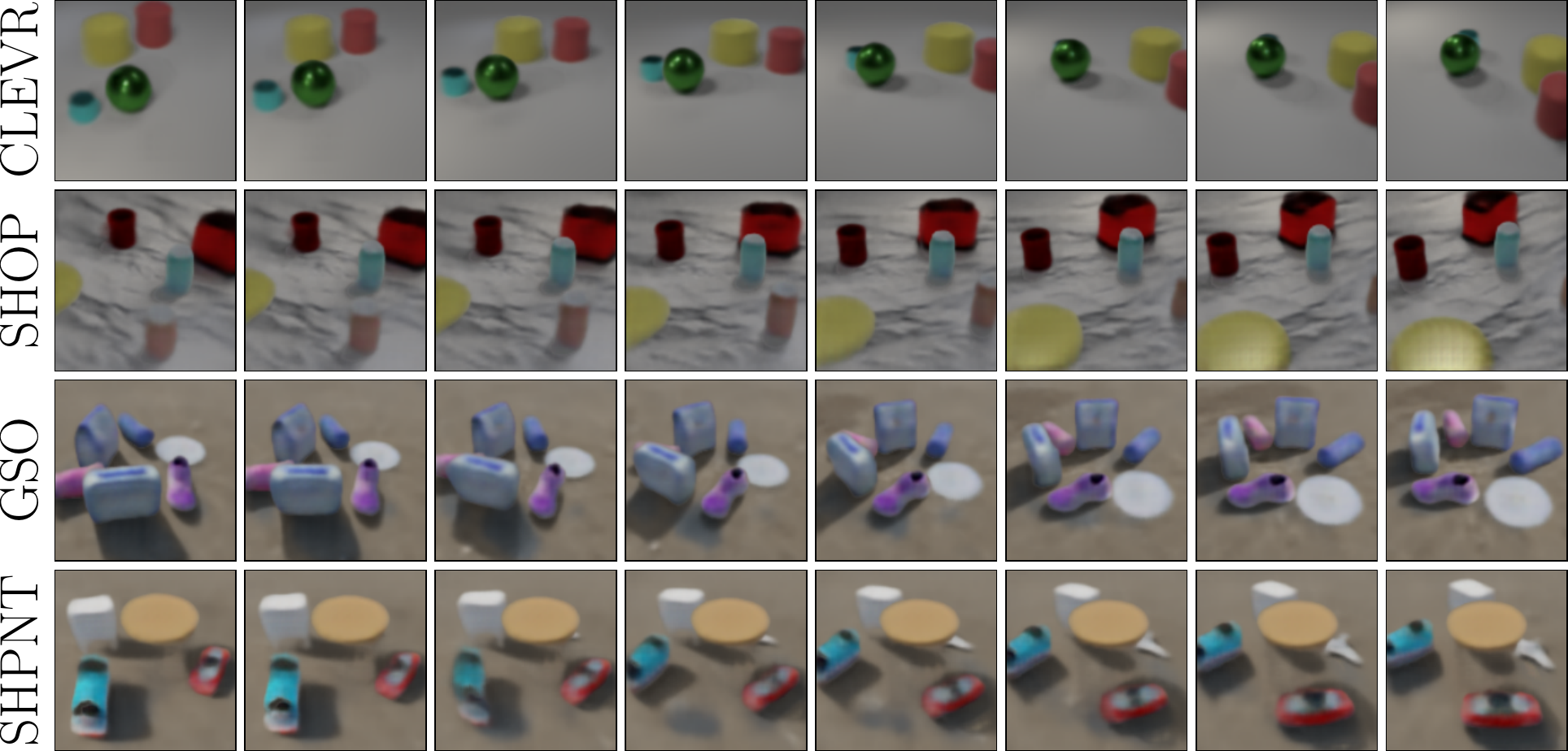}}%
	\\
	\subfloat[Sampling viewpoint latent variables]{\includegraphics[width=0.99\columnwidth]{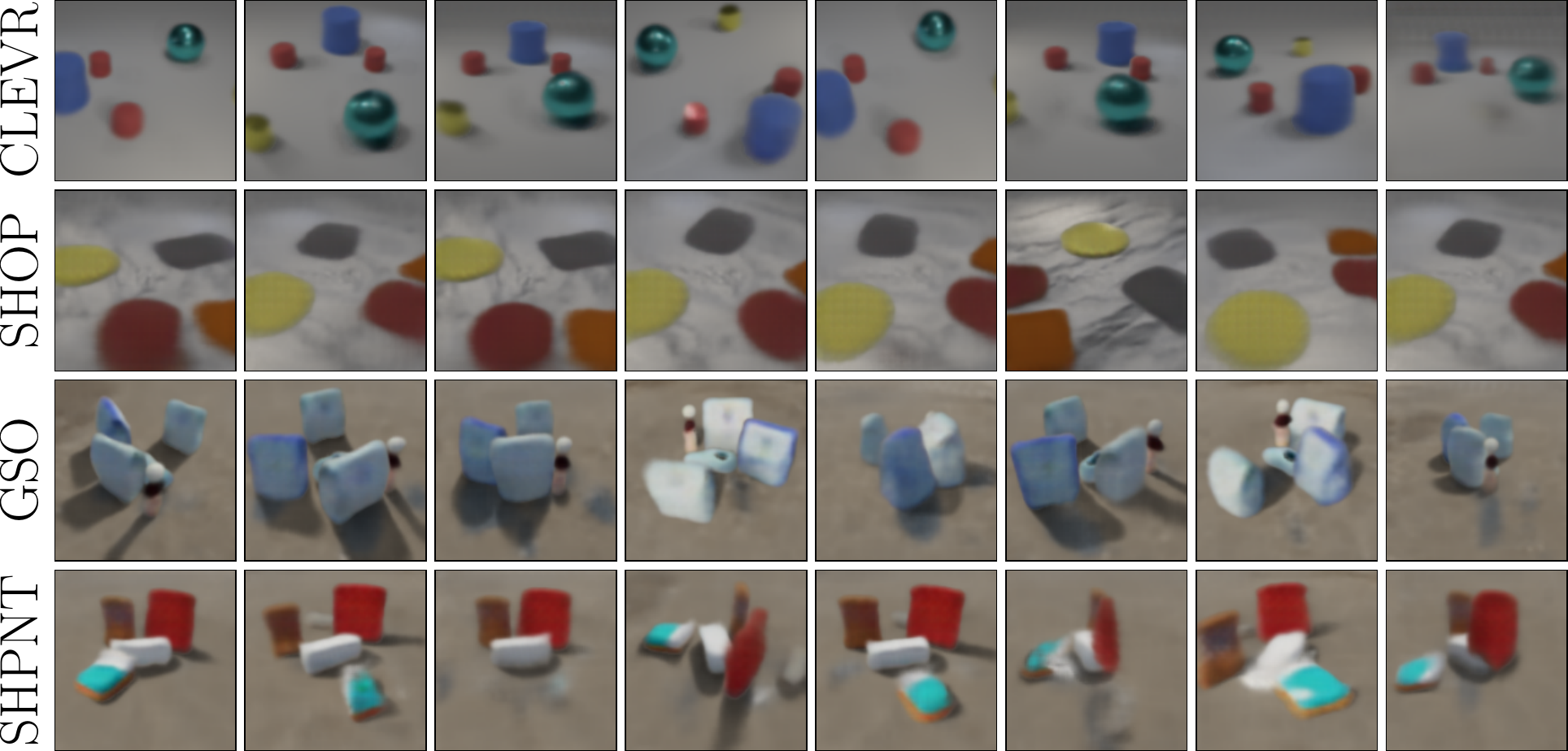}}%
	\caption{Viewpoint interpolation and sampling results of OCLOC.}
	\label{fig:generate_viewpoint}
\end{figure}

Qualitative results of different methods evaluated on the CLEVR and GSO datasets are shown in Figure \ref{fig:decompose_multi_clevr} and Figure \ref{fig:decompose_multi_gso}, respectively. Except SAVi, all the methods can separate objects and achieve \emph{object constancy} relatively well on the CLEVR dataset. As shown in sub-figure (a) of Figure \ref{fig:decompose_multi_clevr}, in a different setting where the model is trained and tested on video sequences, SAVi can achieve significantly better results. This phenomena indicates that SAVi (originally proposed for learning from videos) is not well suited for the considered problem setting. On the more visually complex GSO dataset, the ablation method and the proposed OCLOC decompose visual scenes relatively well, while the other methods do not learn very meaningful object-centric representations. Except the ablation method, all the compared methods cannot estimate the complete shapes of objects because the perceived shapes are directly obtained by normalizing the outputs of the decoder network. In addition, they cannot distinguish between objects and background because the modeling of objects and background is identical. On the CLEVR dataset, MulMON represents the background with a single slot, while SAVi and SIMONe represent the background with multiple slots. On the GSO dataset, these methods all divide the background into several parts. The proposed OCLOC can estimate complete images of objects even if objects are almost fully occluded (e.g., object 2 in column 3 of sub-figure (f) in Figure {\ref{fig:decompose_multi_gso}}) because the complete shapes of objects are explicitly considered in the modeling of visual scenes. In addition, OCLOC is able to not only distinguish between objects and background, but also accurately reconstruct the complete background. Additional results on the SHOP and ShapeNet datasets are provided in the Supplementary Material.

Quantitative comparison of scene decomposition performance is presented in Table \ref{tab:multi_8_avg_test_1}. The reported scores are averaged on datasets with similar properties, i.e., CLEVR \& SHOP, GSO \& ShapeNet. Detailed results of each dataset are included in the Supplementary Material. Because SAVI, SIMONe, and MulMON do not explicitly model the complete shapes and depth ordering of object, the IoU, F1, and OOA scores which require the estimations of complete shapes and depth ordering are not evaluated for them. Although these methods do not model the number of objects in the visual scene, it is still possible to estimate the number of objects and compute the OCA score based on the scene decomposition results, in a reasonable though heuristic way. More specifically, let $\boldsymbol{r} \!\in\! {\{0, 1\}^{M \!\times\! N \!\times\! (K + 1)}}$ be the estimated pixel-wise partition of $K \!+\! 1$ slots in $M$ viewpoints. Whether the object or background represented by the $k$th slot is considered to be included in the visual scene can be computed by $\max_{m}{\max_{n}{\,r_{m,n,k}}}$, and the computation of the estimated number of objects $\tilde{K}$ is described below.
\begin{equation}
	\tilde{K} = \sum\nolimits_{k=0}^{K}{\Big(\max_{m}{\max_{n}{\,r_{m,n,k}}}\Big)} - 1
\end{equation}
The ablation method and the proposed OCLOC explicitly model the varying number of objects and distinguish background from objects. Therefore, the IoU, F1, OCA, and OOA scores are all computed based on the inference results. Compared to the \emph{partially supervised} MulMON, the proposed \emph{unsupervised} OCLOC achieves competitive or better results, which have validated the effectiveness of OCLOC in learning from multiple unspecified viewpoints.

\subsection{Generalizability}

Because visual scenes are modeled compositionally by the proposed method, the trained models are generalizable to novel visual scenes containing more objects than the ones used for training. The performance of different methods when visual scenes contain more objects than the ones used for training is shown in Table \ref{tab:multi_8_avg_test_2}. Although the increased number of objects in the visual scene makes it more difficult to extract compositional scene representations, the proposed method performs reasonably well, which has validated the generalizability of this method.

\subsection{Effectiveness of Shadow Modeling}

The proposed method considers the shadows of objects in the modeling of visual scenes, and the effectiveness of shadow modeling is evaluated both qualitatively and quantitatively. According to Figures \ref{fig:decompose_multi_clevr} and \ref{fig:decompose_multi_gso}, most of the shadows are excluded in the scene decomposition results if shadows are explicitly modeled (sub-figure (f)), while shadows are considered to be parts of objects in methods without explicit shadow modeling (sub-figure (e)). According to Tables \ref{tab:multi_8_avg_test_1} and \ref{tab:multi_8_avg_test_2}, the proposed OCLOC outperforms the ablation method which does not explicitly model shadows in most cases, especially in terms of ARI-A, AMI-A, IoU, and F1 scores. The major reason is that the ablation method tends to treat regions of shadows as objects, while they are considered as background in the ground truth annotations. The behavior that the ablation method treats shadows as parts of objects instead of background is desirable, because the shadows will change accordingly as objects move. The proposed OCLOC uses representations of objects to generate images of shadows in consideration of compositionality, and is able to distinguish the shadows of objects from the objects themselves because the shadows and shapes of objects are modeled differently.

\subsection{Viewpoint Modification}

Multi-viewpoint images of the same visual scene can be generated by first inferring compositional scene representations and then modifying latent variables of viewpoints. Results of interpolating and sampling viewpoint latent variables are illustrated in Figure \ref{fig:generate_viewpoint}. It can be seen from the generated multi-viewpoint images that the proposed method is able to appropriately modify viewpoints.

\section{Conclusions}

In this paper, we have considered a novel problem of learning compositional scene representations from multiple unspecified viewpoints in a fully unsupervised way and proposed a deep generative model called OCLOC to solve this problem. The proposed OCLOC separates latent representations of each visual scene into a viewpoint-independent part and a viewpoint-dependent part, and it performs inference by first randomly initializing and then iteratively updating latent representations using inference networks that can integrate the information contained in different viewpoints. On several specifically designed synthesized datasets, the proposed fully unsupervised method achieves competitive or better results compared with a state-of-the-art method with viewpoint supervision. It also outperforms state-of-the-arts that assume temporal relationships among viewpoints in the considered problem setting. Experimental results have validated the effectiveness of the proposed method in learning compositional scene representations from multiple unknown and unrelated viewpoints without any supervision. In addition, the proposed method can distinguish between objects and background more precisely than the ablation method which does not explicitly consider shadows of objects in the modeling of visual scenes. This ablation study has verified the effectiveness of shadow modeling in the proposed method.

\ifCLASSOPTIONcaptionsoff
  \newpage
\fi



\bibliographystyle{IEEEtran}
\bibliography{reference}
%

%

\begin{IEEEbiography}[{\includegraphics[width=1in,height=1.25in,clip,keepaspectratio]{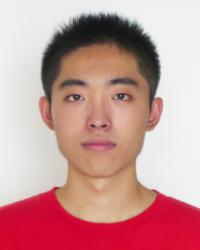}}]{Jinyang Yuan}
received the BS degree in physics from Nanjing University, China, the MS degree in electrical engineering from University of California, San Diego, and the PhD degree in computer science from Fudan University, China. His research interests include computer vision, machine learning, and deep generative models.
\end{IEEEbiography}

\begin{IEEEbiography}[{\includegraphics[width=1in,height=1.25in,clip,keepaspectratio]{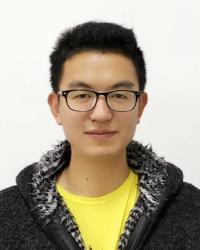}}]{Tonglin Chen}
is currently pursuing the PhD degree in computer science from Fudan University, Shanghai, China. His current research interests include machine learning, deep generative models, and object-centric representation learning.
\end{IEEEbiography}

\begin{IEEEbiography}[{\includegraphics[width=1in,height=1.25in,clip,keepaspectratio]{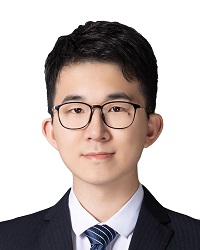}}]{Zhimeng Shen}
is currently pursuing the Master degree in computer science from Fudan University, Shanghai, China. His current research interests include diffusion models and object-centric representation learning.
\end{IEEEbiography}

\begin{IEEEbiography}[{\includegraphics[width=1in,height=1.25in,clip,keepaspectratio]{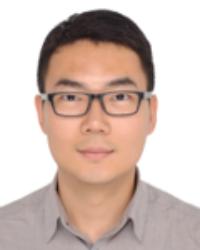}}]{Bin Li}
received the PhD degree in computer science from Fudan University, Shanghai, China. He is an associate professor with the School of Computer Science, Fudan University, Shanghai, China. Before joining Fudan University, Shanghai, China, he was a lecturer with the University of Technology Sydney, Australia and a senior research scientist with Data61 (formerly NICTA), CSIRO, Australia. His current research interests include machine learning and visual intelligence, particularly in compositional scene representation, modeling and inference.
\end{IEEEbiography}

\begin{IEEEbiography}[{\includegraphics[width=1in,height=1.25in,clip,keepaspectratio]{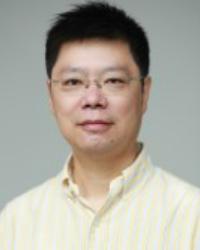}}]{Xiangyang Xue}
received the BS, MS, and PhD
degrees in communication engineering from
Xidian University, Xian, China, in 1989, 1992,
and 1995, respectively. He is currently a professor of computer science with Fudan University,
Shanghai, China. His research interests include
multimedia information processing and machine
learning.
\end{IEEEbiography}

\vfill








\clearpage

\begin{appendices}
\section{Evaluation Metrics}

The evaluation metrics are modified based on the ones described in \cite{Yuan2023Compositional} by considering the object constancy among viewpoints. Details are described below.

\subsection{Adjusted Rand Index (ARI)}

$\hat{K}_{i}$ denotes the ground truth number of objects in the $i$th visual scene of the test set, and $\hat{\boldsymbol{r}}^{i} \!\in\! {\{0, 1\}^{M \!\times\! N \!\times\! (\hat{K}_{i} + 1)}}$ is the ground truth pixel-wise partition of objects and background in the $M$ images of this visual scene. $K$ denotes the maximum
number of objects that may appear in the visual scene, and $\boldsymbol{r}^{i} \!\in\! {\{0, 1\}^{M \!\times\! N \!\times\! (K + 1)}}$ is the estimated partition. ARI is computed using the following expression.
\begin{equation}
	\label{equ:ari}
	\text{ARI} = \frac{1}{I} \sum_{i=1}^{I}{\frac{b_{\text{all}}^{i} - b_{\text{row}}^{i} \cdot b_{\text{col}}^{i} / c^{i}}{\big(b_{\text{row}}^{i} + b_{\text{col}}^{i}\big) / 2 - b_{\text{row}}^{i} \cdot b_{\text{col}}^{i} / c^{i}}}
\end{equation}
Let $\mathcal{S}$ be a set of pixel indexes of $M$ images. The $b_{\text{all}}^{i}$, $b_{\text{row}}^i$, $b_{\text{col}}^i$, and $c^{i}$ in Eq. \eqref{equ:ari} are computed by
\begin{align}
	b_{\text{all}}^{i} & = \sum\nolimits_{\hat{k}=0}^{\hat{K}_{i}}{\sum\nolimits_{k=0}^{K}{C\big(a_{\hat{k},k}^{i}, 2\big)}} \\
	b_{\text{row}}^i & = \sum\nolimits_{\hat{k}=0}^{\hat{K}_i}{C\Big(\sum\nolimits_{k=0}^{K}{a_{\hat{k},k}^{i}}, 2\Big)} \\ b_{\text{col}}^i & = \sum\nolimits_{k=0}^{K}{C\Big(\sum\nolimits_{\hat{k}=0}^{\hat{K}_i}{a_{\hat{k},k}^{i}}, 2\Big)} \\
	c^{i} & = C\Big(\sum\nolimits_{\hat{k}=0}^{\hat{K}_i}{\sum\nolimits_{(m,n) \in \mathcal{S}}{\hat{r}_{m,n,\hat{k}}^i}}, 2\Big)
\end{align}
In the above expressions, $C(\cdot, \cdot)$ is the combinatorial function and $a_{\hat{k},k}^{i}$ is an intermediate variable. The computations of $C(\cdot, \cdot)$ and $a_{\hat{k},k}^{i}$ are described below.
\begin{align}
	C(x, y) & = \frac{x!}{(x - y)! \, y!} \\
	a_{\hat{k},k}^{i} & = \sum\nolimits_{m,n \in \mathcal{S}}{\big(\hat{r}_{m,n,\hat{k}}^{i} \cdot r_{m,n,k}^{i}\big)}
\end{align}
When computing ARI-A, $\mathcal{S}$ is the collection of all the pixels in the $M$ images, i.e., $\mathcal{S} \!=\! \{1, \dots, M\} \!\times\! \{1, \dots, N\}$. When computing ARI-O, $\mathcal{S}$ corresponds to all the pixels belonging to objects in the $M$ images.

\subsection{Adjusted Mutual Information (AMI)}

The meanings of $\hat{K}_{i}$, $\hat{\boldsymbol{r}}^{i}$, $K$, $\boldsymbol{r}^{i}$, and $\mathcal{S}$ are identical to the ones in the descriptions of ARI. Let $(m_j, n_j)$ be the $j$ element in the set $\mathcal{S}$. $\hat{l}_{j}^{i} = \argmax\nolimits_{\hat{k}}{\hat{\boldsymbol{r}}_{m_j,n_j}^{i}} \!\in\! \{0, 1, \dots, \hat{K}_{i}\}$ and $l_{j}^{i} = \argmax\nolimits_{k}{\boldsymbol{r}_{m_j,n_j}^{i}} \!\in\! \{0, 1, \dots, K\}$ are indexes of the ground truth layers and the estimated layers observed at each pixel, respectively. AMI is computed by
\begin{equation}
	\text{AMI} = \frac{1}{I} \sum_{i=1}^{I}{\frac{\MI(\hat{\boldsymbol{l}}^{i}, \boldsymbol{l}^{i}) - \mathbb{E}[\MI(\hat{\boldsymbol{l}}^{i}, \boldsymbol{l}^{i})]}{\big(\!\entropy(\hat{\boldsymbol{l}}^{i}) + \entropy(\boldsymbol{l}^{i})\big) / 2 - \mathbb{E}[\MI(\hat{\boldsymbol{l}}^{i}, \boldsymbol{l}^{i})]}}
\end{equation}
In the above expression, $\MI$ denotes mutual information and $\entropy$ denotes entropy. When computing AMI-A/AMI-O, the choice of $\mathcal{S}$ is the same as ARI-A/ARI-O.

\subsection{Intersection over Union (IoU)}

IoU can be used to evaluate the performance of amodal instance segmentation. Compared to ARI and AMI, it provides extra information about the estimation of occluded regions of objects because complete shapes instead of perceived shapes of objects are used to compute this metric. Let $\hat{\boldsymbol{s}}^{i} \!\in\! [0, 1]^{M \!\times\! N \!\times\! \hat{K}_{i}}$ and $\boldsymbol{s}^{i} \!\in\! [0, 1]^{M \!\times\! N \!\times\! K}$ denote the ground truth and estimated shapes of objects in the $M$ images of the $i$th visual scene of the test set, respectively. Because both the number and the indexes of the estimated objects may be different from the ground truth, $\hat{\boldsymbol{s}}^{i}$ and $\boldsymbol{s}^{i}$ cannot be compared directly. Let $\Xi$ be the set of all the $K!$ possible permutations of the indexes $\{1, 2, \dots, K\}$. $\boldsymbol{\xi}^{i} \!\in\! \Xi$ is a permutation chosen based on the ground truth $\hat{\boldsymbol{r}}^{i}$ and estimated $\boldsymbol{r}^{i}$ partitions of objects and background, and is computed by	$\boldsymbol{\xi}^{i} \!=\! \max_{\boldsymbol{\xi} \in \Xi}{\sum_{k=1}^{\hat{K}_i}{\!\sum_{m=1}^{M}{\!\sum_{n=1}^{N}{\hat{r}_{m,n,k}^{i} \cdot r_{m,n,\xi_k^i}^{i}}}}}$. IoU is computed using the following expression.
\begin{equation}
	\label{equ:iou}
	\text{IoU} = \frac{1}{I} \sum_{i=1}^{I}{\frac{1}{\hat{K}_i} \sum_{k=1}^{\hat{K}_i}{\frac{d_{\text{inter}}}{d_{\text{union}}}}}
\end{equation}
In Eq. \eqref{equ:iou}, $d_{\text{inter}}$ and $d_{\text{union}}$ are computed by
\begin{align}
	d_{\text{inter}} & = \sum\nolimits_{m=1}^{M}{\sum\nolimits_{n=1}^{N}{\min(\hat{s}_{m,n,k}^{i}, s_{m,n,\xi_k^i}^{i})}} \\
	d_{\text{union}} & = \sum\nolimits_{m=1}^{M}{\sum\nolimits_{n=1}^{N}{\max(\hat{s}_{m,n,k}^{i}, s_{m,n,\xi_k^i}^{i})}}
\end{align}
Although the set $\Xi$ contain $K!$ elements, the permutation $\boldsymbol{\xi}^{i}$ can still be computed efficiently by formulating the computation as a linear sum assignment problem.

\subsection{$\boldsymbol{F_1}$ Score (F1)}

$F_1$ score can also be used to assess the performance of amodal segmentation like IoU, and is computed in a similar way. The meanings of $\hat{\boldsymbol{s}}^{i}$, $\boldsymbol{s}^{i}$, $\boldsymbol{\xi}$, and $\Xi$ as well as the computations of $d_{\text{inter}}$ and $d_{\text{union}}$ are identical to the ones in the descriptions of IoU. F1 is computed by
\begin{equation}
	\text{F1} = \frac{1}{I} \sum_{i=1}^{I}{\frac{1}{\hat{K}_i} \sum_{k=1}^{\hat{K}_i}{\frac{2 \cdot d_{\text{inter}}}{d_{\text{inter}} + d_{\text{union}}}}}
\end{equation}

\subsection{Object Counting Accuracy (OCA)}

$\hat{K}_{i}$ and $\tilde{K}_{i}$ denote the ground truth number and the estimated number of objects in the $i$th visual scene of the test set, respectively. Let $\delta$ denote the Kronecker delta function. The computation of OCA is described below.
\begin{equation}
	\text{OCA} = \frac{1}{I} \sum\nolimits_{i=1}^{I}{\delta_{\hat{K}_{i}, \tilde{K}_{i}}}
\end{equation}

\subsection{Object Ordering Accuracy (OOA)}

Let $\hat{t}_{m,k_1, k_2}^{i} \!\in\! \{0, 1\}$ and $t_{m,k_1, k_2}^{i} \!\in\! \{0, 1\}$ denote the ground truth and estimated pairwise depth orderings of the $k_1$th and $k_2$th objects in the $m$th viewpoint of the $i$th image, respectively. The correspondences between the ground truth and estimated indexes of objects are determined based on the permutation of indexes $\boldsymbol{\xi}^{i}$ as described in the computation of IoU. Because the depth ordering of two objects is hard to estimate if these objects do not overlap, the computation of OOA described below measures the importance of different pairs of objects with different weights.
\begin{equation}
	\label{equ:ooa}
	\text{OOA} \!=\! \frac{1}{I} \! \sum_{i=1}^{I}{\!\frac{\sum\nolimits_{k_1=1}^{\hat{K}_i - 1}{\!\sum\nolimits_{k_2=k_1 + 1}^{\hat{K}_i}{w_{m\!,k_1\!,k_2}^{i} \delta_{\hat{t}_{m\!,k_1\!,k_2}^{i}\!, t_{m\!,k_1\!,k_2}^{i}}}}}{\sum\nolimits_{k_1=1}^{\hat{K}_i - 1}{\!\sum\nolimits_{k_2=k_1 + 1}^{\hat{K}_i}{w_{m\!,k_1\!,k_2}^{i}}}}}
\end{equation}
In Eq. \eqref{equ:ooa}, the weight $w_{m\!,k_1\!,k_2}^{i}$ is computed by
\begin{equation}
	w_{m\!,k_1\!,k_2}^{i} = \sum\nolimits_{n=1}^{N}{\hat{s}_{m,n,k_1}^{i} \cdot \hat{s}_{m,n,k_2}^{i}}
\end{equation}
$w_{m,k_1,k_2}^{i}$ measures the overlapped area of the ground truth shapes $\hat{\boldsymbol{s}}^{i}$ of the $k_1$th and the $k_2$th objects. The more the two objects overlap, the easier it is to determine the depth ordering of these objects, and thus the more important it is for the model to estimate the depth ordering correctly.

\section{Choices of Hyperparameters}

\subsection{Proposed Method}

\label{sec:hyperparameter_ocloc}

In the generative model, the standard deviation $\sigma_{\text{x}}$ of the likelihood function is chosen as $0.2$. The maximum number of objects that may appear in the visual scene is $K = 7$ during training, $K = 7$ when testing on the Test 1 split, and $K = 11$ when testing on the Test 2 split. The hyperparameter $\alpha$ is chosen to be $4.5$. The respective dimensionalities of latent variables $\boldsymbol{z}_{m}^{\text{view}}$, $\boldsymbol{z}^{\text{bck}}$, and $\boldsymbol{z}_{k}^{\text{obj}}$ with $1 \!\leq\! k \!\leq\! K$ are chosen as $E_{\text{view}} = 4$, $E_{\text{bck}} = 8$, $E_{\text{obj}} = 64$ for the CLEVR and SHOP datasets, and $E_{\text{view}} = 16$, $E_{\text{bck}} = 32$, $E_{\text{obj}} = 256$ for the GSO and ShapeNet datasets.

In the variational inference, the hyperparameter $T$ is set to $3$. The dimensionalities of intermediate variables $\boldsymbol{r}_{m}^{\text{view}}$ and $\boldsymbol{r}_{k}^{\text{attr}}$, and keys and values in the cross-attention are $D_{\text{vw}} = 8$, $D_{\text{at}} = 128$, $D_{\text{key}} = 64$, $D_{\text{val}} = 136$ for the CLEVR and SHOP datasets, and $D_{\text{vw}} = 32$, $D_{\text{at}} = 512$, $D_{\text{key}} = 256$, $D_{\text{val}} = 544$ for the GSO and ShapeNet datasets.

In the learning, the batch size is chosen to be $4$. The initial learning rate is $1 \times 10^{-4}$, and is decayed exponentially with a factor $0.5$ during the training. We have found that the optimization of neural networks with randomly initialized weights tend to get stuck into undesired local optima. To solve this problem, a better initialization of weights is obtained by using only one viewpoint per visual scene to train neural networks in the first $\num[group-separator={,}]{100000}$ steps for the CLEVR and SHOP datasets, and in the first $\num[group-separator={,}]{200000}$ steps for the GSO and ShapeNet datasets.

The choices of neural networks in both the generative model and the variational inference are included in the provided source code. Instead of adopting a superior but more time-consuming method such as grid search, we manually choose the hyperparameters of neural networks based on experience. Details of the hyperparameters of neural networks are provided below.

\begin{itemize}[leftmargin=*]
	\item In the decoder networks (Figure 4 of the main paper)
	\begin{itemize}
		\item 2: Fully Connected Layers
		\begin{itemize}
			\item Fully Connected, out 256, SiLU
			\item Fully Connected, out 256, SiLU
			\item Fully Connected, out 64, SiLU
		\end{itemize}
		\item 4: Position Embedding
		\begin{itemize}
			\item Fully Connected, out 64, Sinusoid
		\end{itemize}
		\item 5: Transformer Layers (CLEVR and SHOP)
		\begin{itemize}
			\item Transformer Encoder, head 4, out 64, hidden 128, SiLU
			\item Transformer Encoder, head 4, out 64, hidden 128, SiLU
		\end{itemize}
		\item 5: Transformer Layers (GSO and ShapeNet)
		\begin{itemize}
			\item Transformer Encoder, head 4, out 128, hidden 256, SiLU
			\item Transformer Encoder, head 4, out 128, hidden 256, SiLU
		\end{itemize}
		\item 7: ConvTranspose Layers
		\begin{itemize}
			\item ConvTranspose, kernel 4 $\times$ 4, stride 2, out 64, SiLU
			\item ConvTranspose, kernel 3 $\times$ 3, out 32, SiLU
			\item ConvTranspose, kernel 4 $\times$ 4, stride 2, out 32, SiLU
			\item ConvTranspose, kernel 3 $\times$ 3, out 16, SiLU
			\item ConvTranspose, kernel 4 $\times$ 4, stride 2, out 16, SiLU
			\item ConvTranspose, kernel 3 $\times$ 3, out 3, Linear
		\end{itemize}
		\item 10: Fully Connected Layers
		\begin{itemize}
			\item Fully Connected, out 512, SiLU
			\item Fully Connected, out 512, SiLU
			\item Fully Connected, out 1, Linear
		\end{itemize}
		\item 11: Fully Connected Layers
		\begin{itemize}
			\item Fully Connected, out 1024, SiLU
			\item Fully Connected, out 1024, SiLU
			\item Fully Connected, out 128, SiLU
		\end{itemize}
		\item 13: Position Embedding
		\begin{itemize}
			\item Fully Connected, out 128, Sinusoid
		\end{itemize}
		\item 14: Transformer Layers (CLEVR and SHOP)
		\begin{itemize}
			\item Transformer Encoder, head 8, out 128, hidden 256, SiLU
			\item Transformer Encoder, head 8, out 128, hidden 256, SiLU
		\end{itemize}
		\item 14: Transformer Layers (GSO and ShapeNet)
		\begin{itemize}
			\item Transformer Encoder, head 8, out 256, hidden 512, SiLU
			\item Transformer Encoder, head 8, out 256, hidden 512, SiLU
		\end{itemize}
		\item 16: ConvTranspose Layers
		\begin{itemize}
			\item ConvTranspose, kernel 4 $\times$ 4, stride 2, out 128, SiLU
			\item ConvTranspose, kernel 3 $\times$ 3, out 64, SiLU
			\item ConvTranspose, kernel 4 $\times$ 4, stride 2, out 64, SiLU
			\item ConvTranspose, kernel 3 $\times$ 3, out 32, SiLU
			\item ConvTranspose, kernel 4 $\times$ 4, stride 2, out 32, SiLU
			\item ConvTranspose, kernel 3 $\times$ 3, out 3 \!+\! 1 \!+\! 1 \!+\! 1, Linear
		\end{itemize}
	\end{itemize}
	\item In the encoder networks (Figure 4 of the main paper)
	\begin{itemize}
		\item 2: Convolutional Layers (CLEVR and SHOP)
		\begin{itemize}
			\item Convolutional, kernel 4 $\times$ 4, stride 2, out 64, SiLU
			\item Convolutional, kernel 5 $\times$ 5, out 64, SiLU
			\item Convolutional, kernel 5 $\times$ 5, out 64, SiLU
			\item Convolutional, kernel 5 $\times$ 5, out 64, SiLU
			\item Convolutional, kernel 5 $\times$ 5, out 64, SiLU
		\end{itemize}
		\item 2: Convolutional Layers (GSO and ShapeNet)
		\begin{itemize}
			\item Convolutional, kernel 4 $\times$ 4, stride 2, out 256, SiLU
			\item Convolutional, kernel 5 $\times$ 5, out 256, SiLU
			\item Convolutional, kernel 5 $\times$ 5, out 256, SiLU
			\item Convolutional, kernel 5 $\times$ 5, out 256, SiLU
			\item Convolutional, kernel 5 $\times$ 5, out 256, SiLU
		\end{itemize}
		\item 4: Position Embedding (CLEVR and SHOP)
		\begin{itemize}
			\item Fully Connected, out 64, Linear
		\end{itemize}
		\item 4: Position Embedding (GSO and ShapeNet)
		\begin{itemize}
			\item Fully Connected, out 256, Linear
		\end{itemize}
		\item 5: Fully Connected Layers (CLEVR and SHOP)
		\begin{itemize}
			\item LayerNorm
			\item Fully Connected, out 64, SiLU
			\item Fully Connected, out 64, Linear
		\end{itemize}
		\item 5: Fully Connected Layers (GSO and ShapeNet)
		\begin{itemize}
			\item LayerNorm
			\item Fully Connected, out 256, SiLU
			\item Fully Connected, out 256, Linear
		\end{itemize}
		\item 6: Fully Connected Layers (CLEVR and SHOP)
		\begin{itemize}
			\item LayerNorm
			\item Fully Connected, no bias, out 64, Linear
		\end{itemize}
		\item 6: Fully Connected Layers (GSO and ShapeNet)
		\begin{itemize}
			\item LayerNorm
			\item Fully Connected, no bias, out 256, Linear
		\end{itemize}
		\item 7: Fully Connected Layers (CLEVR and SHOP)
		\begin{itemize}
			\item LayerNorm
			\item Fully Connected, no bias, out 136, Linear
		\end{itemize}
		\item 7: Fully Connected Layers (GSO and ShapeNet)
		\begin{itemize}
			\item LayerNorm
			\item Fully Connected, no bias, out 544, Linear
		\end{itemize}
		\item 11: Fully Connected Layers (CLEVR and SHOP)
		\begin{itemize}
			\item LayerNorm
			\item Fully Connected, no bias, out 64, Linear
		\end{itemize}
		\item 11: Fully Connected Layers (GSO and ShapeNet)
		\begin{itemize}
			\item LayerNorm
			\item Fully Connected, no bias, out 256, Linear
		\end{itemize}
		\item 13: GRU Layer (CLEVR and SHOP)
		\begin{itemize}
			\item GRU, out 136
		\end{itemize}
		\item 13: GRU Layer (GSO and ShapeNet)
		\begin{itemize}
			\item GRU, out 544
		\end{itemize}
		\item 14: Fully Connected Layers (CLEVR and SHOP)
		\begin{itemize}
			\item LayerNorm
			\item Fully Connected, out 128, SiLU
			\item Fully Connected, out 136, Linear
		\end{itemize}
		\item 14: Fully Connected Layers (GSO and ShapeNet)
		\begin{itemize}
			\item LayerNorm
			\item Fully Connected, out 512, SiLU
			\item Fully Connected, out 544, Linear
		\end{itemize}
		\item 19: Fully Connected Layers (CLEVR and SHOP)
		\begin{itemize}
			\item Fully Connected, out 512, SiLU
			\item Fully Connected, out 512, SiLU
			\item Fully Connected, out 4 \!+\! 4, Linear
		\end{itemize}
		\item 19: Fully Connected Layers (GSO and ShapeNet)
		\begin{itemize}
			\item Fully Connected, out 512, SiLU
			\item Fully Connected, out 512, SiLU
			\item Fully Connected, out 16 \!+\! 16, Linear
		\end{itemize}
		\item 21: Fully Connected Layers (CLEVR and SHOP)
		\begin{itemize}
			\item Fully Connected, out 512, SiLU
			\item Fully Connected, out 512, SiLU
			\item Fully Connected, out 64 \!+\! 64 \!+\! 2 \!+\! 1, Linear
		\end{itemize}
		\item 21: Fully Connected Layers (GSO and ShapeNet)
		\begin{itemize}
			\item Fully Connected, out 512, SiLU
			\item Fully Connected, out 512, SiLU
			\item Fully Connected, out 256 \!+\! 256 \!+\! 2 \!+\! 1, Linear
		\end{itemize}
		\item 23: Fully Connected Layers
		\begin{itemize}
			\item Fully Connected, out 512, SiLU
			\item Fully Connected, out 512 \!+\! 1, Linear
		\end{itemize}
		\item 27: Fully Connected Layers (CLEVR and SHOP)
		\begin{itemize}
			\item Fully Connected, out 512, SiLU
			\item Fully Connected, out 512, SiLU
			\item Fully Connected, out 8 \!+\! 8, Linear
		\end{itemize}
		\item 27: Fully Connected Layers (GSO and ShapeNet)
		\begin{itemize}
			\item Fully Connected, out 512, SiLU
			\item Fully Connected, out 512, SiLU
			\item Fully Connected, out 32 \!+\! 32, Linear
		\end{itemize}
	\end{itemize}
\end{itemize}

\subsection{Compared Method}

\subsubsection{MulMON} MulMON \cite{Li2020Learning} is trained with the default hyperparameters described in the ``scripts/train\_clevr\_parallel.sh'' file of the official code repository\footnote{\url{https://github.com/NanboLi/MulMON}}
except: 1) the number of training steps is $\num[group-separator={,}]{600000}$; 2) the number of viewpoints for inference is sampled from $n \!\sim\! \mathcal{U}(1, 7)$ and the number of viewpoints for query is $8 - n$; 3) the number of slots $K \!+\! 1$ is $8$; 4) the channels of the last three convolutional layers are changed to $16$ and a $2 \!\times\! 2$ nearest neighbor upsample layer is added before the last convolutional layer in the decoder.

\subsubsection{SIMONe} The architecture and hyperparameters used to train SIMONe \cite{Kabra2021SIMONe} are similar to the ones described in the original paper except: 1) the number of training steps is $\num[group-separator={,}]{4000000}$; 2) the number of slots $K \!+\! 1$ is $8$; 3) for the CLEVR-1, SHOP-1, GSO, and ShapeNet datasets, the batch size is $4$ and the learning rate is $2 \!\times\! 10^{-4}$; 4) for the CLEVR-2 and SHOP-2 datasets, the batch size is $8$ and the learning rate is $2 \!\times\! 10^{-5}$.

\subsubsection{SAVi} SAVi \cite{Kipf2022Conditional} is trained using the official code repository\footnote{\url{https://github.com/google-research/slot-attention-video}} with the default hyperparameters described in the original paper except: 1) the number of training steps is $\num[group-separator={,}]{300000}$; 2) the number of slots $K \!+\! 1$ is $8$; 3) the batch size is 8;  4) the number of input frames is $8$.

\subsubsection{Ablation Method}

The ablation method is derived from the proposed OCLOC and use the same set of hyperparameters as OCLOC.

\section{Extra Experimental Results}

\begin{figure*}[p]
	\captionsetup[subfigure]{labelformat=empty}
	\centering
	\subfloat[(a) SAVi (video)]{\includegraphics[width=0.67\columnwidth]{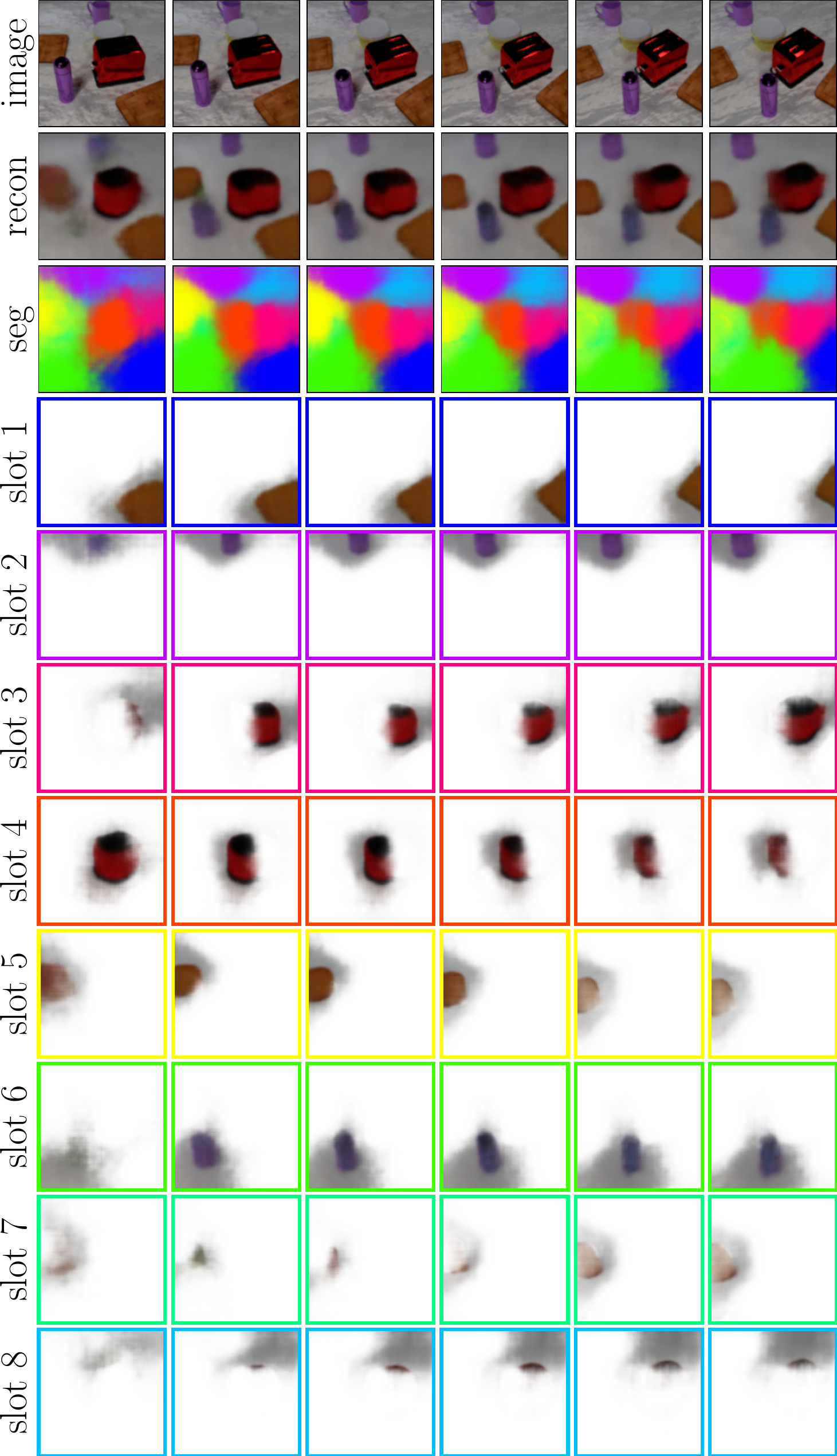}}%
	\hfill
	\subfloat[(b) SAVi]{\includegraphics[width=0.67\columnwidth]{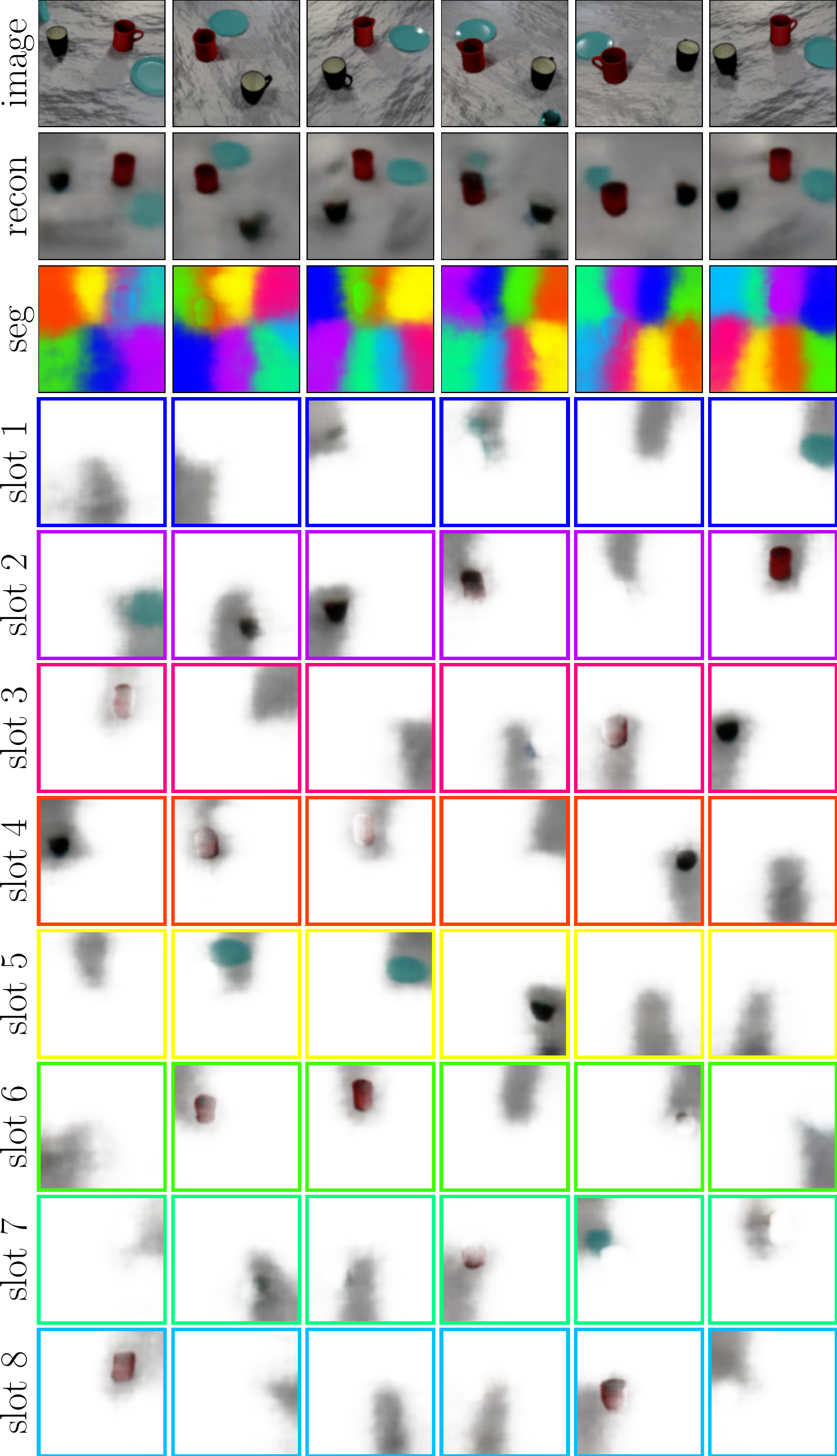}}%
	\hfill
	\subfloat[(c) SIMONe]{\includegraphics[width=0.67\columnwidth]{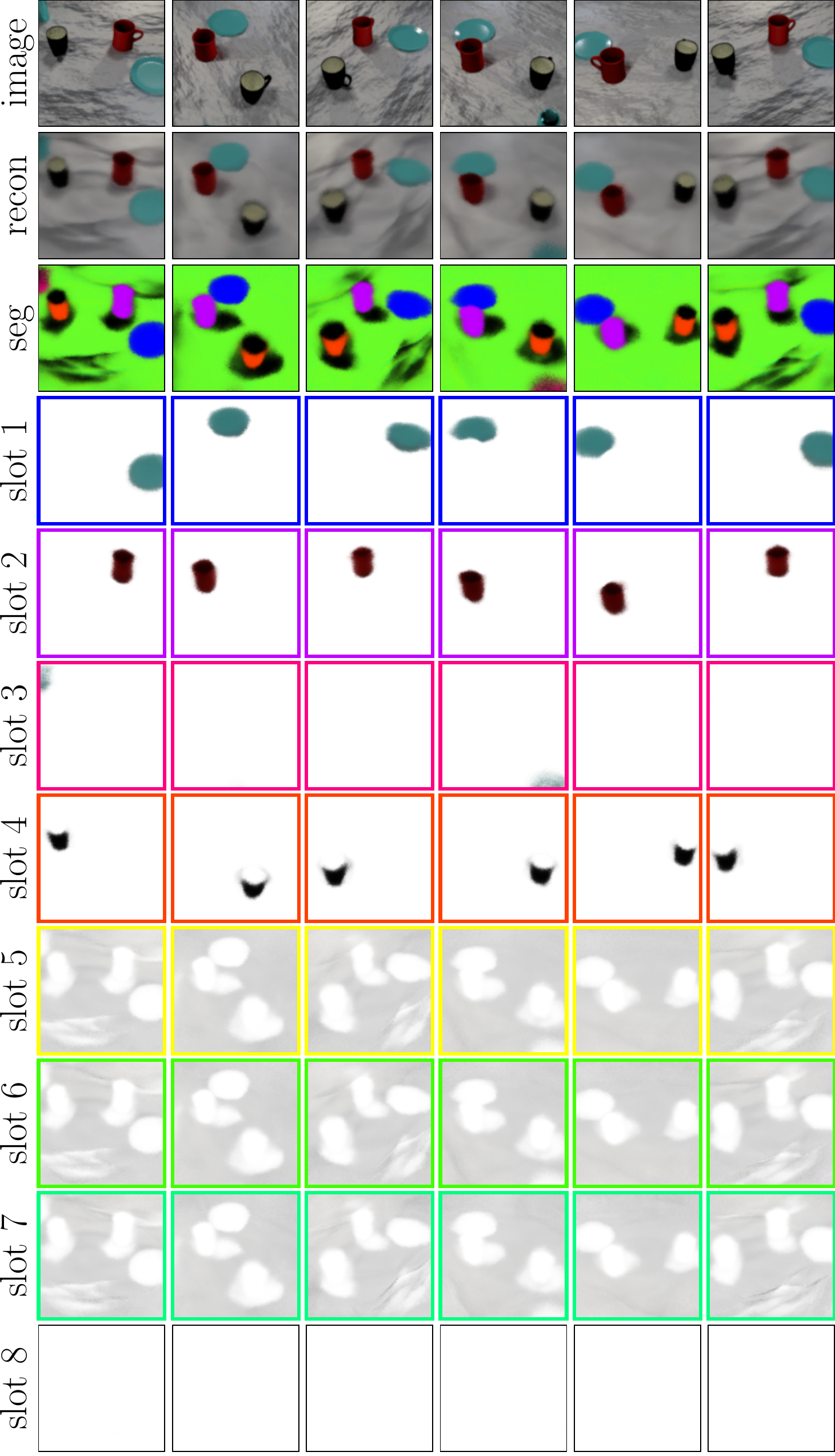}}%
	\\
	\subfloat[(d) MulMON]{\includegraphics[width=0.67\columnwidth]{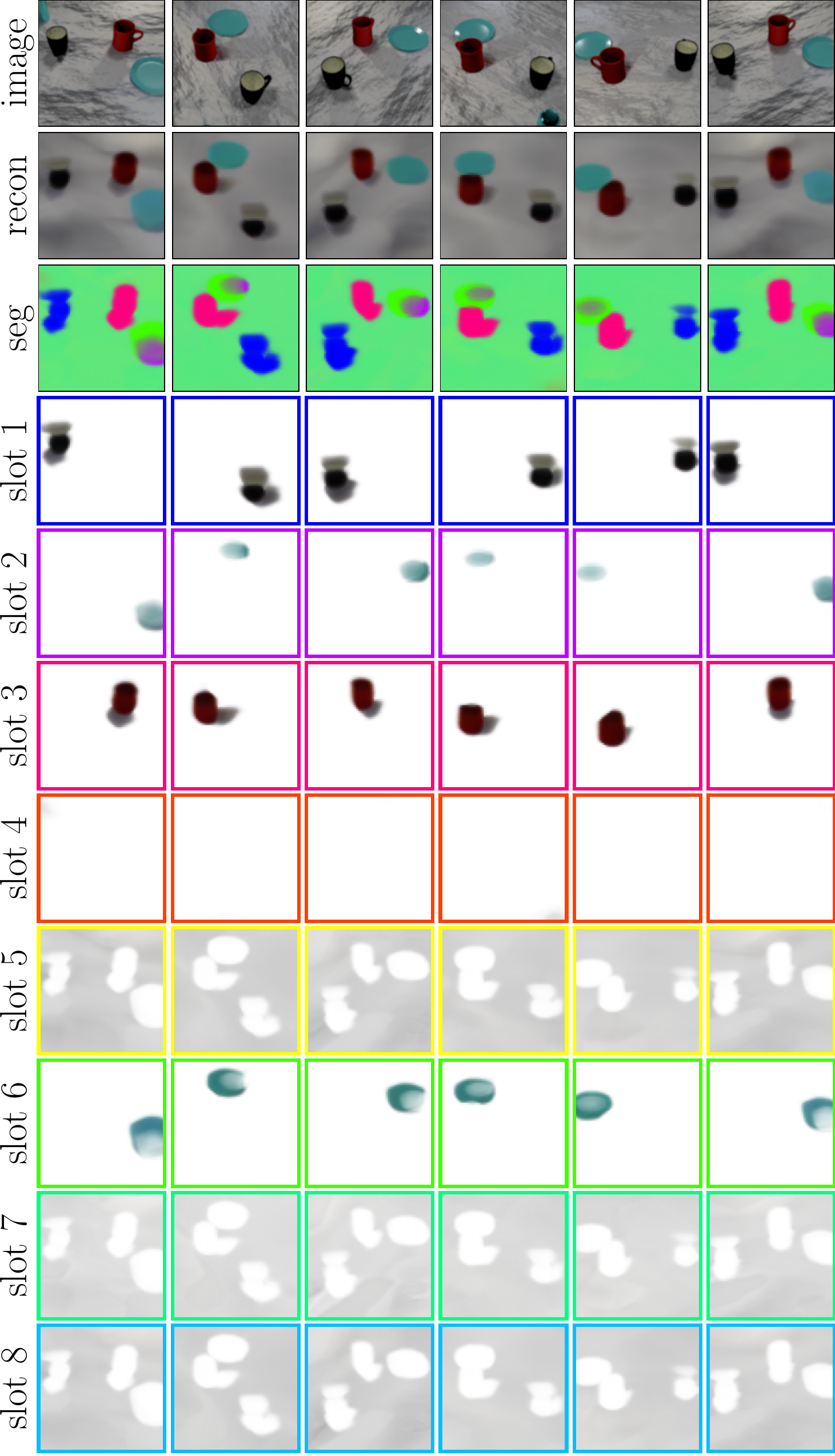}}%
	\hfill
	\subfloat[(e) Ablation]{\includegraphics[width=0.67\columnwidth]{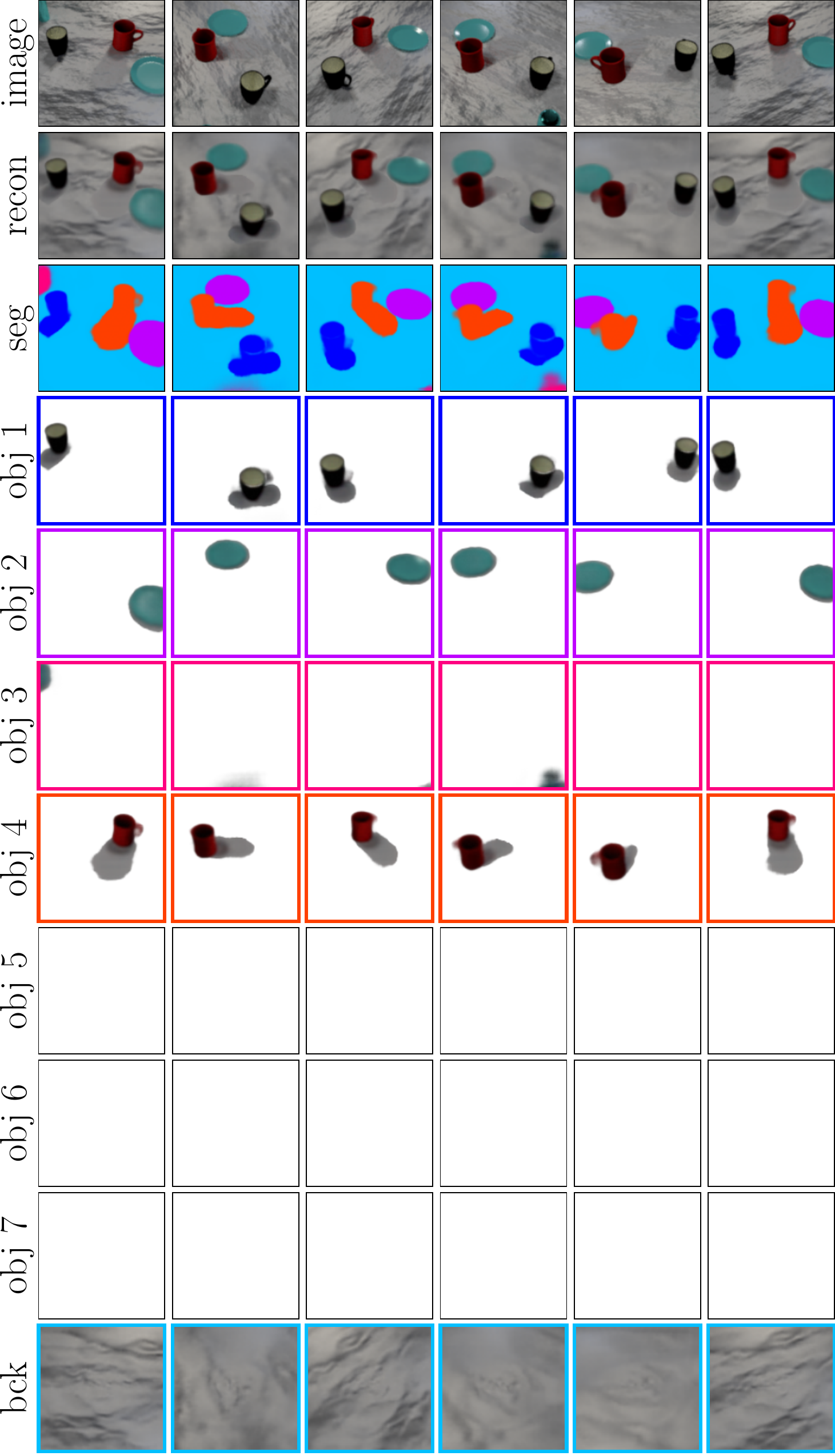}}%
	\hfill
	\subfloat[(f) OCLOC]{\includegraphics[width=0.67\columnwidth]{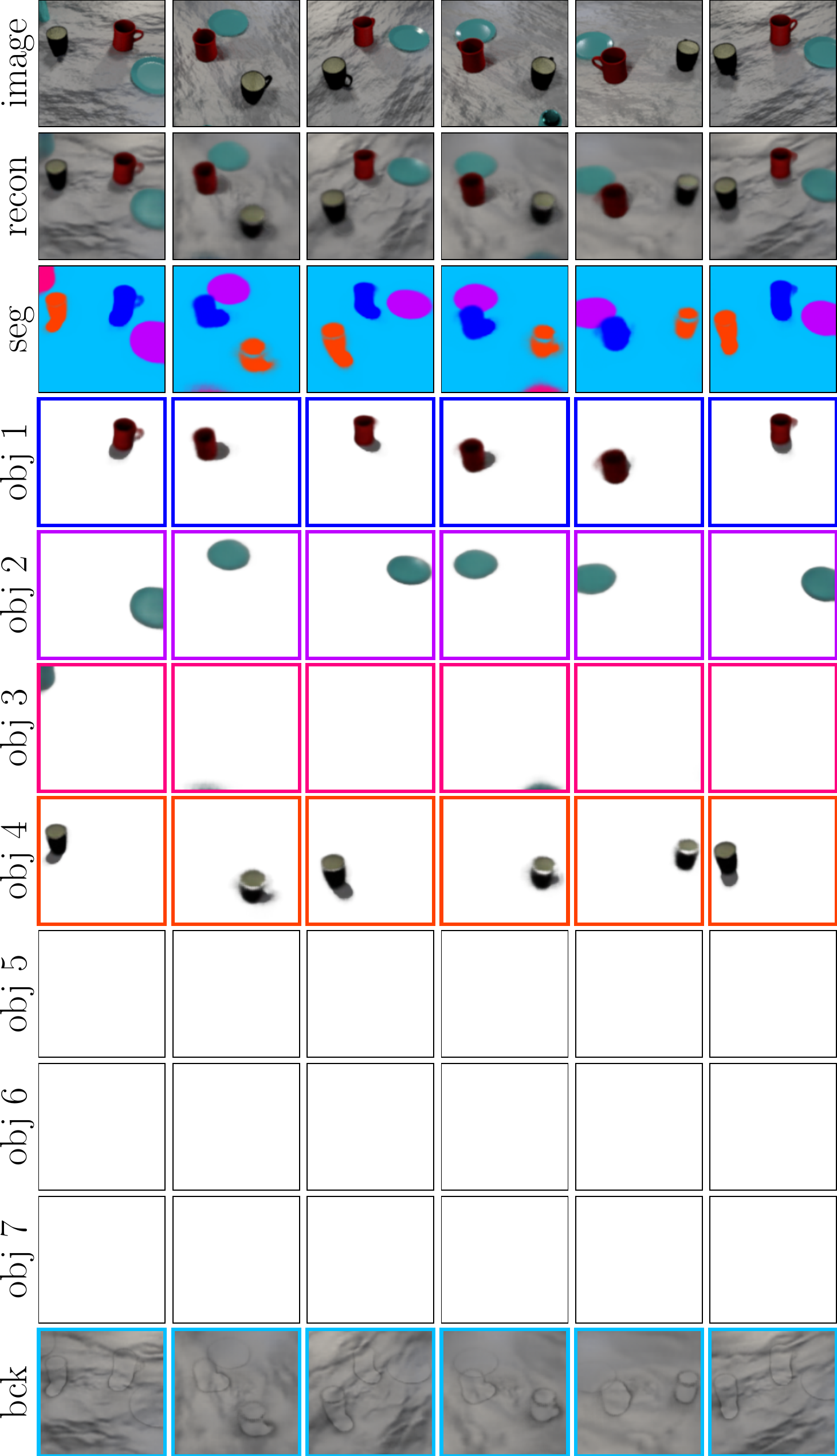}}%
	\caption{Scene decomposition results of different methods on the SHOP dataset.}
	\label{fig:decompose_multi_shop}
\end{figure*}

\begin{figure*}[p]
	\captionsetup[subfigure]{labelformat=empty}
	\centering
	\subfloat[(a) SAVi (video)]{\includegraphics[width=0.67\columnwidth]{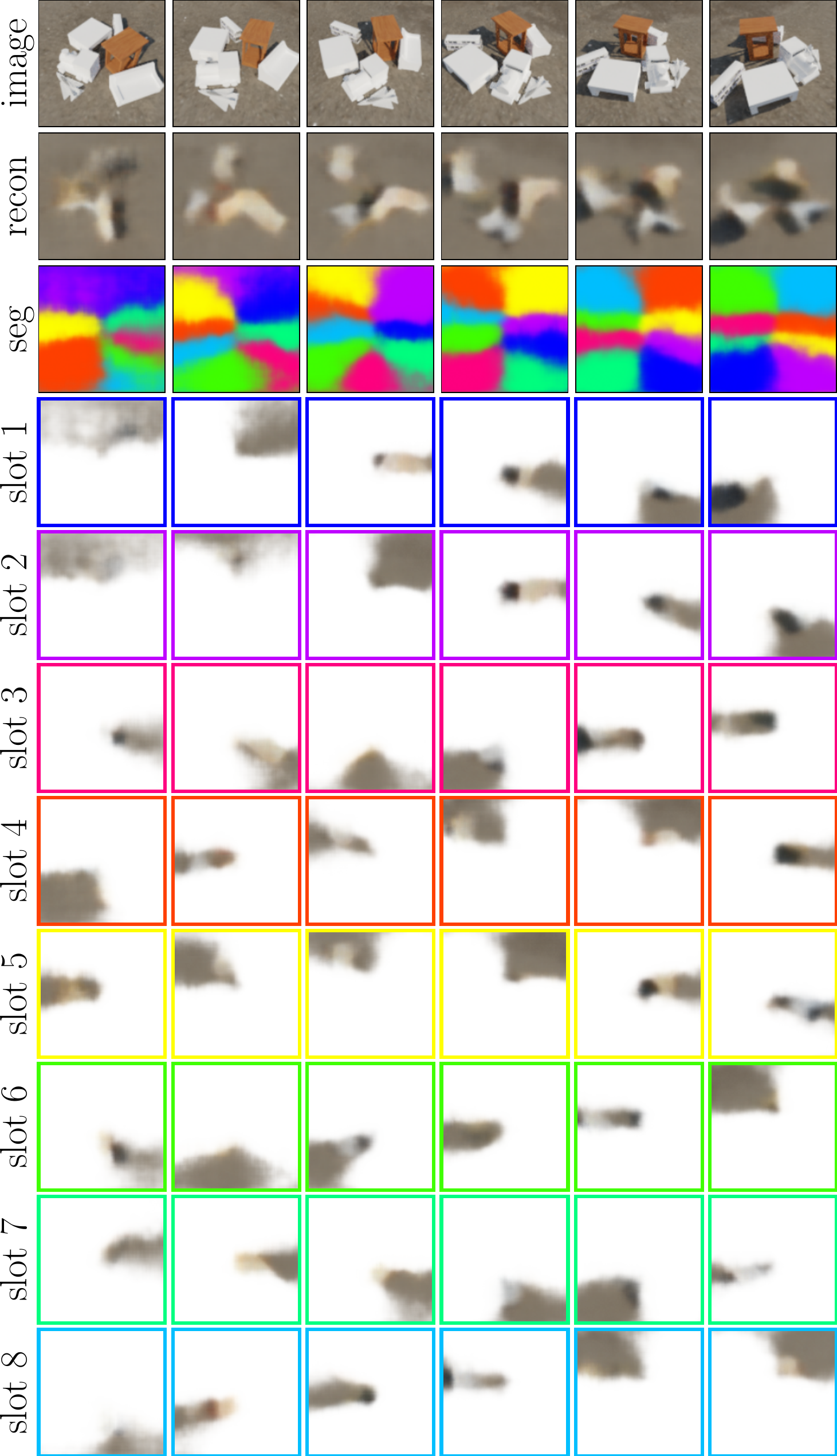}}%
	\hfill
	\subfloat[(b) SAVi]{\includegraphics[width=0.67\columnwidth]{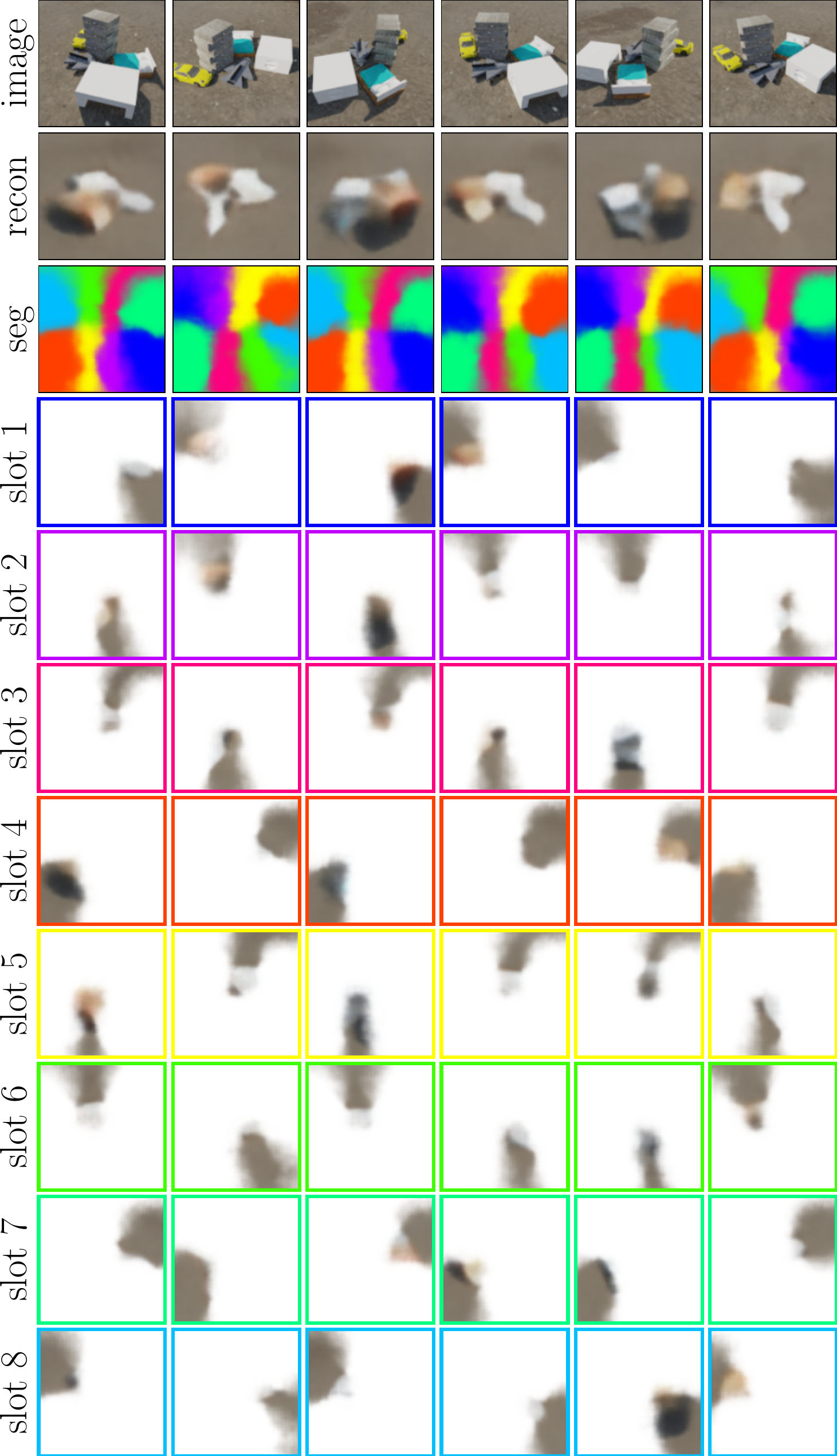}}%
	\hfill
	\subfloat[(c) SIMONe]{\includegraphics[width=0.67\columnwidth]{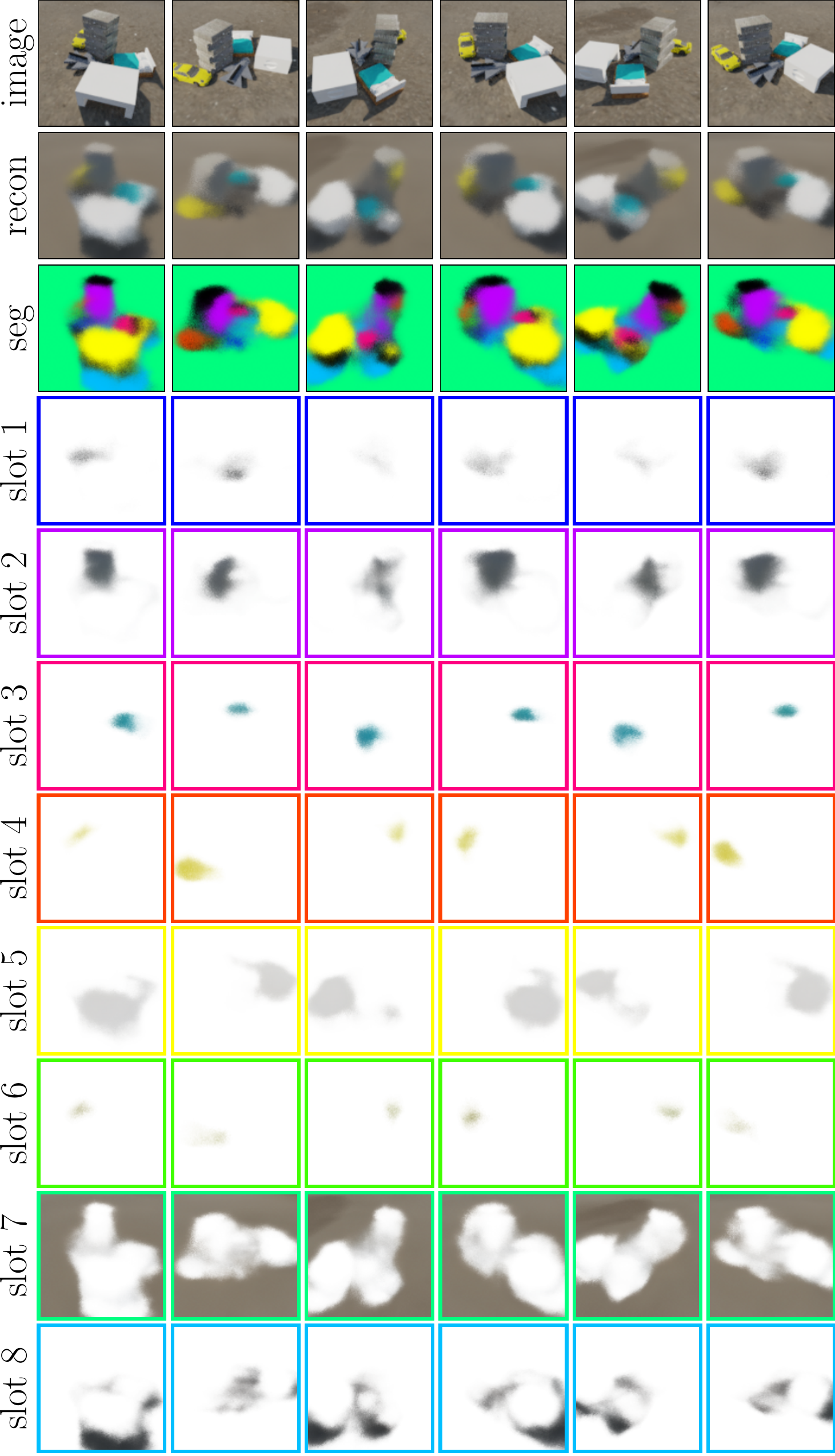}}%
	\\
	\subfloat[(d) MulMON]{\includegraphics[width=0.67\columnwidth]{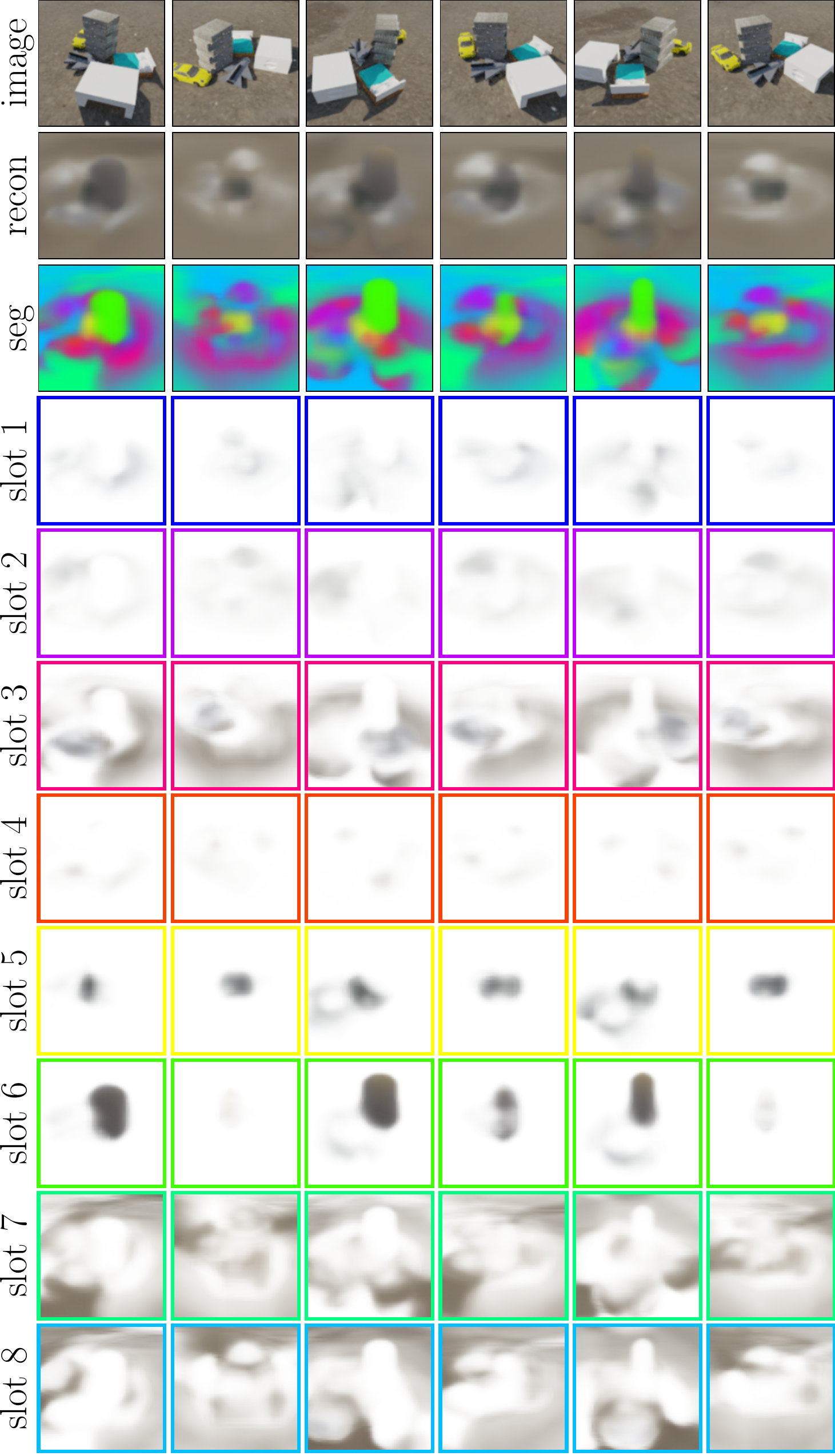}}%
	\hfill
	\subfloat[(e) Ablation]{\includegraphics[width=0.67\columnwidth]{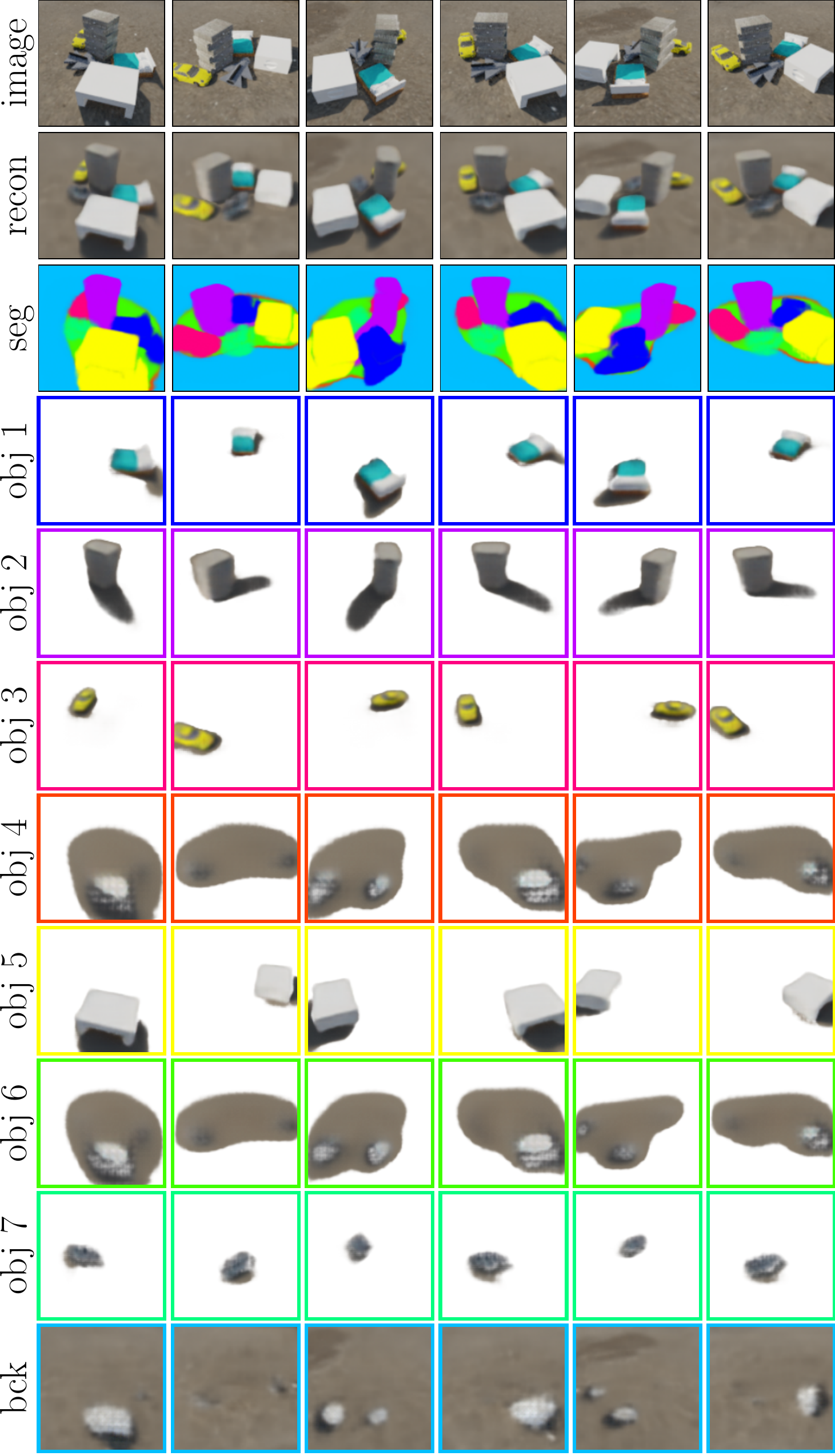}}%
	\hfill
	\subfloat[(f) OCLOC]{\includegraphics[width=0.67\columnwidth]{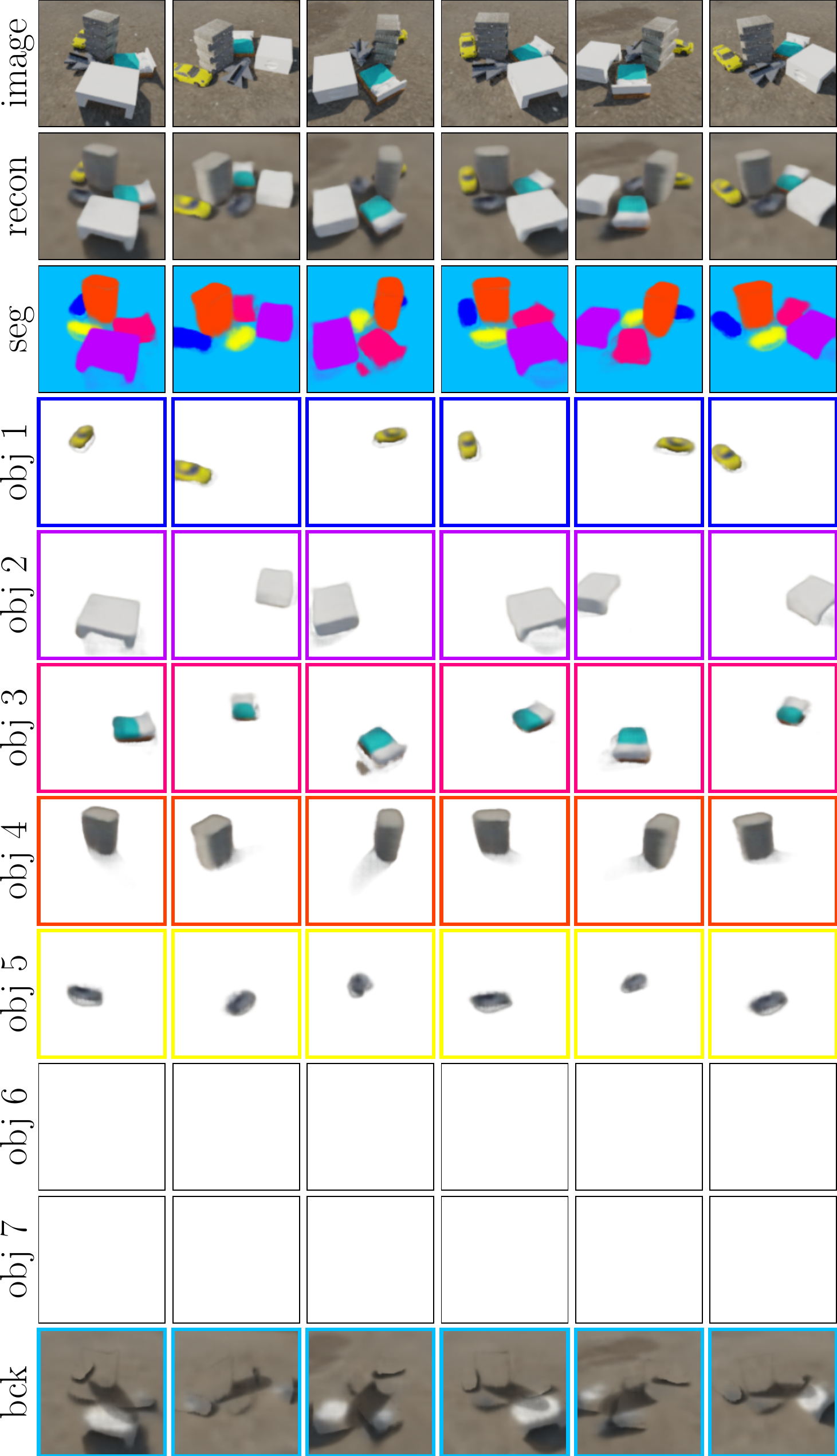}}%
	\caption{Scene decomposition results of different methods on the ShapeNet dataset.}
	\label{fig:decompose_multi_shapenet}
\end{figure*}

Samples of scene decomposition results on the SHOP and ShapeNet datasets are shown in Figure \ref{fig:decompose_multi_shop} and Figure \ref{fig:decompose_multi_shapenet}, respectively. The proposed method can separate different objects accurately on these datasets. Detailed quantitative results are shown in Tables \ref{tab:multi_8_test_1} and \ref{tab:multi_8_test_2}. The proposed method outperforms the compared methods in most cases.

\begin{table*}[ht]
	\centering
	\caption{Comparison of multi-viewpoint learning on the Test 1 splits. All methods are trained with $M \!\in\! [1, 8]$ and $K \!=\! 7$ and tested with $M \!=\! 8$ and $K \!=\! 7$. The top-2 scores are underlined, with the best in bold and the second best in italics.}
	\label{tab:multi_8_test_1}
	\begin{small}
		\addtolength{\tabcolsep}{-4pt}
		\begin{tabular}{c|c|C{0.67in}C{0.67in}C{0.67in}C{0.67in}C{0.67in}C{0.67in}C{0.67in}C{0.67in}}
			\toprule
			Dataset          &  Method  &          ARI-A          &          AMI-A          &          ARI-O          &          AMI-O          &           IoU           &           F1            &           OCA           &           OOA           \\ \midrule
			\multirow{6}{*}{CLEVR} & SAVi (video) & 0.047$\pm$6e-4 & 0.228$\pm$2e-3 & 0.805$\pm$1e-2 & 0.836$\pm$5e-3 & N/A & N/A & 0.000$\pm$0e-0 & N/A \\
 & SAVi & 0.004$\pm$2e-4 & 0.027$\pm$6e-4 & 0.070$\pm$2e-3 & 0.089$\pm$2e-3 & N/A & N/A & 0.000$\pm$0e-0 & N/A \\
 & SIMONe & 0.393$\pm$3e-5 & 0.444$\pm$9e-5 & 0.616$\pm$1e-4 & 0.677$\pm$1e-4 & N/A & N/A & 0.000$\pm$0e-0 & N/A \\
 & MulMON & \underline{\textit{0.581}}$\pm$6e-3 & \underline{\textit{0.545}}$\pm$3e-3 & \underline{\textit{0.907}}$\pm$6e-3 & \underline{\textit{0.888}}$\pm$4e-3 & N/A & N/A & \underline{\textbf{0.399}}$\pm$2e-2 & N/A \\
 & Ablation & 0.176$\pm$4e-7 & 0.318$\pm$8e-7 & 0.905$\pm$4e-6 & 0.876$\pm$4e-6 & \underline{\textit{0.344}}$\pm$1e-7 & \underline{\textit{0.481}}$\pm$2e-7 & 0.010$\pm$0e-0 & \underline{\textbf{0.970}}$\pm$1e-16 \\
 & OCLOC & \underline{\textbf{0.791}}$\pm$2e-6 & \underline{\textbf{0.692}}$\pm$3e-6 & \underline{\textbf{0.924}}$\pm$9e-7 & \underline{\textbf{0.926}}$\pm$1e-6 & \underline{\textbf{0.657}}$\pm$1e-7 & \underline{\textbf{0.777}}$\pm$9e-8 & \underline{\textit{0.320}}$\pm$0e-0 & \underline{\textit{0.956}}$\pm$0e-0 \\
\midrule
\multirow{6}{*}{SHOP} & SAVi (video) & 0.073$\pm$2e-3 & 0.276$\pm$2e-3 & 0.624$\pm$9e-3 & 0.757$\pm$5e-3 & N/A & N/A & 0.000$\pm$0e-0 & N/A \\
 & SAVi & 0.002$\pm$1e-4 & 0.016$\pm$8e-4 & 0.023$\pm$1e-3 & 0.047$\pm$3e-3 & N/A & N/A & 0.000$\pm$0e-0 & N/A \\
 & SIMONe & 0.199$\pm$3e-5 & 0.424$\pm$4e-5 & 0.718$\pm$6e-5 & 0.739$\pm$9e-5 & N/A & N/A & 0.000$\pm$0e-0 & N/A \\
 & MulMON & 0.366$\pm$1e-2 & 0.442$\pm$6e-3 & 0.658$\pm$1e-2 & 0.701$\pm$9e-3 & N/A & N/A & 0.054$\pm$3e-2 & N/A \\
 & Ablation & \underline{\textit{0.614}}$\pm$7e-7 & \underline{\textit{0.589}}$\pm$7e-7 & \underline{\textbf{0.900}}$\pm$8e-7 & \underline{\textbf{0.907}}$\pm$1e-6 & \underline{\textit{0.543}}$\pm$2e-7 & \underline{\textit{0.661}}$\pm$2e-7 & \underline{\textbf{0.380}}$\pm$0e-0 & \underline{\textbf{0.862}}$\pm$0e-0 \\
 & OCLOC & \underline{\textbf{0.677}}$\pm$1e-6 & \underline{\textbf{0.608}}$\pm$7e-7 & \underline{\textit{0.793}}$\pm$2e-6 & \underline{\textit{0.836}}$\pm$2e-6 & \underline{\textbf{0.554}}$\pm$2e-7 & \underline{\textbf{0.666}}$\pm$2e-7 & \underline{\textit{0.090}}$\pm$0e-0 & \underline{\textit{0.825}}$\pm$0e-0 \\
\midrule
\multirow{6}{*}{GSO} & SAVi (video) & 0.020$\pm$3e-4 & 0.093$\pm$1e-3 & 0.199$\pm$3e-3 & 0.275$\pm$3e-3 & N/A & N/A & 0.000$\pm$0e-0 & N/A \\
 & SAVi & 0.004$\pm$7e-5 & 0.027$\pm$4e-4 & 0.048$\pm$9e-4 & 0.084$\pm$1e-3 & N/A & N/A & 0.000$\pm$0e-0 & N/A \\
 & SIMONe & 0.243$\pm$2e-5 & 0.338$\pm$2e-5 & 0.311$\pm$7e-5 & 0.413$\pm$6e-5 & N/A & N/A & 0.000$\pm$0e-0 & N/A \\
 & MulMON & 0.247$\pm$5e-3 & 0.202$\pm$3e-3 & 0.212$\pm$9e-3 & 0.269$\pm$5e-3 & N/A & N/A & \underline{\textit{0.030}}$\pm$6e-3 & N/A \\
 & Ablation & \underline{\textit{0.455}}$\pm$2e-6 & \underline{\textit{0.484}}$\pm$9e-7 & \underline{\textit{0.896}}$\pm$2e-6 & \underline{\textit{0.852}}$\pm$1e-6 & \underline{\textit{0.531}}$\pm$2e-7 & \underline{\textit{0.684}}$\pm$2e-7 & 0.000$\pm$0e-0 & \underline{\textit{0.968}}$\pm$0e-0 \\
 & OCLOC & \underline{\textbf{0.856}}$\pm$3e-6 & \underline{\textbf{0.765}}$\pm$3e-6 & \underline{\textbf{0.946}}$\pm$3e-6 & \underline{\textbf{0.919}}$\pm$4e-6 & \underline{\textbf{0.746}}$\pm$4e-7 & \underline{\textbf{0.847}}$\pm$3e-7 & \underline{\textbf{0.820}}$\pm$0e-0 & \underline{\textbf{0.985}}$\pm$1e-16 \\
\midrule
\multirow{6}{*}{ShapeNet} & SAVi (video) & 0.008$\pm$3e-4 & 0.058$\pm$1e-3 & 0.112$\pm$3e-3 & 0.182$\pm$4e-3 & N/A & N/A & 0.000$\pm$0e-0 & N/A \\
 & SAVi & 0.005$\pm$5e-5 & 0.019$\pm$3e-4 & 0.034$\pm$8e-4 & 0.054$\pm$1e-3 & N/A & N/A & 0.000$\pm$0e-0 & N/A \\
 & SIMONe & \underline{\textit{0.566}}$\pm$9e-5 & 0.441$\pm$1e-4 & 0.343$\pm$1e-4 & 0.452$\pm$2e-4 & N/A & N/A & 0.000$\pm$0e-0 & N/A \\
 & MulMON & 0.192$\pm$6e-3 & 0.197$\pm$2e-3 & 0.239$\pm$8e-3 & 0.278$\pm$5e-3 & N/A & N/A & \underline{\textit{0.019}}$\pm$1e-2 & N/A \\
 & Ablation & 0.403$\pm$1e-6 & \underline{\textit{0.454}}$\pm$1e-6 & \underline{\textit{0.872}}$\pm$2e-6 & \underline{\textit{0.834}}$\pm$3e-6 & \underline{\textit{0.498}}$\pm$9e-8 & \underline{\textit{0.652}}$\pm$8e-8 & 0.000$\pm$0e-0 & \underline{\textit{0.938}}$\pm$0e-0 \\
 & OCLOC & \underline{\textbf{0.805}}$\pm$3e-6 & \underline{\textbf{0.711}}$\pm$4e-6 & \underline{\textbf{0.922}}$\pm$2e-6 & \underline{\textbf{0.902}}$\pm$2e-6 & \underline{\textbf{0.668}}$\pm$3e-7 & \underline{\textbf{0.787}}$\pm$2e-7 & \underline{\textbf{0.610}}$\pm$0e-0 & \underline{\textbf{0.945}}$\pm$0e-0 \\
\bottomrule
		\end{tabular}
	\end{small}
\end{table*}
\begin{table*}[ht]
	\centering
	\caption{Comparison of multi-viewpoint learning on the Test 2 splits. All methods are trained with $M \!\in\! [1, 8]$ and $K \!=\! 7$ and tested with $M \!=\! 8$ and $K \!=\! 11$. The top-2 scores are underlined, with the best in bold and the second best in italics.}
	\label{tab:multi_8_test_2}
	\begin{small}
		\addtolength{\tabcolsep}{-4pt}
		\begin{tabular}{c|c|C{0.67in}C{0.67in}C{0.67in}C{0.67in}C{0.67in}C{0.67in}C{0.67in}C{0.67in}}
			\toprule
			Dataset          &  Method  &          ARI-A          &          AMI-A          &          ARI-O          &          AMI-O          &           IoU           &           F1            &           OCA           &           OOA           \\ \midrule
			\multirow{6}{*}{CLEVR} & SAVi (video) & 0.051$\pm$9e-4 & 0.314$\pm$1e-3 & 0.771$\pm$7e-3 & 0.827$\pm$4e-3 & N/A & N/A & 0.000$\pm$0e-0 & N/A \\
 & SAVi & 0.005$\pm$1e-4 & 0.048$\pm$3e-4 & 0.068$\pm$2e-3 & 0.126$\pm$6e-4 & N/A & N/A & 0.000$\pm$0e-0 & N/A \\
 & SIMONe & 0.333$\pm$7e-5 & 0.408$\pm$7e-5 & 0.545$\pm$1e-4 & 0.615$\pm$1e-4 & N/A & N/A & 0.000$\pm$0e-0 & N/A \\
 & MulMON & \underline{\textit{0.558}}$\pm$5e-3 & \underline{\textbf{0.574}}$\pm$3e-3 & \underline{\textbf{0.891}}$\pm$3e-3 & \underline{\textbf{0.881}}$\pm$2e-3 & N/A & N/A & \underline{\textit{0.176}}$\pm$2e-2 & N/A \\
 & Ablation & 0.128$\pm$3e-7 & 0.351$\pm$1e-6 & 0.829$\pm$3e-6 & 0.822$\pm$3e-6 & \underline{\textit{0.230}}$\pm$5e-8 & \underline{\textit{0.340}}$\pm$5e-8 & 0.000$\pm$0e-0 & \underline{\textbf{0.928}}$\pm$0e-0 \\
 & OCLOC & \underline{\textbf{0.596}}$\pm$1e-6 & \underline{\textbf{0.574}}$\pm$1e-6 & \underline{\textit{0.852}}$\pm$1e-6 & \underline{\textit{0.873}}$\pm$1e-6 & \underline{\textbf{0.473}}$\pm$2e-8 & \underline{\textbf{0.602}}$\pm$4e-8 & \underline{\textbf{0.260}}$\pm$0e-0 & \underline{\textit{0.841}}$\pm$0e-0 \\
\midrule
\multirow{6}{*}{SHOP} & SAVi (video) & 0.069$\pm$8e-4 & 0.344$\pm$2e-3 & 0.608$\pm$3e-3 & 0.765$\pm$3e-3 & N/A & N/A & 0.000$\pm$0e-0 & N/A \\
 & SAVi & 0.003$\pm$8e-5 & 0.035$\pm$2e-4 & 0.028$\pm$1e-4 & 0.077$\pm$5e-4 & N/A & N/A & 0.000$\pm$0e-0 & N/A \\
 & SIMONe & 0.177$\pm$4e-5 & 0.385$\pm$3e-5 & 0.601$\pm$2e-5 & 0.631$\pm$4e-5 & N/A & N/A & 0.000$\pm$0e-0 & N/A \\
 & MulMON & 0.355$\pm$1e-2 & 0.473$\pm$4e-3 & 0.645$\pm$4e-3 & 0.720$\pm$2e-3 & N/A & N/A & 0.086$\pm$1e-2 & N/A \\
 & Ablation & \underline{\textit{0.401}}$\pm$2e-6 & \underline{\textit{0.509}}$\pm$7e-7 & \underline{\textbf{0.844}}$\pm$2e-6 & \underline{\textbf{0.854}}$\pm$1e-6 & \underline{\textbf{0.431}}$\pm$7e-8 & \underline{\textbf{0.557}}$\pm$8e-8 & \underline{\textbf{0.200}}$\pm$0e-0 & \underline{\textbf{0.889}}$\pm$4e-4 \\
 & OCLOC & \underline{\textbf{0.494}}$\pm$9e-7 & \underline{\textbf{0.523}}$\pm$1e-6 & \underline{\textit{0.782}}$\pm$2e-6 & \underline{\textit{0.815}}$\pm$2e-6 & \underline{\textit{0.424}}$\pm$1e-7 & \underline{\textit{0.541}}$\pm$1e-7 & \underline{\textit{0.090}}$\pm$0e-0 & \underline{\textit{0.748}}$\pm$0e-0 \\
\midrule
\multirow{6}{*}{GSO} & SAVi (video) & 0.025$\pm$3e-4 & 0.143$\pm$1e-3 & 0.185$\pm$3e-3 & 0.305$\pm$3e-3 & N/A & N/A & 0.000$\pm$0e-0 & N/A \\
 & SAVi & 0.006$\pm$2e-4 & 0.048$\pm$3e-4 & 0.045$\pm$4e-4 & 0.107$\pm$8e-4 & N/A & N/A & 0.000$\pm$0e-0 & N/A \\
 & SIMONe & 0.223$\pm$2e-5 & 0.299$\pm$3e-5 & 0.217$\pm$3e-5 & 0.340$\pm$3e-5 & N/A & N/A & 0.000$\pm$0e-0 & N/A \\
 & MulMON & 0.325$\pm$9e-3 & 0.387$\pm$3e-3 & 0.452$\pm$8e-3 & 0.538$\pm$4e-3 & N/A & N/A & 0.014$\pm$1e-2 & N/A \\
 & Ablation & \underline{\textit{0.393}}$\pm$1e-6 & \underline{\textit{0.492}}$\pm$6e-7 & \underline{\textit{0.791}}$\pm$2e-6 & \underline{\textit{0.778}}$\pm$1e-6 & \underline{\textit{0.442}}$\pm$1e-7 & \underline{\textit{0.595}}$\pm$1e-7 & \underline{\textit{0.020}}$\pm$0e-0 & \underline{\textit{0.931}}$\pm$1e-16 \\
 & OCLOC & \underline{\textbf{0.750}}$\pm$8e-7 & \underline{\textbf{0.670}}$\pm$9e-7 & \underline{\textbf{0.838}}$\pm$2e-6 & \underline{\textbf{0.823}}$\pm$2e-6 & \underline{\textbf{0.600}}$\pm$2e-5 & \underline{\textbf{0.720}}$\pm$3e-5 & \underline{\textbf{0.320}}$\pm$0e-0 & \underline{\textbf{0.946}}$\pm$2e-4 \\
\midrule
\multirow{6}{*}{ShapeNet} & SAVi (video) & 0.013$\pm$1e-4 & 0.096$\pm$6e-4 & 0.108$\pm$7e-4 & 0.208$\pm$1e-3 & N/A & N/A & 0.000$\pm$0e-0 & N/A \\
 & SAVi & 0.006$\pm$1e-4 & 0.039$\pm$5e-4 & 0.038$\pm$8e-4 & 0.083$\pm$1e-3 & N/A & N/A & 0.000$\pm$0e-0 & N/A \\
 & SIMONe & \underline{\textit{0.428}}$\pm$3e-5 & 0.333$\pm$4e-5 & 0.224$\pm$4e-5 & 0.382$\pm$5e-5 & N/A & N/A & 0.000$\pm$0e-0 & N/A \\
 & MulMON & 0.258$\pm$7e-3 & 0.369$\pm$2e-3 & 0.446$\pm$3e-3 & 0.524$\pm$2e-3 & N/A & N/A & 0.002$\pm$4e-3 & N/A \\
 & Ablation & 0.351$\pm$2e-6 & \underline{\textit{0.470}}$\pm$2e-6 & \underline{\textit{0.753}}$\pm$2e-6 & \underline{\textit{0.755}}$\pm$3e-6 & \underline{\textit{0.412}}$\pm$9e-8 & \underline{\textit{0.561}}$\pm$9e-8 & \underline{\textit{0.020}}$\pm$0e-0 & \underline{\textit{0.869}}$\pm$0e-0 \\
 & OCLOC & \underline{\textbf{0.665}}$\pm$3e-6 & \underline{\textbf{0.611}}$\pm$3e-6 & \underline{\textbf{0.794}}$\pm$1e-6 & \underline{\textbf{0.798}}$\pm$2e-6 & \underline{\textbf{0.524}}$\pm$1e-7 & \underline{\textbf{0.653}}$\pm$8e-8 & \underline{\textbf{0.200}}$\pm$0e-0 & \underline{\textbf{0.890}}$\pm$1e-16 \\
\bottomrule
		\end{tabular}
	\end{small}
\end{table*}
\end{appendices}

\end{document}